\documentclass[10pt,journal,compsoc]{IEEEtran}

\ifCLASSOPTIONcompsoc
  \usepackage[nocompress]{cite}
\else
  \usepackage{cite}
\fi
\ifCLASSINFOpdf
\else
\fi
\usepackage[utf8]{inputenc} 
\usepackage[T1]{fontenc}    
\usepackage{hyperref}       
\usepackage{url}            
\usepackage{booktabs}       
\usepackage{amsfonts}       
\usepackage{nicefrac}       
\usepackage{microtype}      
\usepackage{graphicx}
\usepackage{wrapfig}
\usepackage{amsmath}
\usepackage{algorithm,algpseudocode}
\usepackage{multirow}
\usepackage{wrapfig}
\usepackage{bm}
\usepackage{bbm}
\usepackage{pgf, tikz}
\usepackage[keeplastbox]{flushend}
\newcommand{\tabincell}[2]{\begin{tabular}{@{}#1@{}}#2\end{tabular}}

\begin{document}
%
\title{A Survey on Machine Learning from\\ Few Samples}
%
%
%
%

\author{Jiang~Lu,
        Pinghua Gong,
        Jieping Ye,~\IEEEmembership{Fellow,~IEEE},
        Jianwei Zhang,~\IEEEmembership{Member,~IEEE},\\
        and~Changshui Zhang,~\IEEEmembership{Fellow,~IEEE}
\IEEEcompsocitemizethanks{
\IEEEcompsocthanksitem J. Lu 
and C. Zhang are with the Institute for Artificial Intelligence, Tsinghua University (THUAI), the State Key Laboratory of Intelligence Technologies and Systems, the Beijing National Research Center for Information Science and Technologies (BNRist) , the Department of Automation, Tsinghua University, Beijing 100084, China. E-mail: lu-j13@tsinghua.org.cn; zcs@mail.tsinghua.edu.cn.

\IEEEcompsocthanksitem P. Gong is with the Didi Research Institute, Didi Chuxing, Beijing 100085, China. E-mail:  gongpinghua@didichuxing.com.

\IEEEcompsocthanksitem J. Ye is with Didi AI Labs, Beijing 100085, China and University of Michigan, Ann Arbor. E-mail: jpye@umich.edu.

\IEEEcompsocthanksitem J. Zhang is with the Faculty of Mathematics Computer Science and Natural Sciences, TAMS Group, University of Hamburg, Hamburg 20146, Germany.  E-mail: zhang@informatik.uni-hamburg.de.

}
}

%
%

\markboth{}%
{Lu \MakeLowercase{\textit{et al.}}: A Survey on Machine Learning from Few Samples}
%



\IEEEtitleabstractindextext{%
\begin{abstract}
Few sample learning (FSL) is significant and challenging in the field of machine learning. The capability of learning and generalizing from very few samples successfully is a noticeable demarcation separating artificial intelligence and human intelligence since humans can readily establish their cognition to novelty from just a single or a handful of examples whereas machine learning algorithms typically entail hundreds or thousands of  supervised samples to guarantee generalization ability. Despite the long history dated back to the early 2000s and the widespread attention in recent years with booming deep learning technologies, little surveys or reviews for FSL are available until now. In this context, we extensively review 300+ papers of FSL spanning from the 2000s to 2019 and provide a timely and comprehensive  survey for FSL. In this survey, we review the evolution history as well as the current progress on FSL, categorize FSL approaches into the generative model based and discriminative model based kinds in principle, and emphasize particularly on the meta learning based FSL approaches. We also summarize several recently emerging extensional topics of FSL and review the latest advances on these topics. Furthermore, we highlight the important FSL applications covering many research hotspots in computer vision, natural language processing, audio and speech, reinforcement learning and robotic, data analysis, etc. Finally, we conclude the survey with a discussion on promising trends in the hope of providing guidance and insights to follow-up researches.
\end{abstract} 

\begin{IEEEkeywords}
few sample learning, learn to learn, survey, few-shot learning, meta learning
\end{IEEEkeywords}}

\maketitle

\IEEEdisplaynontitleabstractindextext

%
\IEEEpeerreviewmaketitle

\IEEEraisesectionheading{\section{Introduction}\label{sec:introduction}}
\IEEEPARstart{O}{ne} impressive hallmark of human intelligence is the ability to rapidly establish cognition to novel concepts from just a single or a handful of examples.  Many cognitive and psychological evidences~\cite{landau1988importance,markman1989categorization,xu2007word} have shown that humans can recognize visual objects through very few images~\cite{biederman1987recognition} and even children can remember a novel word by a single encounter~\cite{carey1978acquiring,clark2009first}. Although exactly what support the human capability of learning and generalizing from very few samples remains a profound mystery, some neurobiological works~\cite{rougier2005prefrontal,braver2009flexible,kehagia2010learning} have argued that the prominent human learning ability benefits from prefrontal cortex (PFC) and working memory in human brain, especially the interaction between PFC-specific neurobiological mechanism and previous experience stored in the brain. By contrast, most cutting-edge machine learning algorithms are data-hungry, especially the most widely known deep learning~\cite{lecun2015deep} that has pushed artificial intelligence to a new climax. As an important milestone in the development of machine learning, deep learning has scored remarkable achievement in a broad spectrum of research fields including vision~\cite{krizhevsky2012imagenet,szegedy2015going,he2016deep}, language~\cite{mikolov2010recurrent,sutskever2014sequence}, speech~\cite{hinton2012deep}, game~\cite{silver2016mastering}, demography~\cite{gebru2017using}, medicine~\cite{esteva2017dermatologist}, phytopathology~\cite{ghosal2018explainable} and zoology~\cite{norouzzadeh2018automatically}, etc. Generally, the  successes of deep learning can be owned to three key factors: powerful computing resources (e.g., GPU), sophisticated neural networks (e.g., CNN~\cite{krizhevsky2012imagenet}, LSTM~\cite{hochreiter1997long}) and large-scale datasets (e.g., ImageNet~\cite{russakovsky2015imagenet}, Pascal-VOC~\cite{everingham2010pascal}). However, many realistic application scenarios, such as in the field of medicine, military and finance, do not allow us access sufficient labeled training samples, due to some factors including privacy, security or high labeling costs for data, etc. Thus, it becomes an eagerly-awaited blueprint for almost all machine learning researchers to enable learning systems to efficiently learn and generalize from very few samples.

From a high-level perspective, the theoretical and practical significance of studying few sample learning (FSL) mainly comes from three aspects. First, the FSL approach is expected not to rely on large-scale training samples, thus eschewing the prohibitive costs on data preparation in some specific applications. Second, FSL can shrink the gap between human intelligence and artificial intelligence, being a necessary trip to develop universal AI~\cite{legg2007universal}. Third, FSL can achieve a low-cost and quick model deployment for one emerging task for which just a few samples are temporarily available, beneficial to shed light on the potential laws earlier in the task. 

Despite these encouraging virtues, the research of FSL progresses more slowly in the past decades than that of large sample learning due to its intrinsic difficulty. Clearly, we illustrate this difficulty from the optimization viewpoint. Consider a general machine learning problem, which is described by a prepared  supervised training set $\mathcal{D}_{t}=\{(x_i, y_i)\}_{i=1}^{n}$ with $x\in \mathcal{X}$, $y\in\mathcal{Y}$ drawn from the joint distribution $P_{\mathcal{X} \times \mathcal{Y}}$. The goal of the learning algorithm is to produce a mapping function $f \in \mathcal{F}: \mathcal{X} \rightarrow \mathcal{Y}$ such that the expected error $\mathcal{E}_{\mathrm{ex}}=\mathbb{E}_{(x,y)\sim P_{\mathcal{X} \times \mathcal{Y}}} L(f(x), y)$ is minimized, where $L(f(x), y)$ denotes the loss that compares the prediction $f(x)$ to its supervision target $y$. In fact, the joint distribution $P_{\mathcal{X} \times \mathcal{Y}}$ is unknown, and thus the learning algorithms are intended to minimize the empirical error $\mathcal{E}_{\mathrm{em}}=\mathbb{E}_{(x,y)\sim \mathcal{D}_t} L(f(x), y)$. In this context, a typical problem is that if the function space $\mathcal{F}$ from which the learning algorithm selects $f$ is too large, the generalization error $\mathcal{E}=|\mathcal{E}_{\mathrm{ex}}-\mathcal{E}_{\mathrm{em}}|$ would become big and thereby overfitting may arise easily. We can re-look the problem from the following perspective
\begin{equation}
\mathop{\min}\nolimits_{f} \mathcal{E}_{em},\quad
\mathrm{s.t.}\ \  f(x_i) = y_i,\  \forall (x_i,y_i)\in \mathcal{D}_t.
\label{eq:optimization}
\end{equation} 
If $\mathcal{D}_t$ contains more supervised samples, there will be more constraints on $f$, which implies the  space of function $f$ will be smaller, then it will bring a good generalization. Conversely, a scarce supervised training set would naturally lead to a poor generalization performance. Essentially, the constraint formed by each supervised sample can be regarded as a regularization on the function $f$, which is able to compress the redundant optional space of function $f$ and thereby reduce its generalization error.
Thus, it can be concluded that if one learning algorithm deal with one FSL task just by the vanilla learning techniques without any sophisticated learning strategies or specific network design, the learning algorithm would be faced with the serious overfitting. 

Few sample learning (FSL), also known as small or one sample learning,  few-shot or one-shot learning, can date back to the early 2000s. 
Despite the nearly 20 years of research history and its importance at the level of theory and application, few related surveys or reviews are available until now. In this article, we extensively investigate almost all FSL-related scientific papers spanning from the 2000s to 2019 to elaborate a systematic FSL survey. We must emphasize that the FSL  discussed here is orthogonal to zero-shot learning (ZSL)~\cite{wang2019survey}, which is another hot topic for machine learning. The setting of ZSL entails concept-specific side information to support the cross-concept knowledge transfer, varying greatly from that of FSL. To our best knowledge, there only have two FSL-related preprinted surveys~\cite{shu2018small,wang2019few} until now. Compared with them, the novelties and contributions of this survey mainly come from five major aspects:

(1) We give a more comprehensive and timely review which encompasses 300+ FSL-related papers spanning from the 2000s to 2019, covering all FSL approaches from the very earliest Congealing model~\cite{miller2000learning} to the latest meta learning approaches. The exhaustive exposition is conducive to the grasp of the whole development process of FSL as well as the construction of the complete knowledge hierarchy to FSL.

(2) We provide an understandable hierarchical taxonomy that categorizes existing FSL approaches into the generative model based approaches and discriminative model based approaches in light of their modeling principles to FSL problems. Within each class, we further conduct a more detailed categorization according to the generalizable properties.

(3) We put emphasis on current mainstream FSL approaches, i.e., the meta learning based FSL approaches, and categorize them into five major classes in light of what they hope to learn to learn via meta learning strategy, including Learn-to-Measure, Learn-to-Finetune,  Learn-to-Parameterize, Learn-to-Adjust and Learn-to-Remember. Moreover, the underlying development relationship between various meta learning based FSL approaches is revealed in this survey.

(4) We conclude several extensional research topics beyond  vanilla FSL that are emerging lately and review the latest advances towards these topics. These topics include Semi-supervised FSL, Unsupervised FSL, Cross-domain FSL, Generalized FSL and Multimodal FSL, which are challenging whilst endowing prominent practical significance to the solution for many realistic machine learning problems. These extensional topics were rarely covered by previous reviews.

(5) We extensively summarize existing FSL applications in various fields including computer vision, natural language processing, audio and speech, reinforcement learning and robotic, data analysis, etc, and current FSL performance on benchmarks, aiming to provide a handbook for follow-up researches, which were not studied  by previous reviews. 


The rest of this paper is organized as follows. In Section~\ref{sec:overview}, we give a general overview including the evolution  history of FSL, the notations and definition we will use later, and the proposed taxonomy for existing FSL approaches. The generative model based approaches and the discriminative model based approaches are discussed in detail in Section~\ref{sec:gmb} and Section~\ref{sec:dmb}, respectively. Then, several emerging extensional  topics for FSL are summarized in Section~\ref{sec:extensional}. In Section~\ref{sec:applications}, we extensively investigate the FSL applications in various fields and the benchmark performance of FSL. In Section~\ref{sec:conclusion}, we conclude this survey with a discussion on future directions.

\begin{figure*}[t!]
\centering
\includegraphics[width=1\linewidth]{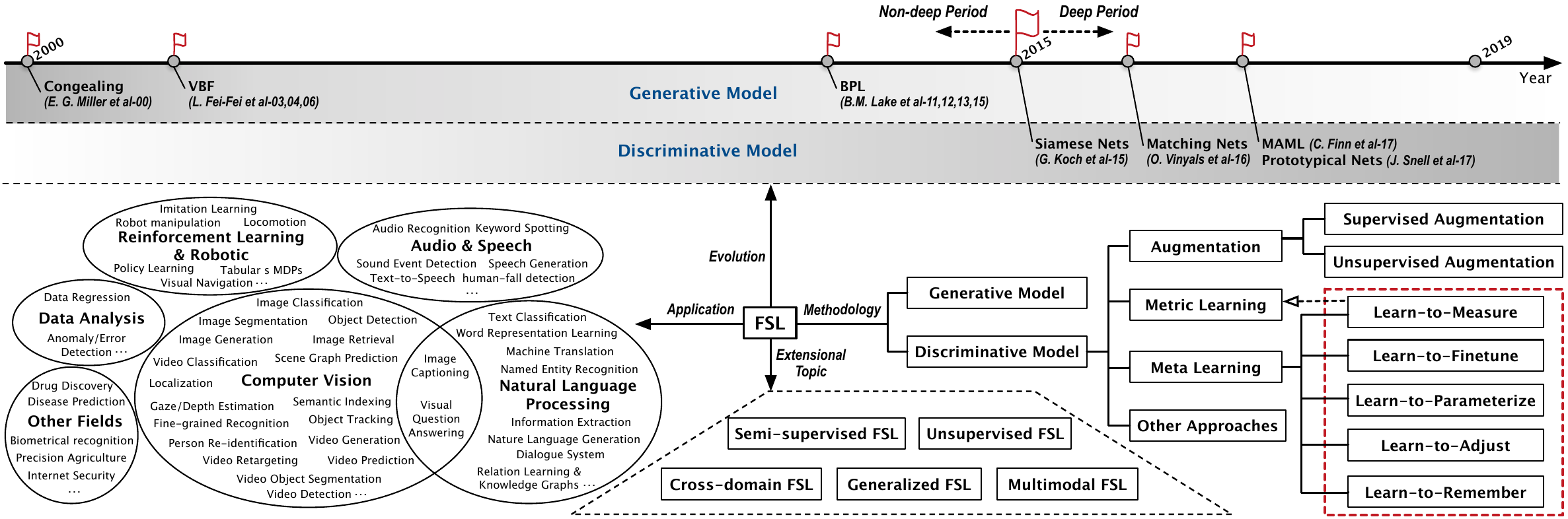}
\vspace{-1.em}
\caption{Outline of our survey. The main contents include the evolution history, methodology, extensional topics, and applications of FSL.}
\label{fig:outline}
\end{figure*}

\section{Overview}
\label{sec:overview}
In this section, we first briefly review the FSL evolution history in Section~\ref{subsec:evolution}. Then, some notations and definitions are introduced in Section~\ref{subsec:notations}. Finally, we provide a high-level taxonomy for existing FSL approaches in Section~\ref{subsec:taxonomy}.

\subsection{Evolution History}
\label{subsec:evolution}
The general regime of machine learning is to make predictions on the future data using the statistical models that are learned on previously prepared training samples. In most cases, the generalization ability of the models is guaranteed by a sufficient quantity of training samples. In many realistic applications, nevertheless, we might be allowed to access only very few training data for novel concepts, in the limit, just one example per concept. For instance, we may need to recognize several kinds of uncommon animals whereas only several annotated pictures are at hand due to their rarity. Similarly, we may be required to authenticate the identity for some new users based on mobile sensor information given a handful of historical usage records from them. The problem of learning from very few examples firstly attracted the attention of E. G. Miller \emph{et al.} in 2000~\cite{miller2000learning}, who postulated a shared density on digit transforms and proposed a Congealing algorithm to bring test digit image into correspondence with class-specific congealed digit image. Thereafter, more and more efforts were devoted to FSL research.


The development process of FSL research can be roughly divided into two periods, non-deep period (from 2000 to 2015) and deep period (from 2015 to now), as depicted in Fig.~\ref{fig:outline}. The watershed separating them is the first combination of deep learning techniques and FSL problems introduced by G. Koch \emph{et al.} in 2015~\cite{koch2015siamese}. Before that, all solutions proposed for FSL problems are based on non-deep learning methodologies or techniques. In particular, most of the famous early FSL approaches in non-deep period are based on the generative model. They seek to estimate the joint distribution $P(\mathcal{X},\mathcal{Y})$ or the conditional distribution $P(\mathcal{X}|\mathcal{Y})$ given a supervision (e.g., a class), albeit on very few observed training samples and then make predictions for test samples from the point of Bayesian decision. 
Several milestones among these generative model based FSL approaches in non-deep period include Congealing algorithm by E. G. Miller \emph{et al.}~\cite{miller2000learning}, Variational Bayesian framework (VBF) by L. Fei-Fei \emph{et al.}~\cite{fe2003bayesian,fei2004learning,fei2006one}, and Bayesian Program Learning (BPL) by B. M. Lake \emph{et al.}~\cite{lake2011one,lake2012concept,lake2013one,lake2015human}. Congealing algorithm~\cite{miller2000learning} is the earliest founder for studying how to learn from very few samples, while VBF~\cite{fe2003bayesian} is the first work to articulate the term of ``one-shot learning''. Comparably, BPL~\cite{lake2015human} reaches a human-level one-shot character classification performance by capitalizing on the human abilities of compositionality, causality and imagination in the cognition of novel concepts. In this non-deep period, there also have several discriminative model based FSL approaches~\cite{fink2005object,wolf2005robust,bart2005cross,tommasi2009more,yu2010attribute,tang2010optimizing}, though they were not the mainstream at this period. Opposite to generative model,  discriminative model based FSL approaches pursue a conditional distribution $P(\mathcal{Y}|\mathcal{X})$ which can directly predict a probability given one observed sample. Despite the efforts above,  the FSL research in the non-deep period still evolves very slowly.

With deep learning booming, especially the great success achieved by CNNs on visual tasks~\cite{krizhevsky2012imagenet,szegedy2015going,he2016deep}, many FSL researchers began to shift their sights from non-deep models to deep models. In 2015, G. Koch \emph{et al.}~\cite{koch2015siamese} took the lead in incorporating deep learning into the solution for FSL issues by proposing a Siamese CNN to learn a class-irrelevant similarity metric on pairwise samples, which marks the beginning of a new era for FSL, i.e., the deep period. After that, the subsequent FSL approaches made full use of the advantages of deep neural networks in  feature representation and end-to-end model optimization to address FSL problems from different angles including data augmentation~\cite{tanner1987calculation}, metric learning~\cite{xing2003distance} and meta learning~\cite{vilalta2002perspective}, etc, pushing FSL researches into a new period of rapid development. Although a few generative model based approaches, such as Neural Statistician~\cite{edwards2016towards} and Sequential Generative Model~\cite{rezende2016one}, were proposed in this deep period, discriminative model based FSL approaches dominate the evolution of FSL study. Especially, a large number of meta learning based FSL approaches have been springing up in recent years,
such as Matching Nets by O. Vinyals \emph{et al.}~\cite{vinyals2016matching}, MAML by C. Finn \emph{et al.}~\cite{finn2017model}, Meta-Learner LSTM by S. Ravi and H. Larochelle~\cite{ravi2016optimization}, MANN by A. Santoro \emph{et al.}~\cite{Santoro2016One}, MetaNet by T. Munkhdalai and H. Yu~\cite{munkhdalai2017meta}, Prototypical Nets by J. Snell \emph{et al.}~\cite{snell2017prototypical}, Relation Net by F. Sung \emph{et al.}~\cite{yang2018learning} and LGM-Nets by H. Li \emph{et al.}~\cite{li2019lgm}, etc. Noticeably, meta learning strategies become the prevailing ideology for FSL. In this period, furthermore, these advanced FSL approaches have been  directly applied to or improved to tackle various applications in computer vision, natural language processing, audio and speech, data analysis, robotics, etc. Meanwhile, more and more challenging extensional topics relating to FSL, such as Semi-supervised FSL, Unsupervised FSL, Cross-domain FSL, Generalized FSL and Multimodal FSL, have been unearthed. 

In brief, the evolution history of FSL witnessed a transition from non-deep period to deep period, an alternation of mainstream approaches between generative model and discriminative model, and a resurgence of the classical meta learning idea. Today, FSL related works frequently appear in many top venues, most notably in machine learning or their applications, attracting wide attention of the machine learning community.

\subsection{Notations and Definitions}
\label{subsec:notations}

Formally, we use $x$ to represent input data, $y$ to represent supervision target,  $\mathcal{X}$ and $\mathcal{Y}$ to denote the space of input data and supervision target, respectively. An FSL task $T$ is described by a $T$-specific dataset $D_{T}=\{D_{\mathrm{trn}}, D_{\mathrm{tst}}\}$  with $D_{\mathrm{trn}} = \{(x_i,y_i)\}_{i=1}^{N_{\mathrm{trn}}}$ and $D_{\mathrm{tst}} = \{x_j\}_{j=1}^{N_{\mathrm{tst}}}$, $x_i, x_j\in \mathcal{X}_T\subset \mathcal{X}, y_i \in \mathcal{Y}_T\subset \mathcal{Y}$. The samples $x_i, x_j$ for task $T$ come from one specific domain $\mathcal{D}_T=\{\mathcal{X}_T, P(\mathcal{X}_T)\}$ consisting of a data space $\mathcal{X}_T$ and a marginal probability distribution $P(\mathcal{X}_T)$.
Usually, there are $C$ task classes and only $K$ (very small, 1, 5, for example) samples per class in $D_{\mathrm{trn}}$, that is, $N_{\mathrm{trn}}=CK$, then $T$ is also called as $C$-way $K$-shot task. The goal is to produce a target predictive function $f \in \mathcal{F}: \mathcal{X}\rightarrow\mathcal{Y}$ which can make predictions for test samples in $D_{\mathrm{tst}}$. Based on our analysis in Section~\ref{sec:introduction}, it is hard to build a high-quality $f$ just with the scarce $D_{\mathrm{trn}}$. In most cases, therefore, researchers are allowed to leverage one supervised auxiliary dataset $D_{A}=\{(x^a_i,y^a_i)\}_{i=1}^{N_{\mathrm{aux}}}$, $x^a_i \in \mathcal{X}_A\subset \mathcal{X}, y^a_i\in \mathcal{Y}_A\subset \mathcal{Y}$, that includes sufficient samples and classes ($N_a \gg N_{\mathrm{trn}}$, $|\mathcal{Y}_A|\gg |\mathcal{Y}_T|$) collected based on previously seen concepts. It needs to be noted that $D_{A}$ does not contain data belonging to classes in $T$, that is, $\mathcal{Y}_T \cap \mathcal{Y}_A = \emptyset$, and the data of $D_{T}$ and those of $D_{A}$ come from the same domain, that is, $\mathcal{D}_T = \mathcal{D}_A$, $\mathcal{X}_T=\mathcal{X}_A$ and $P(\mathcal{X}_T)=P(\mathcal{X}_A)$, where $\mathcal{D}_A=\{\mathcal{X}_A, P(\mathcal{X}_A)\}$. The setting is solid and reasonable since $D_{A}$ is easy to be acquired from many historical, offline or publicly well-labeled data that are relevant to task $T$, especially in today's big data era.

On these basis, we give a unified definition of FSL.\\

\vspace{-0.8em} 
\noindent \textbf{Definition 1} \emph{(Few Sample Learning)} Given a task $T$ described by a $T$-specific dataset $D_{T}$ with only a few supervised information available, and a $T$-irrelevant auxiliary dataset $D_{A}$ (if any), \emph{few sample learning} aims to build a function $f$ for task $T$ that maps its inputs to targets using the very few supervision information in $D_{T}$ and the knowledge in $D_{A}$. \\

\vspace{-0.6em}
The term of $T$-irrelevant in above definition implies the targets in $D_T$ and $D_A$ are orthogonal, that is, $\mathcal{Y}_T \cap \mathcal{Y}_A = \emptyset$.  If $D_A$ covers the classes in $T$, i.e., $\mathcal{Y}_T \cap \mathcal{Y}_A = \mathcal{Y}_T$, the FSL problem will collapse to a traditional large sample learning problem. In particular, if $|\mathcal{Y}_T|=2$, $T$ is a binary FSL task, and if $|\mathcal{Y}_T|>2$,  then we call $T$ is a multiclass FSL task. Besides, we conclude several important extensional topics of FSL in light of the above notations and definition.

\vspace{0.2em}
\textbf{Semi-supervised FSL}. In addition to the $CK$ supervised samples, $D_{\mathrm{trn}}$ also contains some unlabeled training samples. 

\vspace{0.2em}
\textbf{Unsupervised FSL}. $D_A$ is fully unsupervised despite it  contains sufficient samples from non-task classes. 

\vspace{0.2em}
\textbf{Cross-domain FSL}. The samples in $D_T$ and $D_A$ come from two different data domains, that is, $\mathcal{D}_T\neq \mathcal{D}_A$. 

\vspace{0.2em}
\textbf{Generalized FSL}. Function $f$ is required to make inference on united label space $\mathcal{Y}_T \cup \mathcal{Y}_A$ rather than  single $\mathcal{Y}_T$. 

\vspace{0.2em}
\textbf{Multimodal FSL}. It has two cases, multimodal matching and multimodal fusion. In the former case, the target $y_i$ in a labeled sample pair $(x_i, y_i)$ is not the simple class label, but one data in another modality different from the modality of input $x_i$. In the latter case, the additional information $I_i$ for $x_i$ belonging to other modality is provided. 

We give the detailed problem description and literature review for the above five extensional topics in Section~\ref{sec:extensional}.


\subsection{Taxonomy}
\label{subsec:taxonomy}
As shown in Fig.~\ref{fig:outline}, we organize FSL approaches into two major categories, i.e., generative model based approaches and discriminative model based approaches in light of the modeling principles to FSL problems. For one test sample $x_j$, all FSL solutions pursue the following statistical model which can predict the posterior probability of class given $x_j$
\begin{equation}
\hat{y}_j = \mathop{\arg \max}\nolimits_{y\in\mathcal{Y}_T} p(y|x_j),
\label{eq:posterior probability}
\end{equation}
where $\hat{y}_j$ denotes the predicted target by this model. Discriminative model based FSL approaches aim to directly model the posterior probability $p(y|x)$, which takes $x$ as the input of discriminative model and outputs one probability distribution of $x$ belonging to $C$ task classes. By contrast, generative model based approaches tackle it using Bayesian decision $p(y|x) = p(x|y)p(y)/p(x,y)$.Thus the maximization of posterior probability in Eq.~(\ref{eq:posterior probability}) becomes to
\begin{equation}
\hat{y}_j = \mathop{\arg\max}\nolimits_{y\in\mathcal{Y}_T} p(x_j|y)p(y),
\label{eq:prior probability}
\end{equation}
where $p(y)$ is the prior distribution of target class, $p(x_j|y)$ is the conditional distribution of data given class $y$. In most cases, $p(y)$ is assumed to be a uniform distribution among  classes or computed as the frequency ratio of data in different classes. Consequently, the core aim of generative model based FSL approaches is to compare $p(x|y)$, $y\in \mathcal{Y}_T$. 

\begin{table*}[t]
\begin{center}
\caption{Summary of different generative model based FSL approaches}
\vspace{-1em}
\label{tab:gmb_comparison}
\scalebox{0.855}{
\begin{tabular}{lllll}
\toprule[1pt]
\textbf{Approaches} & \textbf{Latent Variable} & \textbf{Task Type} & \textbf{Experimental Dataset} & \textbf{Remark}   \\
\cmidrule[0.5pt](lr){1-5}
\multirow{2}{*}{Congealing~\cite{miller2000learning}}& \multirow{2}{*}{Transformation $\mathbf{z}_{\mathrm{tran}}$} & \multirow{2}{*}{Multi-class image classification} & \multirow{2}{*}{NIST Special Database 19~\cite{grother1995nist}} & the founder of FSL/only applicable to simple \\
& & & &  digit or letter character grayscale images \\
\cmidrule[0.5pt](lr){1-5}
\multirow{2}{*}{VBF~\cite{fe2003bayesian,fei2004learning,fei2006one}} & \multirow{2}{*}{Parameters $\mathbf{z}_{\mathrm{para}}$} & \multirow{2}{*}{Binary image classification} & Caltech 4 Data Set~\cite{weber2000unsupervised,fe2003bayesian}, & the first work to propose ``one-shot learning''/\\
& & & Caltech 101 Data Set~\cite{fei2004learning,fei2006one} & hard to adapt to multi-class tasks\\
\cmidrule[0.5pt](lr){1-5}
\multirow{2}{*}{HB~\cite{salakhutdinov2012one}} & \multirow{2}{*}{Superclass $\mathbf{z}_{\mathrm{sup}}$} & \multirow{2}{*}{Binary image classification} & MNIST~\cite{lecun1998gradient}, & relies on the underlying hierarchical inter-class    \\
& & & MSR Cambridge dataset~\cite{salakhutdinov2012one}& relationship/hard to adapt to multi-class tasks\\
\cmidrule[0.5pt](lr){1-5}
\multirow{2}{*}{BPL~\cite{lake2011one,lake2012concept,lake2013one,lake2015human}} & \multirow{2}{*}{Programs $\mathbf{z}_{\mathrm{prog}}$} & Multi-class image classification, & \multirow{2}{*}{Omniglot~\cite{lake2015human}} & requires the dynamic stroke information and  \\ 
& &Image generation & & the production rules of image objects\\
\cmidrule[0.5pt](lr){1-5}
\multirow{2}{*}{Chopping~\cite{fleuret2006pattern}} & \multirow{2}{*}{Splits $\mathbf{z}_{\mathrm{spl}}$} & \multirow{2}{*}{Binary image classification} & COIL-100 database~\cite{nene1996object}, & like a probabilistic ``ensemble'' method/hard \\
& & & LATEX symbols~\cite{fleuret2006pattern} & to adapt to multi-class tasks\\
\cmidrule[0.5pt](lr){1-5}
\multirow{3}{*}{CPM~\cite{wong2015one}} & \multirow{3}{*}{Reconstruction $\mathbf{z}_{\mathrm{rec}}$} & \multirow{3}{*}{Multi-class image classification} & \multirow{3}{*}{MNIST~\cite{lecun1998gradient}, USPS~\cite{hull1994database}}& only applicable to simple digit or letter \\
& & &  &character grayscale images/does not need the\\
& & & &    auxiliary set $D_A$   \\
\cmidrule[0.5pt](lr){1-5}
\multirow{2}{*}{Neural Statistician~\cite{edwards2017towards}} & \multirow{2}{*}{Statistics $\mathbf{z}_{\mathrm{stat}}$} & Multi-class image classification, & MNIST~\cite{lecun1998gradient}, Omniglot~\cite{lake2015human}, & an extension of a variational autoencoder/ \\
& & Image generation & Youtube Faces database~\cite{wolf2011face}& contains some deep neural networks\\

\bottomrule[1pt]
\end{tabular}}
\end{center}
\end{table*}

For the category of generative model, researchers bridge the connection between $x$ and $y$ using some intermediate latent variables such that the conditional distribution $p(x|y)$ can be computed mathematically. Most of FSL approaches in this category require some necessary assumptions on the distribution of the latent variables. We will briefly discuss the generative model based approaches in Section~\ref{sec:gmb}.

For the category of discriminative model, three mainstreams are summarized, which include augmentation, metric learning and meta learning. The augmentation approaches are further divided into supervised augmentation and unsupervised augmentation according to whether extra supervision information (e.g., attribute annotation, word embedding, etc) has been used. As the most popular treatment to FSL problems recently, the meta learning based approaches involve various views to reach the goal of learn-to-learn. We divide the existing meta learning based FSL approaches into five major genres in light of what is hoped to be meta-learned behind the meta strategy, Learn-to-Measure, Learn-to-Finetune, Learn-to-Parameterize, Learn-to-Adjust and Learn-to-Remember. In a broad sense, Learn-to-Measure approaches fall into the scope of metric learning since they all pursue a metric space rendering homogeneous samples close and inhomogeneous samples far apart. Even so, the important basis by which we assign Learn-to-Measure approaches into meta learning is the use of meta learning strategy. Also, there exist other niche directions to tackle FSL problems. We will review the discriminative model based FSL approaches in Section~\ref{sec:dmb}.

\section{Generative Model based Approaches}~\label{sec:gmb}
As mentioned in Section~\ref{subsec:taxonomy}, the generative model based FSL approaches seek to model the posterior probability $p(x|y)$. In most cases, however, the probabilistic relationship between data $x$ and target $y$ is not straightforward. For instance, in few-shot image classification,  $x$ denotes one image and $y$ denotes its class label, and the mathematical connection between them can not be described directly. A feasible strategy to bridge the connection between $x$ and $y$ is to introduce an intermediate latent variable $\mathbf{z}$ as follows:
\begin{equation}
p(x|y) = \int_{\mathbf{z}}p(x,\mathbf{z}|y)d\mathbf{z}=\int_{\mathbf{z}}p(\mathbf{z}|y)p(x|\mathbf{z},y)d\mathbf{z}.
\label{eq:latent_variable}
\end{equation}
Almost all generative model based FSL approaches follow this high-level strategy, even if they differ in the specific form of $\mathbf{z}$. Several classic forms of $\mathbf{z}$ are summarized as follows.

\begin{itemize}
\item \emph{Transformation}
\end{itemize}

As the first work that attempts to learn from one sample, Congealing~\cite{miller2000learning} algorithm assumes  there exists one latent image for each digit class and all observed images belonging to this class are produced from the latent image through some underlying transformations $\mathbf{z}_{\mathrm{tran}}$. Moreover, the  density over transformations are supposed to be shared across different classes, which implies that the transformation probability is independent of class. Thus, Eq.~(\ref{eq:latent_variable}) can be written into
\begin{equation}
p(x|y) =\int_{\mathbf{z}_{\mathrm{tran}}}p(\mathbf{z}_{\mathrm{tran}})p(x|\mathbf{z}_{\mathrm{tran}},y)d\mathbf{z}_{\mathrm{tran}}.
\label{eq:latent_variable_trans}
\end{equation}
Importantly, $p(\mathbf{z}_{\mathrm{tran}})$ can be learned on auxiliary set $D_{A}$. We must emphasize that Congealing algorithm is only applicable to the simple digit or letter character grayscale images since it is unrealistic
 to model such class-shared transformation mathematically for other natural RGB images.


\begin{itemize}
\item \emph{Parameters}
\end{itemize}

VBF~\cite{fe2003bayesian,fei2004learning,fei2006one}  measures the probability that an object exists in one RGB image using probabilistic models. The probabilistic models involve many parameters $\mathbf{z}_{\mathrm{para}}$ that needs to be learned. Thus, VBF defines the $p(\mathbf{z}_{\mathrm{para}}|y)$ using a so-called constellation model and utilizes variational {}methods to estimate $\mathbf{z}_{\mathrm{para}}$  on auxiliary set $D_{A}$.

\begin{itemize}
\item \emph{Superclass}
\end{itemize}

In~\cite{salakhutdinov2012one}, a hierarchical Bayesian (HB) model was developed by introducing the superclass relationship over classes. Its key insight is that the classes under the same superclass inherit the same similarity metric. By the superclass variable $\mathbf{z}_{\mathrm{sup}}$, Eq.~(\ref{eq:latent_variable}) can be turned into
\begin{equation}
p(x|y) =\sum\nolimits_{\mathbf{z}_{\mathrm{sup}}} p(\mathbf{z}_{\mathrm{sup}}^y)p(x|\mathbf{z}_{\mathrm{sup}}^y),
\label{eq:latent_variable_superclass}
\end{equation}
where $p(\mathbf{z}_{\mathrm{sup}}^y)=p(\mathbf{z}_{\mathrm{sup}}|y)$ is the prior distribution of the superclass that $y$ belongs to, and $p(x|\mathbf{z}_{\mathrm{sup}}^y)=p(x|\mathbf{z}_{\mathrm{sup}},y)$ is the data distribution conditioned on the superclass $\mathbf{z}_{\mathrm{sup}}^y$.

\begin{itemize}
\item \emph{Programs}
\end{itemize}

BPL~\cite{lake2011one,lake2012concept,lake2013one,lake2015human} uses a Bayesian process to model the generation process of character objects as a probabilistic program. This program will experience a bottom-up parsing analysis of primitives, sub-parts, parts, types, tokens and images. Furthermore, the intermediate types and tokens within the generation program are treated as the latent variable $\mathbf{z}_{\mathrm{prog}}$. With the explicit probabilistic program for each character concept, BPL is able to access the compositionality and causality of character objects and can perform one-shot classification, generate new exemplars given one sample, and generate new character classes as well. 

\begin{itemize}
\item \emph{Splits}
\end{itemize}

Chopping model~\cite{fleuret2006pattern} introduces the random data splits of the auxiliary set $D_A$ as the latent variable $\mathbf{z}_{\mathrm{spl}}$ to bridge the mathematical dependence between raw image $x$ and the label $y$. It makes many splits on $D_A$ by assigning label 1 to half of auxiliary classes and 0 to the others, and then trains a predictor for each split. For one image in $D_T$, Chopping model will combine the predictions from all split-specific predictors to achieve the Bayesian posterior decision.

\begin{itemize}
\item \emph{Reconstruction}
\end{itemize}

Unlike BPL~\cite{lake2011one,lake2012concept,lake2013one,lake2015human}, a compositional patch model (CPM) that does not rely on the knowledge of dynamic strokes in character images is proposed in~\cite{wong2015one}. Similar to BPL, the core assumption of this model is that the congener character images share the same patch-based structure. Thus, this model first segments the single sample in $D_{\mathrm{trn}}$ for each class into a set of components, and then utilizes an AND-OR graph to reconstruct the test sample in $D_{\mathrm{tst}}$. The reconstruction is essentially the latent variable $\mathbf{z}_{\mathrm{rec}}$, which is used to make the final one-shot classification for test samples.

\begin{itemize}
\item \emph{Statistics}
\end{itemize}

Neural Statistician model~\cite{edwards2017towards} deploys a deep network to produce statistics that encapsulate a generative model for each $D_{\mathrm{trn}}$. Concretely, the statistics are described by a mean and variance specifying a Gaussian distribution in the latent space. Using the latent variable $\mathbf{z}_{\mathrm{stat}}$, Neural Statistician can realize one-shot generation and classification. 

Table~\ref{tab:gmb_comparison} presents an intuitive comparison between the above mainstream generative model based FSL approaches that constructed the latent variable $\mathbf{z}$ from different perspectives. Except for the Neural Statistician, the remainders were born in the non-deep period of FSL development process, and most of them are tailored in light of the specific task form or data form, lacking the scalability to more general cases. Besides, these early works were validated on various experimental datasets with different evaluation settings, having not formed some comparable benchmarks for subsequent FSL researches at that time. 


\section{Discriminative Model based Approaches}~\label{sec:dmb}
Unlike the above FSL approaches based on generative model, the discriminative model based FSL approaches attempt to model the posterior probability $p(y|x)$ directly for task $T$ using the scarce training set $D_{\mathrm{trn}}$. The computation model for $p(y|x)$ generally contains a feature extractor and a predictor. For few-shot image recognition tasks, for example, the feature extractor and the predictor respectively might be a CNN and softmax layer. Due to the sample scarcity in $D_{\mathrm{trn}}$, it would be easy to trap into overfitting when fitting $p(y|x)$ only with $D_{\mathrm{trn}}$. Therefore, existing discriminative model based FSL approaches pursue the construction of $p(y|x)$ from different perspectives. We summarize them into the following classes. 

The first one is based on augmentation, which advocates learning a general augmentation function $\mathcal{A}(\cdot)$ from the auxiliary dataset $D_A$ to augment the samples or the features of samples in $D_{\mathrm{trn}}$. The augmentation based FSL approaches are reviewed in Section~\ref{sec:dmb:aug}. The second one is based on metric learning, which aims to learn a pairwise similarity metric $\mathcal{S}(\cdot,\cdot)$ on $D_A$. By this metric, a nearest-neighbor classifier can be used for final prediction. The metric learning based FSL approaches are introduced in Section~\ref{sec:dmb:metric}. The third is based on meta learning, which leverages $D_A$ to construct many tasks similar to the task $T$ and adopts the cross-task training strategy to distill some transferrable models, algorithms or parameters. The meta learning based FSL approaches are detailed in Section~\ref{sec:dmb:meta}. Besides, there also exist some other FSL approaches, which  are discussed in Section~\ref{sec:dmb:other}.

\subsection{Augmentation}~\label{sec:dmb:aug}
\begin{figure}[h!]
\centering
\includegraphics[width=1\linewidth]{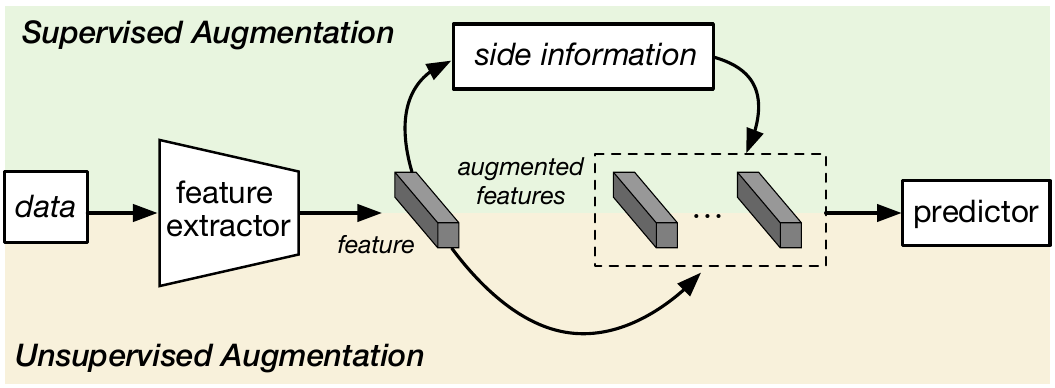}
\vspace{-1.5em}
\caption{General framework of augmentation based FSL approaches.}
\label{fig:augmentation}
\end{figure}

Augmentation is an intuitive way to increase the number of training samples and enhance data diversity. In the field of vision, some basic augmentation operations include rotating, flipping, cropping, translation, and adding noise into images~\cite{krizhevsky2012imagenet,chatfield2014return,zeiler2014visualizing}. For FSL tasks, these low-level augmentation means are insufficient to bring essential gains in the generalization ability of FSL models. In this context, more sophisticated augmentation models, algorithms or networks customized for FSL were proposed, and they mainly occurred in the deep period. Fig.~\ref{fig:augmentation} illustrates the general framework of augmentation based FSL approaches. Except for DAGAN~\cite{antoniou2018data} that augments the samples in $D_{\mathrm{trn}}$ at the data level,  other approaches achieve the feature-level augmentation for training samples in task $T$. According to whether their augmentation relies on external side information (such as semantic attributes~\cite{lampert2013attribute}, word vectors~\cite{mikolov2013distributed}, etc), we further divide the existing augmentation based FSL approaches into supervised and unsupervised ones.

\begin{figure}[t!]
\centering
\includegraphics[width=1\linewidth]{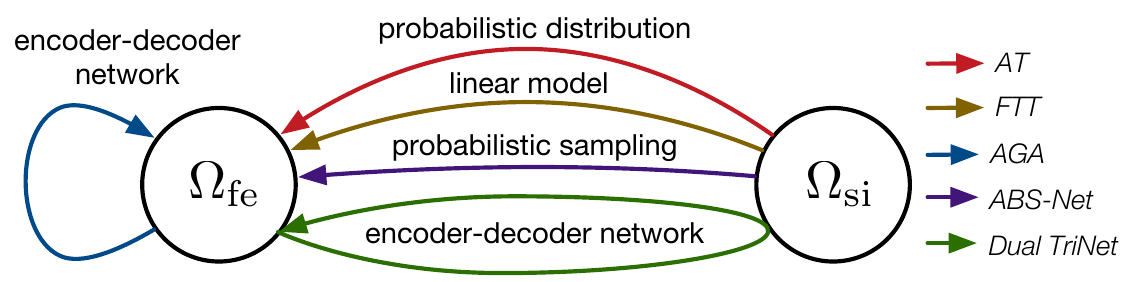}
\vspace{-2em}
\caption{Mapping relationship between the feature and side information in terms of different supervised augmentation approaches.}
\label{fig:supervised_augmentation}
\end{figure}

\subsubsection{Supervised Augmentation}
Several FSL approaches based on supervised augmentation include Feature Trajectory Transfer (FTT)~\cite{kwitt2016one}, AGA~\cite{dixit2017aga}, Dual TriNet~\cite{chen2018semantic,chen2019multi}, Author-Topic (AT)~\cite{yu2010attribute} and ABS-Net~\cite{lu2018attribute}. For ease of notation, let $\Omega_{\mathrm{fe}}$ be the feature space, and $\Omega_{\mathrm{si}}$ be the side information space. The augmentation $\mathcal{A}(\cdot)$ learned by these approaches, essentially, is a mapping relationship between $\Omega_{\mathrm{fe}}$ and $\Omega_{\mathrm{si}}$, although they differ in the mapping direction and mapping module, as shown in Fig.~\ref{fig:supervised_augmentation}.

\begin{table*}[t]
\begin{center}
\caption{Summary of supervised (top part) or unsupervised (bottom part) augmentation based FSL approaches}
\vspace{-1em}
\label{tab:supervised_aug_comparison}
\scalebox{0.85}{
\begin{tabular}{llllll}
\toprule[1pt]
\textbf{Approaches} &\textbf{Side Information} & \textbf{Mapping Direction} & \textbf{Mapping Module} & \textbf{Task Type} & \textbf{Experimental Dataset}   \\
\cmidrule[0.5pt](lr){1-6}
\multirow{2}{*}{FTT~\cite{kwitt2016one}} & transient attributes & \multirow{2}{*}{$\Omega_{\mathrm{si}}$$\rightarrow$$\Omega_{\mathrm{fe}}$} & \multirow{2}{*}{linear model} & scene location & Transient Attributes Database (TADB)~\cite{laffont2014transient}\\
& (\textit{rainy, sunny, etc}) & & & classification &SUN Attributes Database (SADB)~\cite{patterson2014sun}\\
\cmidrule[0.5pt](lr){1-6}
\multirow{2}{*}{AGA~\cite{dixit2017aga}}& attribute strength  & \multirow{2}{*}{$\Omega_{\mathrm{fe}}$$\stackrel{\Omega_{\mathrm{si}}}{\longrightarrow}$$\Omega_{\mathrm{fe}}$} & encoder-decoder & 2D/3D object & \multirow{2}{*}{SUN RGB-D~\cite{song2015sun}} \\
& (\textit{depth, pose}) & & network (MLP) &classification \\
\cmidrule[0.5pt](lr){1-6}
\multirow{2}{*}{AT~\cite{yu2010attribute}}& discrete attributes & \multirow{2}{*}{$\Omega_{\mathrm{si}}$$\rightarrow$$\Omega_{\mathrm{fe}}$} & \multirow{2}{*}{probabilisic distribution} & \multirow{2}{*}{image classification} & \multirow{2}{*}{Animals with Attributes (AwA)~\cite{lampert2009learning}} \\
& (\textit{black, fierce, etc}) &  \\
\cmidrule[0.5pt](lr){1-6}
\multirow{2}{*}{Dual TriNet~\cite{chen2018semantic,chen2019multi}}& word vectors,  & \multirow{2}{*}{$\Omega_{\mathrm{fe}}$$\rightarrow$$\Omega_{\mathrm{si}}$$\rightarrow$$\Omega_{\mathrm{fe}}$} & encoder-decoder & \multirow{2}{*}{image classification} & \emph{mini}ImageNet~\cite{vinyals2016matching}, Cifar-100~\cite{krizhevsky2009learning}, \\
& discrete attributes & & network (CNN)& & CUB~\cite{wah2011caltech}, Caltech-256~\cite{griffin2007caltech} \\
\cmidrule[0.5pt](lr){1-6}
\multirow{2}{*}{ABS-Net~\cite{lu2018attribute}}& discrete attributes  & \multirow{2}{*}{$\Omega_{\mathrm{si}}$$\rightarrow$$\Omega_{\mathrm{fe}}$} & \multirow{2}{*}{probabilisic sampling} & \multirow{2}{*}{image classification} & \multirow{2}{*}{Colored MNIST~\cite{lu2018attribute}} \\
& (\textit{ForColor, BackColor}) &  \\

\bottomrule[0.5pt]
\toprule[0.5pt]
\multirow{2}{*}{GentleBoostKO~\cite{wolf2005robust}}& \multirow{2}{*}{--}  & \multirow{2}{*}{$\Omega_{\mathrm{fe}}$$\rightarrow$$\Omega_{\mathrm{fe}}$} &  knockout (feature & binary image & \multirow{2}{*}{Caltech datasets~\cite{fergus2003object}} \\
&  & & element replacement)&  classification& \\

\cmidrule[0.5pt](lr){1-6}
\multirow{2}{*}{SH~\cite{hariharan2017low}}& \multirow{2}{*}{--}  & \multirow{2}{*}{$\Omega_{\mathrm{fe}}$$\rightarrow$$\Omega_{\mathrm{fe}}$} &  quadruplet-based MLP & \multirow{2}{*}{image classification} & \multirow{2}{*}{ImageNet1k~\cite{russakovsky2015imagenet}} \\
&  & & (3 features $\rightarrow$ 1 feature)& & \\

\cmidrule[0.5pt](lr){1-6}
\multirow{2}{*}{Hallucinator~\cite{wang2018low}}& \multirow{2}{*}{--}  & \multirow{2}{*}{$\Omega_{\mathrm{fe}}$$\rightarrow$$\Omega_{\mathrm{fe}}$} &  MLP-based generator  & \multirow{2}{*}{image classification} & \multirow{2}{*}{ImageNet1k~\cite{russakovsky2015imagenet}} \\
&  & & (1 features $\rightarrow$ 1 feature)& & \\

\cmidrule[0.5pt](lr){1-6}
CP-ANN~\cite{gao2018low} &- & latent space$\rightarrow$$\Omega_{\mathrm{fe}}$ &  GAN  & image classification & ImageNet1k~\cite{russakovsky2015imagenet} \\

\cmidrule[0.5pt](lr){1-6}
\multirow{3}{*}{$\Delta$-encoder~\cite{schwartz2018delta}}& \multirow{3}{*}{--}  & \multirow{3}{*}{$\Omega_{\mathrm{fe}}$$\rightarrow$$\Omega_{\mathrm{fe}}$} &  encoder-decoder  & \multirow{3}{*}{image classification} & \emph{mini}ImageNet~\cite{vinyals2016matching}, Cifar-100~\cite{krizhevsky2009learning}, \\
&  & & network (MLP)& &CUB~\cite{wah2011caltech}, Caltech-256~\cite{griffin2007caltech},  \\
&  & & (3 features $\rightarrow$ 1 feature) & & AwA~\cite{lampert2009learning}, aPascal\&aYahoo (APY)~\cite{farhadi2009describing}  \\

\cmidrule[0.5pt](lr){1-6}
\multirow{2}{*}{DAGAN~\cite{antoniou2018data}}& \multirow{2}{*}{--}  & \multirow{2}{*}{$\Omega_{\mathrm{da}}$$\rightarrow$$\Omega_{\mathrm{da}}$} &  \multirow{2}{*}{GAN}  & image generation, & Omniglot~\cite{lake2015human}, EMNIST~\cite{cohen2017emnist} \\
&  & & &image classification & VGG-Faces~\cite{parkhi2015deep}\\

\cmidrule[0.5pt](lr){1-6}
\multirow{2}{*}{IDeMe-Net~\cite{chen2019image}}& \multirow{2}{*}{--}  & \multirow{2}{*}{$\Omega_{\mathrm{da}}$$\rightarrow$$\Omega_{\mathrm{da}}$} &  Deformation Sub-network & \multirow{2}{*}{image classification} & \multirow{2}{*}{ImageNet1k~\cite{russakovsky2015imagenet}, \emph{mini}ImageNet~\cite{vinyals2016matching}}\\
&  & & (2 images $\rightarrow$ 1 images)& & \\

\bottomrule[1pt]

\end{tabular}}
\end{center}
\footnotesize{Note: the term in the colume of ``Task Type'' without ``binary'' all indicates multi-class classification.}
\end{table*}

\begin{figure}[b!]
\centering
\includegraphics[width=0.85\linewidth]{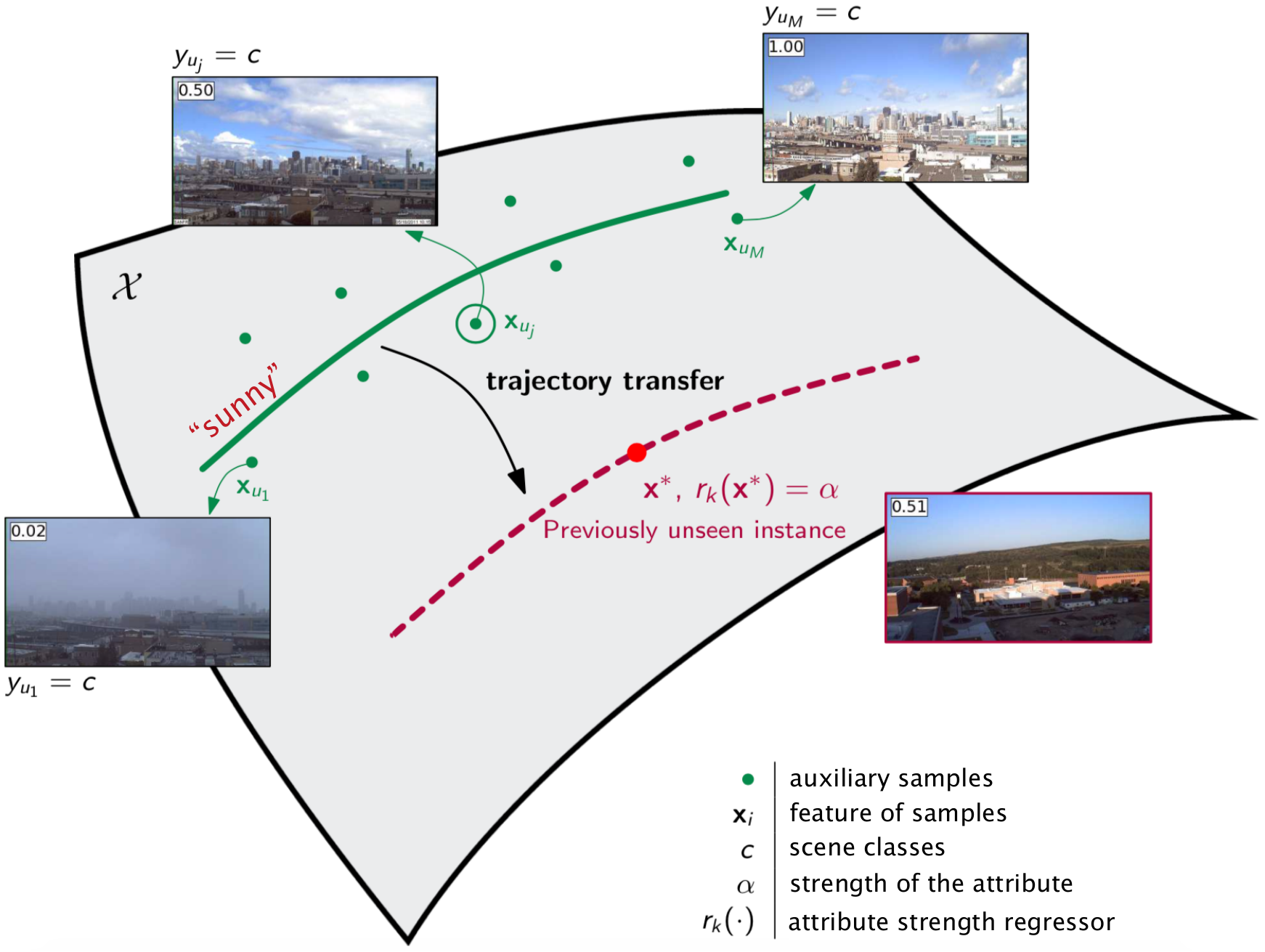}
\vspace{-1em}
\caption{Illustration of Feature Trajectory Transfer (FTT)~\cite{kwitt2016one}. Number at 
upper right corner in each scene image describes the strength of ``sunny''. }
\label{fig:FFT}
\end{figure}

FFT~\cite{kwitt2016one} focuses on one-shot scene image classification, which leverages the consecutive attributes in scene images (e.g., ``rainy'', ``dark'' or ``sunny'') to directionally synthesize the features for the one-sample task scene class. In particular, FFT suggests to learn a linear mapping trajectory on auxiliary scene classes that maps attribute $a \in \mathbb{R}_+$ to feature $\mathbf{x} \in \mathbb{R}^d$:
\begin{equation}
\mathbf{x} = \mathbf{w} \cdot a + \mathbf{b} + \mathbf{\epsilon},
\label{eq:fft}
\end{equation}
where $\mathbf{w}, \mathbf{b} \in \mathbb{R}^d$ are learnable parameters and $\mathbf{\epsilon}$ denotes Gaussian noise. This mapping trajectory is expected to be transferrable from auxiliary  classes to task classes. As shown in Fig.~\ref{fig:FFT}, given only one training sample for a task scene class, one can artificially set the strength of its attribute (e.g., the degree of \emph{sunny}) to produce many synthetic features by the well-learned linear mapping trajectory in Eq.~(\ref{eq:fft}). However, we must note that FFT requires the fine-grained and consecutive attribute annotation, which is a prohibitive cost for data preparation.

\begin{figure}[b!]
\centering
\includegraphics[width=0.9\linewidth]{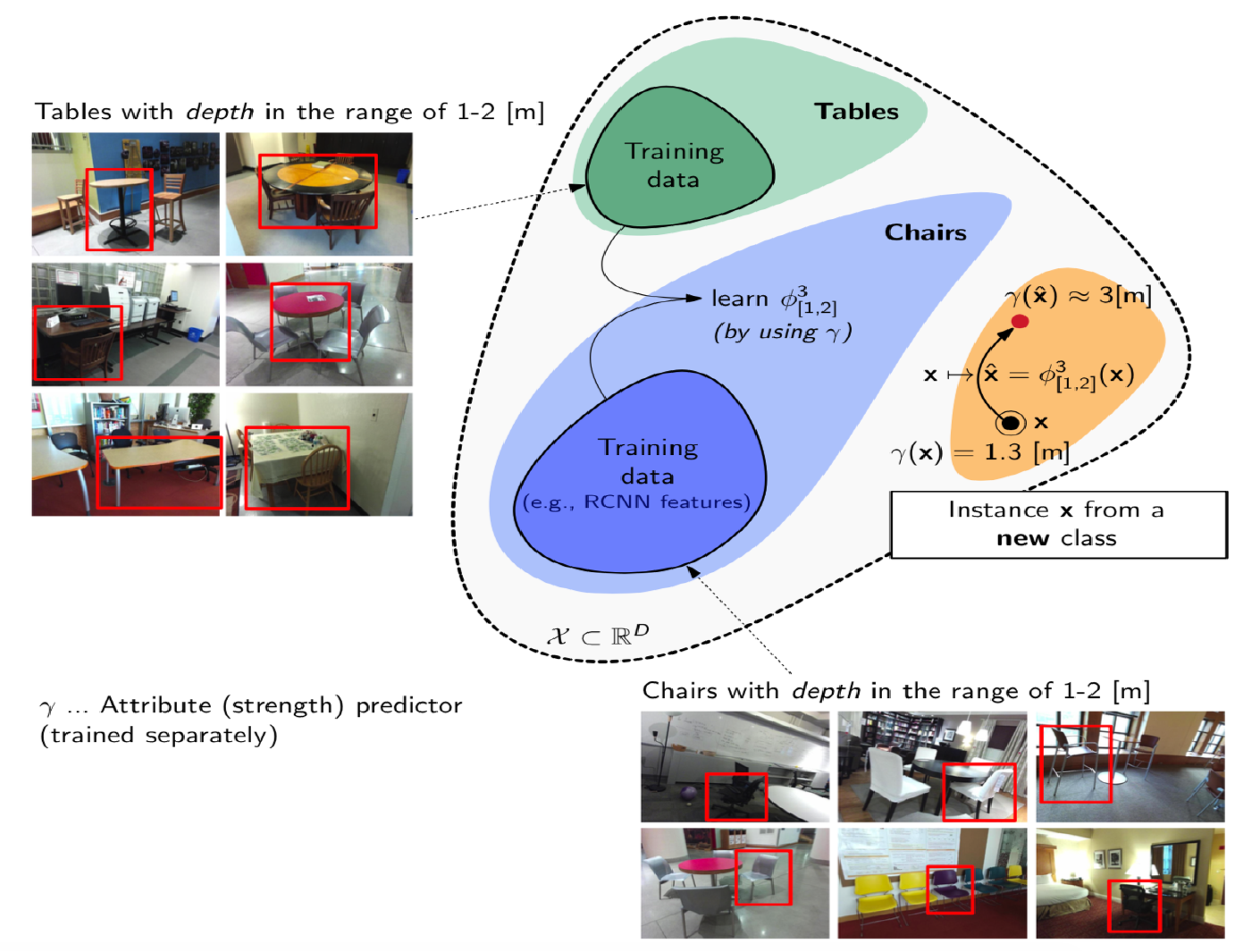}
\vspace{-0.5em}
\caption{Illustration of Attribute-Guided Augmentation (AGA)~\cite{dixit2017aga}. \emph{Depth} is the attribute. Tables and Chairs are two auxiliary classes. $\phi_{[1,2]}^3$ is an encoder-decoder network that transfers the feature of an object with depth in the range of 1-2 [m] into another feature with depth of 3 [m]. }
\label{fig:AGA}
\end{figure}

Comparably, AGA~\cite{dixit2017aga} develops an encoder-decoder network to map the feature of a sample into another synthetic feature at a different attribute strength with the input feature. For example, as shown in Fig.~\ref{fig:AGA}, AGA aims to learn a class-agnostic feature transfer module $\phi_{[1,2]}^{3}$ on auxiliary classes (e.g., Tables, Chairs) that takes the features of objects with depth in the range of 1-2 [m] as inputs and outputs their synthetic features with the depth of 3 [m]. Using this feature transfer module, one can augment the single sample of one task class with various depth strength.
This idea seems similar to FTT~\cite{kwitt2016one}, but two different points exist between them. First, the mapping direction of AGA is $\Omega_{\mathrm{fe}}$$\rightarrow$$\Omega_{\mathrm{fe}}$, while that of FTT is $\Omega_{\mathrm{si}}$$\rightarrow$$\Omega_{\mathrm{fe}}$. Second, the feature synthesis in FTT is guided by directly allocating the desired attribute strength into its linear mapping model, but that in AGA is achieved by the encoder-decoder network specializing in the feature mapping between two explicit attribute strength.

Similarly, Dual TriNet~\cite{chen2018semantic,chen2019multi} also utilizes an encoder-decoder network to achieve the feature-level augmentation. Apart from the difference in the architecture of encoder-decoder network used by them (Dual TriNet uses CNN, while AGA uses MLP), another major difference is the bottleneck embedding between the encoder and decoder: the bottleneck embedding of Dual TriNet is a semantic attribute or word vector, while that of AGA is  trivial latent embedding. Dual TriNet models the bottleneck embedding space as a Semantic Gaussian or a Semantic Neighbourhood, in which  large amounts of semantic vectors can be sampled to be decoded into synthetic features. From this point of view, the mapping direction of Dual TriNet is $\Omega_{\mathrm{fe}}$$\rightarrow$$\Omega_{\mathrm{si}}$$\rightarrow$$\Omega_{\mathrm{fe}}$. 

AT~\cite{yu2010attribute}  uses a topic model~\cite{rosen2010learning} to model the relationship between images and attributes, where each image is treated as a document containing a mixture of topics (i.e., attributes), and each topic is represented by a probabilistic distribution of words (i.e., features). The parameters of this probabilistic distribution are estimated on the auxiliary dataset $D_A$. By the explicit distribution, a large amount of features of a specific class can be generated given the attributes of this class. 

ABS-Net~\cite{lu2018attribute} first conducts an attribute learning process on the auxiliary dataset $D_A$, which allows the establishment of a repository of attribute features. Given the attribution description of one class, a probabilistic sampling operation is performed on the repository, which maps the attributes to the pseudo features of this class.

The top part of Table~\ref{tab:supervised_aug_comparison} summarizes the main characteristics of these supervised augmentation based FSL approaches.
Considering the labeling cost of side information, the above approaches are more suitable for the task or dataset containing some side information.


\subsubsection{Unsupervised Augmentation}
Typical unsupervised augmentation based FSL approaches include GentleBoostKO~\cite{wolf2005robust}, Shrinking and Hallucinating (SH)~\cite{hariharan2017low}, Hallucinator~\cite{wang2018low}, CP-ANN~\cite{gao2018low}, $\Delta$-encoder~\cite{schwartz2018delta}, DAGAN~\cite{antoniou2018data} and IDeMe-Net~\cite{chen2019image} etc, which seek to augment the data or features without any external side information.



GentleBoostKO~\cite{wolf2005robust} is a straightforward FSL solution in the early non-deep period that synthesizes features by a knockout procedure. The knockout is realized by replacing one element of a feature with an element of another feature in the same coordinate. Its key insight is to create corrupted copies of the very few samples to increase the robustness.

\begin{figure}[t!]
\centering
\includegraphics[width=0.92\linewidth]{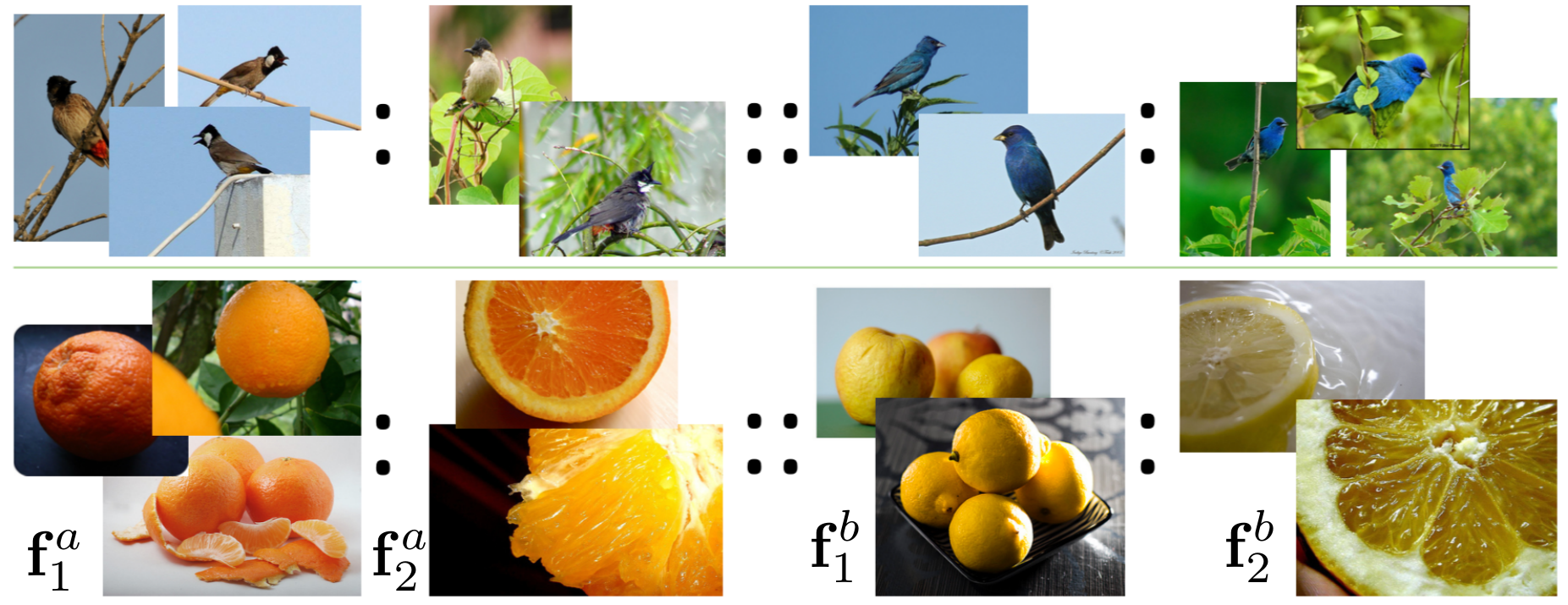}
\vspace{-0.5em}
\caption{Illustration of transformation analogies in the form of quadruplets used by SH~\cite{hariharan2017low}. Top row: birds with sky background \emph{versus} birds with greenery background. Bottom row: whole fruits \emph{versus} cut fruit.}
\label{fig:SH}
\end{figure}

SH~\cite{hariharan2017low} was built on the motivation that the  intra-class variation can generalize across classes (e.g., pose transformations), which is similar to the intuition of FTT~\cite{kwitt2016one} and AGA~\cite{dixit2017aga}. The difference among them is that the intra-class variation in FTT~\cite{kwitt2016one} and AGA~\cite{dixit2017aga}  can be explicitly described by the side information (e.g., the strength of \textit{sunny} attribute, the \textit{depth} value of object), while the underlying intra-class variation in SH needs to be mined from implicit transformation analogies in the form of quadruplets $(\mathbf{f}_1^{a}, \mathbf{f}_2^{a}, \mathbf{f}_1^{b}, \mathbf{f}_2^{b})$, as shown in Fig.~\ref{fig:SH}, where $a, b$ denote two classes. These quadruplets are mined from the auxiliary set by an unsupervised clustering and many heuristic steps. Furthermore,  an MLP-based mapping module $G$ can be learned based on these quadruplets, which takes three features as inputs and outputs a synthetic feature, i.e., $\hat{\mathbf{f}}_2^b$$ = $$G(\mathbf{f}_1^{a}, \mathbf{f}_2^{a}, \mathbf{f}_1^{b})$. Given only one training sample for a task class, one can deduce analogically other synthetic features for this class by $G$.

\begin{figure}[b!]
\centering
\includegraphics[width=0.95\linewidth]{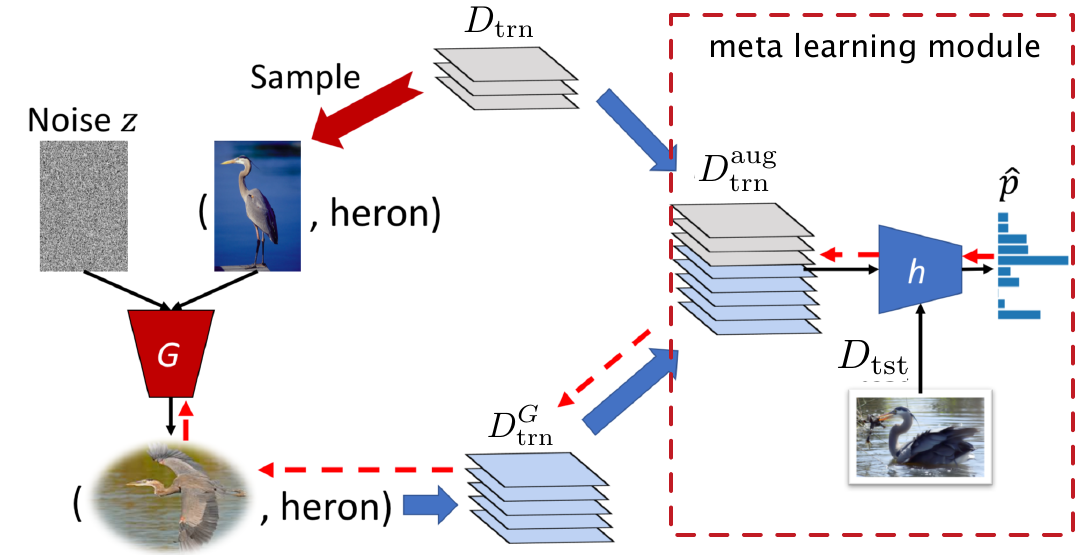}
\vspace{-0.5em}
\caption{Framework of Hallucinator~\cite{wang2018low}. $D_{\mathrm{trn}}^G$: the augmented sample set.}
\label{fig:Hallucinator}
\end{figure}

The high-level motivation of  $\Delta$-encoder~\cite{schwartz2018delta} is similar to that of AGA~\cite{dixit2017aga}, FTT~\cite{kwitt2016one} and SH~\cite{hariharan2017low}. It also suggests to extract transferrable intra-class variation (called $\Delta$) from auxiliary set $D_A$ and apply this variation to the novel task classes so as to synthesize new samples for the task classes. Similar to SH~\cite{hariharan2017low}, $\Delta$-encoder also transfers $\Delta$ based on the underlying quadruplet analogy, and the main difference is the specific  mapping module that deals with the quadruplet relationship:  SH~\cite{hariharan2017low} uses a trivial MLP but $\Delta$-encoder develops an encoder-decoder network whose bottleneck embedding is expected to capture the intra-class variation $\Delta$.

As shown in Fig.~\ref{fig:Hallucinator}, Hallucinator~\cite{wang2018low} uses an MLP-based generator $G$ to augment features for the training samples in $D_{\mathrm{trn}}$, i.e., $\hat{\mathbf{f}} = G(\mathbf{f}, \mathbf{z})$, where $\mathbf{f}$ is an original feature and $\mathbf{z}$ is a noise vector. This generator was devised into a plug-and-play module that can be incorporated into a variety of ready-made meta learning modules, such as Matching Nets~\cite{vinyals2016matching}, Prototypical Nets~\cite{snell2017prototypical} or Prototype Matching Nets~\cite{wang2018low}. The meta learning FSL approaches will be reviewed in Section~\ref{sec:dmb:meta}.

CP-ANN~\cite{gao2018low} achieved feature augmentation for the few support samples via a Generative Adversarial Networks (GAN)~\cite{goodfellow2014generative} based set-to-set translation, which was designed to preserve the covariance of auxiliary samples during augmentation. DAGAN~\cite{antoniou2018data} takes the samples in $D_{\mathrm{trn}}$ as input and generates the within-class data directly (i.e., $\Omega_{\mathrm{da}}$$\rightarrow$$\Omega_{\mathrm{da}}$) by a conditional GAN. Z. Chen \emph{et al.}~\cite{chen2019image} insisted the visual fusion between two similar images may maintain critical semantic information and contribute to formulating the decision boundaries of the final classifier, and thus they proposed IDeMe-Net to generate the deformed images for the small amounts of support samples. 
Similar to Hallucinator~\cite{wang2018low}, both DAGAN~\cite{antoniou2018data} and IDeMe-Net~\cite{chen2019image} were designed to work in coordination with other ready-made meta learning based FSL approaches like Matching Nets~\cite{vinyals2016matching} and Prototypical Nets~\cite{snell2017prototypical}. A high-level summary for the above unsupervised augmentation based FSL approaches is made in the bottom part of Table~\ref{tab:supervised_aug_comparison}.


\begin{table*}[t!]
\begin{center}
\caption{Summary of metric learning based FSL approaches}
\vspace{-1em}
\label{tab:metric_comparison}
\scalebox{0.85}{
\begin{tabular}{lllll}
\toprule[1pt]
\textbf{Approaches} &\textbf{Similarity Metric $\mathcal{S}(\cdot,\cdot)$} &  \textbf{Metric Loss} & \textbf{Task Type} & \textbf{Experimental Dataset}   \\
\cmidrule[0.5pt](lr){1-5}
CRM~\cite{fink2005object}& $d(x_i,x_j)$ (Mahalanobis distance)  & hinge loss & image classification & Latin Character database~\cite{fink2005object} \\
\cmidrule[0.5pt](lr){1-5}
KernelBoost~\cite{hertz2006learning} &$K(x_i,x_j)$ (kernel function) & exponential loss & image classification, image retrieval  & UIC~\cite{blake1998uci}, MNIST~\cite{lecun1998gradient}, YaleB~\cite{georghiades2000few} \\
\cmidrule[0.5pt](lr){1-5}
Siamese Nets~\cite{koch2015siamese} &$\mathbf{p}(x_i,x_j)$ (siamese CNN) & binary cross-entropy loss & image classification  & Omniglot~\cite{lake2015human} \\
\cmidrule[0.5pt](lr){1-5}
Triplet Ranking Nets~\cite{ye2018deep} & $d(x_i,x_j)$ (Euclidean distance) & triple ranking loss & image classification &Omniglot~\cite{lake2015human}, \textit{mini}ImageNet~\cite{vinyals2016matching}\\
\cmidrule[0.5pt](lr){1-5}
SRPN~\cite{mehrotra2017generative} &$\mathbf{p}(x_i,x_j)$ (GAN+siamese CNN) & adversarial loss & image classification  & Omniglot~\cite{lake2015human}, \textit{mini}ImageNet~\cite{vinyals2016matching} \\
\cmidrule[0.5pt](lr){1-5}
MM~\cite{kaiser2017learning} &$d(x_i,x_j)$ (memory+dot product) & memory loss & image classification, translation  & Omniglot~\cite{lake2015human}, WMT14~\cite{kaiser2017learning} \\
\cmidrule[0.5pt](lr){1-5}
\multirow{2}{*}{AdaptHistLoss~\cite{scott2018adapted}} &\multirow{2}{*}{$d(x_i,x_j)$ (cosine distance)} & \multirow{2}{*}{histogram loss} & \multirow{2}{*}{image classification, translation}  & MNIST~\cite{lecun1998gradient}, Isolet of  UIC~\cite{blake1998uci},\\
 & & &&Omniglot~\cite{lake2015human}, tinyImageNet~\cite{scott2018adapted} \\
\bottomrule[1pt]
\end{tabular}}
\end{center}
\end{table*}


\subsubsection{Discussion}
We must emphasize that augmentation based FSL approaches do not conflict with  other FSL approaches, such as those based on metric learning or meta learning to be discussed in Section~\ref{sec:dmb:metric} and~\ref{sec:dmb:meta}. On the contrary, most of the augmentation based FSL approaches are complementary to them and they can be used as the plug-and-play module: one can first adopt these augmentation strategies to enrich $D_{\mathrm{trn}}$ and then learn on the augmented $D_{\mathrm{trn}}^{\mathrm{aug}}$ through other FSL approaches. 

\begin{figure}[b!]
\centering
\includegraphics[width=0.95\linewidth]{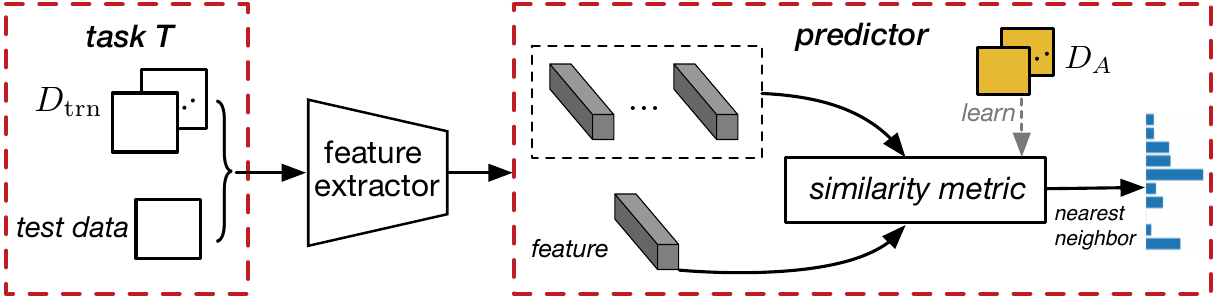}
\vspace{-0.5em}
\caption{General framework of metric learning based FSL approaches.}
\label{fig:metric_learning}
\end{figure}

\subsection{Metric Learning}~\label{sec:dmb:metric}
The general objective of metric learning~\cite{xing2003distance} is to learn a pairwise similarity metric $\mathcal{S}(\cdot,\cdot)$ under which a similar sample pair can obtain a high similarity score while the dissimilar pair gets a low similarity score. All FSL approaches based on metric learning adhere to this principle, as shown in Fig.~\ref{fig:metric_learning}, which create the similarity metric using auxiliary dataset $D_A$ and generalize it to the novel classes of task $T$. The similarity metric could be a simple distance measurement,  a sophisticated network or other feasible modules or algorithms as long as they can estimate the pairwise similarity between samples or features. Several representative metric learning based FSL approaches include Class Relevance Metrics (CRM)~\cite{fink2005object}, KernelBoost~\cite{hertz2006learning}, Siamese Nets~\cite{koch2015siamese}, Triplet Ranking Nets~\cite{ye2018deep}, Skip Residual Pairwise Net (SRPN)~\cite{mehrotra2017generative}, Memory Module (MM)~\cite{kaiser2017learning} and  AdaptHistLoss~\cite{scott2018adapted}. They developed various forms of similarity metrics associated with different metric loss functions to address FSL tasks.


CRM~\cite{fink2005object} is a foundation work of metric learning based FSL approach proposed in the non-deep period. It uses the Mahalanobis distance to measure the pairwise similarity:
\begin{equation}
 d(x_i,x_j)=\sqrt{(x_i-x_j)^{\top}A(x_i-x_j)} = \big\|Wx_i-Wx_j\big\|_2,
\end{equation}
where $A=W^\top W$ is a symmetric positive semi-definite matrix that needs to be learned from  auxiliary dataset $D_A$. The learning objective of CRM follows the form of hinge loss to make the distance of positive sample pair $(x_i,x_i^+)$ smaller than that of negative sample pair $(x_j,x_j^-)$ by $\gamma$ at least:
\begin{equation}
 d(x_i,x_i^+)\leq d(x_j,x_j^-) - \gamma.
\end{equation}
Once trained on $D_A$, the Mahalanobis distance is applied to the task $T$ to enable the nearest neighbor (NN) classification.


KernelBoost~\cite{hertz2006learning} suggests to learn the pairwise distance in the form of a kernel function through a boosting algorithm. The kernel function is defined as a combination of some weak kernel functions, $K(x_i,x_j)=\sum\nolimits_{t=1}^{T}\alpha_tK_t(x_i,x_j)$. Each weak kernel $K_t(\cdot,\cdot)$ learns a Gaussian Mixture Model (GMM) of the data, and $K_t(x_i,x_j)$ represents the probability that both $x_i$ and $x_j$ belong to a same Gaussian component within the $t$-th GMM. The kernel $K$ is optimized by an exponential loss:
\begin{equation}
\ell = \sum\nolimits_{i,j} \exp\big(-y_{ij}K(x_i,x_j)\big),
\end{equation}
where $y_{ij}$ is 1 if $x_i$ and $x_j$ are from the same class, and -1 otherwise. Finally, a kernel  NN classifier can be formed.

\begin{figure}[b!]
\centering
\includegraphics[width=0.98\linewidth]{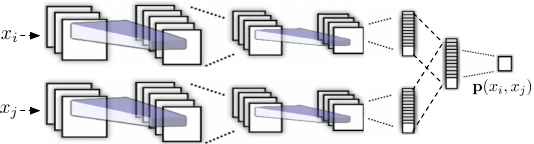}
\vspace{-0.5em}
\caption{Architecture of Siamese Nets~\cite{koch2015siamese}.  Twin CNNs share weights.}
\label{fig:SN}
\end{figure}
Siamese Nets~\cite{koch2015siamese} is the first work that brings deep neural networks into FSL tasks. It consists of  twin CNNs that share the same weights. The twin CNNs accept a pair of samples $(x_i,x_j)$ as inputs and their outputs at the top layer are combined in order to output a single pairwise similarity score $\mathbf{p}(x_i,x_j)$, as depicted in Fig.~\ref{fig:SN}. The twin CNNs are trained through the following binary cross-entropy loss:
\begin{equation}
\ell = \sum\nolimits_{i,j} y_{ij}\log\mathbf{p}(x_i,x_j)+(1-y_{ij})\log(1-\mathbf{p}(x_i,x_j)),
\end{equation} 
where $y_{ij}$$=$$1$ when $x_i$ and $x_j$ belong to the same class, and 0 otherwise.
The well-trained twin CNNs are frozen and used as a fixed similarity metric to make inference on FSL task $T$ in the manner of NN.

\begin{figure}[t!]
\centering
\includegraphics[width=0.98\linewidth]{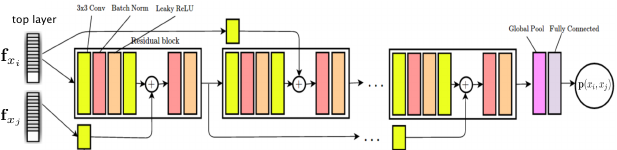}
\vspace{-0.8em}
\caption{Architecture of Skip Residual Pairwise Net (SRPN)~\cite{mehrotra2017generative}.}
\label{fig:SRPN}
\end{figure}

Triplet Ranking Nets~\cite{ye2018deep} extended Siamese Nets~\cite{koch2015siamese} from pairwise samples to triplets and used the triplet ranking loss~\cite{wang2014learning} to optimize the metric space and then capture the similarity between samples.
SRPN~\cite{mehrotra2017generative} is also an evolution of Siamese Nets~\cite{koch2015siamese}, which involves two main modifications: (1) Replacing the simple pairwise combination at top layer used by Siamese Nets with a more sophisticated skip residual network~\cite{he2016deep} that separates the intermediate computations for the pair of samples, as shown in Fig.~\ref{fig:SRPN}. (2) Using an additional GAN~\cite{goodfellow2014generative} to regularize the skip residual network by taking the skip residual network as a discriminator network and introducing another auto-encoder based generator. Thus, the metric loss of SRPN is naturally incorporated into the adversarial loss of GAN.


MM~\cite{kaiser2017learning} develops a life-long memory module to learn the similarity metric, which regards the feature of test data as its query $q$ and stores many continuously updated keys associated with values (\emph{i.e.}, class labels). This memory is optimized by the following hinge-based memory loss:
\begin{equation}
\ell = \sum\nolimits_{q}\big[q^\top k^- - q^\top k^+  + \gamma\big]_+,
\end{equation}
where $k^-$ and $k^+$ are respectively the negative and postive key in terms of $q$. Whether a key is positive or negative is determined by comparing its value to the class label of $q$.

\begin{figure}[t!]
\centering
\includegraphics[width=0.75\linewidth]{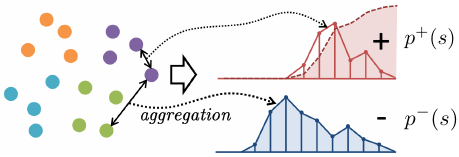}
\vspace{-0.8em}
\caption{Computation of histogram loss~\cite{ustinova2016learning} for a batch of samples. Dots denote the features of samples. The same color indicates the same class.}
\label{fig:hist}
\end{figure}

AdaptHistLoss~\cite{scott2018adapted} advise to adopt histogram loss~\cite{ustinova2016learning} to learn a feature space where the simple cosine distance can effectively measure the similarity between two features. Histogram loss suggests to construct two sets of similarities, $S^+ = \{s(\mathbf{f}_{x_i},\mathbf{f}_{x_j})|y_i=y_j\}$ and $S^- = \{s(\mathbf{f}_{x_i},\mathbf{f}_{x_j})|y_i\neq y_j\}$, where $\mathbf{f}_{x_i}$ denotes the feature of data $x_i$ with class label $y_i$ and $s(\cdot,\cdot)$ is the feature-level similarity metric (i.e., cosine similarity). Using $S^+$ and $S^-$, one can estimate the similarity distributions of positive and negative pairs as the histograms, as shown in Fig.~\ref{fig:hist}, which are denoted as $p^+(s)$ and $p^-(s)$ respectively. Then, the histogram loss is defined as the reverse probability that the similarity in a random negative pair is more than the similarity in a random positive pair:
\begin{equation}
\ell = \int_{-1}^1p^-(s)\Big[\int_{-1}^s p^+(z)dz \Big]ds = \mathbb{E}_{s\sim p^-}\Big[\int_{-1}^s p^+(z)dz\Big].
\end{equation}
Since histogram loss only focuses on the similarity distributions of positive pairs and negative pairs but agnostic to class labels, this metric can be directly transferred to FSL task $T$. 

Table~\ref{tab:metric_comparison} summarizes the main characteristics of the above metric learning based FSL approaches. In addition to these approaches, it should be noted that the Learn-to-Measure FSL approaches (see Section~\ref{sec:dmb:meta:measure}), strictly speaking, all belong to the scope of metric learning. Considering they pursue the similarity metric under the paradigm of meta learning, we discuss them in the section of meta learning.

\begin{figure}[b!]
\centering
\includegraphics[width=0.95\linewidth]{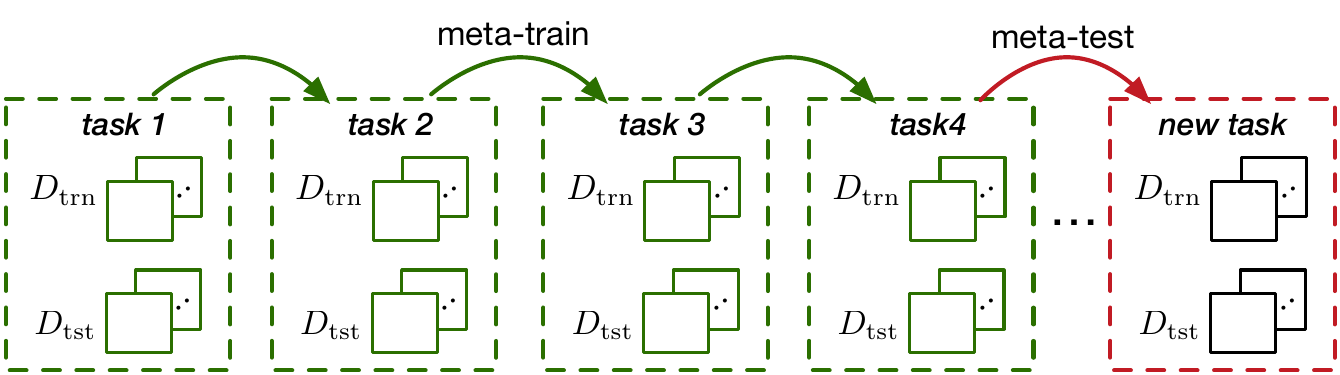}
\vspace{-0.5em}
\caption{General framework of meta learning based FSL approaches.}
\label{fig:meta_learning}
\end{figure}

\subsection{Meta Learning}~\label{sec:dmb:meta}
The idea of meta learning was proposed as early as the 1990s~\cite{bengio1990learning,naik1992meta,thrun1998learning}. As deep learning grew in popularity, some works proposed to utilize the meta learning policy to learn to optimize deep models~\cite{andrychowicz2016learning,li2017learning,chen2017learning}. In general, meta learning advocates to learn across tasks and then adapt to new tasks, as shown in Fig.~\ref{fig:meta_learning}, which aims to learn on the level of tasks instead of samples, and learns the task-agnostic learning systems instead of task-specific models.

\begin{figure*}[t!]
\centering
\includegraphics[width=0.98\linewidth]{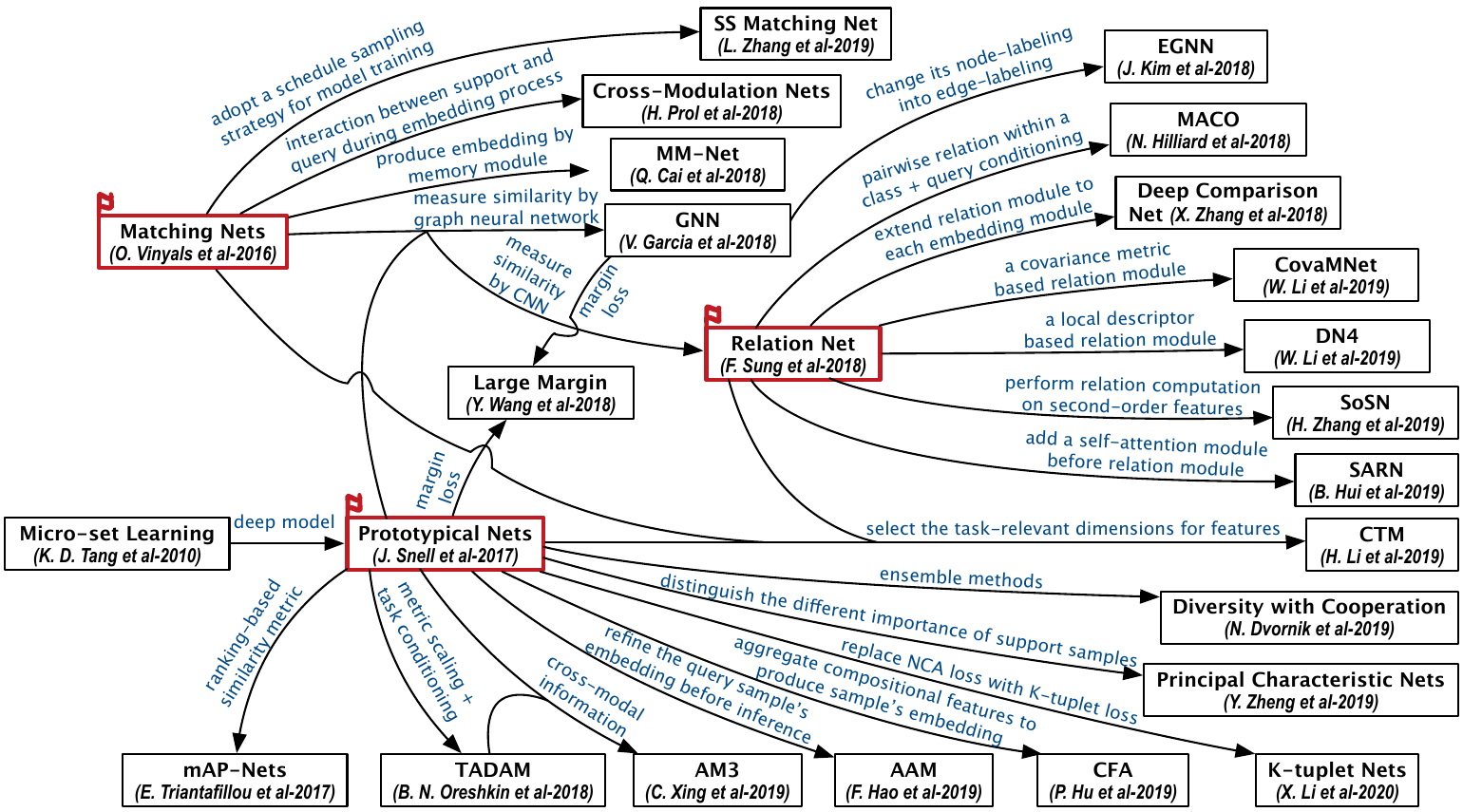}
\vspace{-0.5em}
\caption{Development relationship between different Learn-to-Measure FSL approaches.}
\label{fig:L2M}
\end{figure*}

FSL is a natural testbed to validate the capability of meta learning approaches across tasks where only a few labeled samples are given per task. Meta learning approaches process FSL problems in two stages: meta-train and meta-test. In meta-train, the model is exposed to many independent  supervised tasks $T$$\sim$$p(T)$ that are constructed on the auxiliary dataset $D_A$ (also called ``episode''~\cite{vinyals2016matching,finn2017model}) to learn how to adapt to future related tasks, where $P(T)$ defines a task distribution and the word of \emph{related} means that all tasks are from $P(T)$ and follow the same task paradigm, e.g., all tasks are $C$-way $K$-shot problems. Each meta-train task $T$ entails a task-specific dataset $D_{T}$$=$$\{D_{\mathrm{trn}}, D_{\mathrm{tst}}\}$, where $D_{\mathrm{trn}}$$=$$\{(x_i,y_i)\}_{i=1}^{N_{\mathrm{trn}}}$ and $D_{\mathrm{tst}}$$=$$\{(x_i,y_i)\}_{i=1}^{N_{\mathrm{tst}}}$. In meta-test, the model is tested on a new task $T$$\sim$$p(\mathcal{T})$ whose label space is disjoint with the labels seen during meta-train. In most cases, $D_{\mathrm{trn}}$ is called as support or description set and $D_{\mathrm{tst}}$ is called as query set. Accordingly, the samples of them are called as support samples and query samples, respectively. The meta learning objective is to find the model parameters $\theta$ that minimize the excepted loss $L(\cdot;\theta)$ across all tasks:
\begin{equation}
\min\nolimits_{\theta}\mathbb{E}_{T\sim P(T)}L(D_T;\theta).
\end{equation}
We must emphasize that the meta learning is a high-level cross-task learning strategy rather than a specific FSL model. Based on what the meta learning model seeks to meta-learn behind this learning strategy, we generally summarize the meta learning based FSL methods into five sub-categories: Learn-to-Measure (L2M), Learn-to-Finetune (L2F), Learn-to-Parameterize (L2P), Learn-to-Adjust (L2A) and Learn-to-Remember (L2R).

\subsubsection{Learn-to-Measure}~\label{sec:dmb:meta:measure}
The L2M  approaches inherit the main idea of metric learning in essence as shown in Fig.~\ref{fig:metric_learning}, but they are different from the metric learning based FSL approaches as described in Section~\ref{sec:dmb:metric} in the implementation level: the L2M approaches adopt the meta learning policy to learn the similarity metric that is expected to be transferrable across different tasks. L2M has always been an important branch of meta learning based FSL approaches, and several milestone meta learning approaches such as Matching Nets~\cite{vinyals2016matching}, Prototypical Nets~\cite{snell2017prototypical}, and Relation Net~\cite{yang2018learning} all belong to the L2M category.

We first describe the general pipeline of L2M mathematically. For a task $T$, let $x_i$ be a support sample in $D_{\mathrm{trn}}$ and $x_j$ be a query sample in $D_{\mathrm{tst}}$, and let $f(\cdot;\theta_f)$ and $g(\cdot;\theta_g)$ be the embedding models that map the support and query samples into features respectively. Moreover, all L2M approaches contain a metric module $S(f,g;\theta_S)$ to measure the similarity between support and query samples, which might be a parameter-free distance metric (e.g., Euclidean distance, cosine distance) or a learnable network. The similarity output by this metric module is used to form the final predicted probability of the query sample. Existing L2M approaches are different mainly in the model design and selection of $f$, $g$ and $S$, and we draw the development relationship between different L2M approaches in Fig.~\ref{fig:L2M} in a highly abstract way.

\begin{itemize}
\item \emph{Prototypical Nets and its Variants}
\end{itemize}
The pioneer of L2M is Micro-set Learning~\cite{tang2010optimizing}, although the concept of meta learning was not mentioned by it at that time. This approach artificially constructs many mirco-sets like test scenarios from the auxiliary dataset $D_A$, and each micro-set contains several support and query samples belonging to a few non-task classes. Both the embedding models $f$ and $g$ are realized through a weight-shared linear projection (i.e., $f=g$) and the similarity metric $S$ is achieved by Euclidean distance. Moreover, NCA~\cite{goldberger2005neighbourhood} is used to measure the final probability. Actually, the micro-sets are equivalent to the so-called \emph{episodes} nowadays, and each micro-set is a meta-train task $T$. Importantly, if we replace the linear projection model with a deep learning based embedding model such as CNN,  Micro-set Learning~\cite{tang2010optimizing} would evolve into the classic Prototypical Nets~\cite{snell2017prototypical}. It takes the center of congener support samples' embeddings as the prototype of this class
\vspace{-.2em} 
\begin{equation}
p_c = \frac{1}{K}\sum\nolimits_{(x_i,y_i)\in D_{\mathrm{trn}}} \mathbbm{1}(y_i==c) f(x_i;\theta_f),
\end{equation} 
and then also leverages the Euclidean distance based NCA like Micro-set Learning~\cite{tang2010optimizing}  to predict the probability:
\vspace{-.2em}
\begin{equation}
\label{eq:NCA}
P(y_j=c|x_j) =\frac{\exp\Big(-d\big(g(x_j;\theta_g),p_c\big)\Big)}{\sum_{c'=1}^{C}\exp\Big(-d\big(g(x_j;\theta_g),p_{c'}\big)\Big)},
\end{equation} 
where $f$ and $g$ are also weight-shared embedding models (i.e., $f=g$). This L2M framework is an important cornerstone of many subsequent FSL approaches. In~\cite{triantafillou2017few}, mAP-Nets were proposed to learn an informative similarity metric  from the perspective of information retrieval. It chooses to optimize an mAP-based ranking loss within each meta-train task using Structure SVM~\cite{tsochantaridis2005large} or Direct Loss Minimization~\cite{hazan2010direct}.
TADAM~\cite{oreshkin2018tadam} further optimized the similarity metric $S$ of Prototypical Nets by introducing a metric scaling factor $\alpha$ and transformed the original task-irrelevant $f$ into a task-conditioning embedding model through a task embedding network (TEN)~\cite{oreshkin2018tadam}. The TEN follows the key idea of FILM conditioning layer~\cite{perez2018film}, and customizes some adjustment parameters (e.g., scaling and shift meta parameters) for the embedding model $f$ in light of the current task representation. AM3~\cite{xing2019adaptive} incorporate extra cross-modal information (e.g., semantic representations) into Prototypical Nets and TADAM to enhance the metric learning process. Specifically, it used GloVe~\cite{pennington2014glove} to extract word embeddings for the semantic class labels and then built a new prototype by a convex combination of both the visual feature and word embedding. AAM~\cite{hao2019instance} proposed to refine the query sample's embedding before Eq.~(\ref{eq:NCA}) to render it closer to its corresponding class center. CFA~\cite{hu2019weakly} achieved a compositional feature extraction for images instead of the vanilla image-to-vector mapping. K-tuplet Nets~\cite{li2020revisiting} changed the NCA loss of Prototypical Nets into a K-tuplet metric loss. Y. Zheng~\emph{et al.}~\cite{zheng2019principal} believed that the average prototype ignores the different importance of different support samples and thus proposed Principal Characteristic Nets. Diversity with Cooperation~\cite{dvornik2019diversity} achieved an ensemble of Prototypical Nets to encourage all individual networks to cooperate while encourage prediction diversity.

\begin{figure}[b!]
\centering
\includegraphics[width=0.8\linewidth]{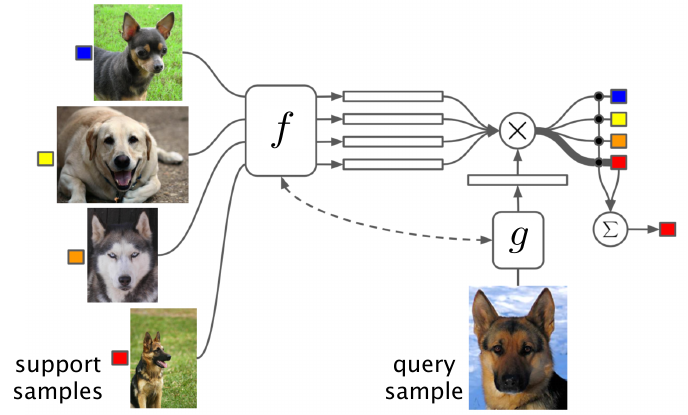}
\vspace{-0.5em}
\caption{Matching Nets architecture~\cite{vinyals2016matching} (4-way 1-shot task for example).}
\label{fig:MN}
\end{figure}

\begin{itemize}
\item \emph{Matching Nets and its Variants}
\end{itemize}
The first deep learning based L2M approach is Matching Nets~\cite{vinyals2016matching}. As shown in Fig.~\ref{fig:MN}, it predicts the probability of query sample $x_j$ by measuring the cosine similarity between the embedding of $x_j$ and each support sample's embedding:
\begin{equation}
p(\hat{y}_j|x_j, D_{\mathrm{trn}}) = \sum\nolimits_{(x_i,y_i)\in D_{\mathrm{trn}}}a(x_j,x_i)\cdot \mathbf{y}_i,
\end{equation}
where $\mathbf{y}_i$ is a $C$-dimensional one-hot label vector ($C$-way FSL task) corresponding to the label $y_i\in \mathbb{R}^1$, and 
\begin{equation}
a(x_j, x_i)=\frac{\exp\Big(c\big(g(x_j;\theta_g), f(x_i;\theta_f)\big)\Big)}{\sum\nolimits_{(x,y)\in D_{\mathrm{trn}}}\exp\Big(c\big(g(x_j;\theta_g), f(x;\theta_f)\big)\Big)}.
\end{equation}
Matching Nets is different from Prototypical Nets in two aspects. First, the embedding model $f$ and $g$ of Matching Nets are two different networks. Concretely, $f$ is a combination of CNN and BiLSTM~\cite{hochreiter1997long} that aims to achieve full context embeddings (FCE)~\cite{vinyals2016matching} for the few support samples, while $g$ is a $f$-conditioning model that generates the features for query samples by a content attention mechanism. Second, the similarity metric $S$ in Matching Nets is cosine distance instead of Euclidean distance. Several follow-up works have made some modifications and extensions based on Matching Nets. For example, Cross-Modulation Nets~\cite{prol2018cross} modified the conditioning mechanism between $f$ and $g$ into a cross-modulation mechanism using FILM layers~\cite{perez2018film}, which allows support and query samples to interact during the feature embedding process. MM-Net~\cite{cai2018memory} developed a memory module~\cite{weston2014memory} to produce feature embeddings, and the parameters of $g$ are generated by this memory module. SS Matching Nets~\cite{zhang2019scheduled} considered the semantic diversity and similarity of class labels and exploited a scheduled sampling strategy to facilitate the model training of Matching Nets.

\begin{figure}[b!]
\centering
\includegraphics[width=0.98\linewidth]{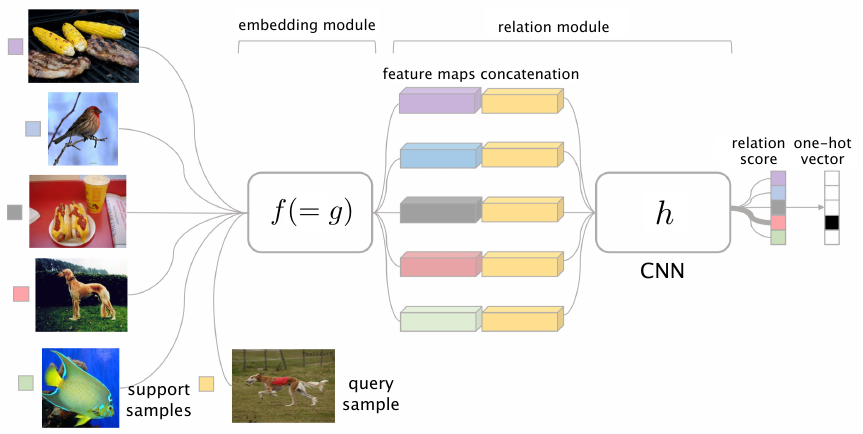}
\vspace{-0.5em}
\caption{Relation Net architecture~\cite{yang2018learning} (5-way 1-shot task for example).}
\label{fig:RN}
\end{figure}

\begin{figure*}[t!]
\centering
\includegraphics[width=0.95\linewidth]{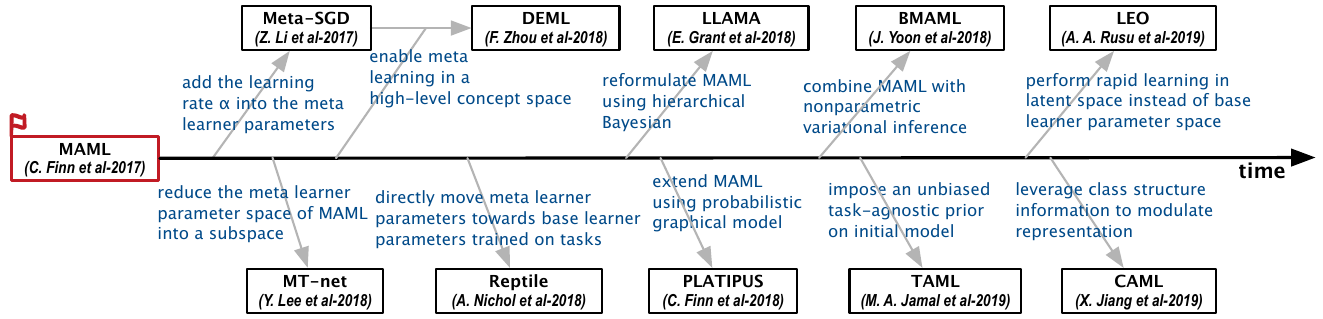}
\vspace{-1.em}
\caption{Development relationship between MAML~\cite{finn2017model} and its variants.}
\label{fig:L2F}
\end{figure*}

\begin{itemize}
\item \emph{Relation Net and its Variants}
\end{itemize}
Unlike Prototypical Nets~\cite{snell2017prototypical} and Matching Nets~\cite{vinyals2016matching} which use the non-parametric Euclidean distance or cosine distance to measure the similarity between pairwise features, Relation Net~\cite{yang2018learning} adopted a learnable CNN (denoted by $h(\cdot;\theta_h)$ here) to measure pairwise similarity, which takes the concatenation of feature maps of support sample $x_i$ and query sample $x_j$ as input and outputs their relation score $r(x_i,x_j)$, as shown in Fig.~\ref{fig:RN}, which is formulated as
\begin{equation}
r(x_i,x_j)=h\Big(\mathcal{C}\big(f(x_i;\theta_f), g(x_j;\theta_g)\big);\theta_h\Big) \in [0,1],
\end{equation}
where $f=g$ and $\mathcal{C}$ denotes the feature maps concatenation. It needs to be noted that, for Relation Net, the embeddings output by $f$ (or $g$) are  feature maps rather than feature vectors. Based on Relation Net, MACO~\cite{hilliard2018few} designed a relational stage after $f$ to form the pairwise relation features within one class and then used a query conditioning operation to predict the probability of query samples. Deep Comparison Net~\cite{zhang2018deep} extended Relation Net by deploying the relation module to each layer of the embedding model $f$. CovaMNet~\cite{li2019distribution} and  DN4~\cite{li2019revisiting} replaced the relation module in Relation Net~\cite{yang2018learning} by a covariance metric network and a  deep local descriptor based image-to-class metric module, respectively. SoSN~\cite{zhang2019power} chose to perform the relation computation on the second-order representation of feature maps. SARN~\cite{hui2019self} introduced a self-attention mechanism into Relation Net for capturing non-local features and enhancing representation.

\begin{itemize}
\item \emph{Other L2M Approaches}
\end{itemize}

In addition to the above three mainstreams, there have also several hybrid variants based on them. For example, GNN~\cite{garcia2018few} replaced the Euclidean distance in Prototypical Nets~\cite{snell2017prototypical} and the cosine distance in Matching Net~\cite{vinyals2016matching} with a learnable graph neural network where the nodes are set to the samples' embeddings and the edges are treated as the similarity between two samples. Conversely, EGNN~\cite{kim2019edge} exchanged the roles of both nodes and edges in GNN~\cite{garcia2018few}, transforming it from a node-labeling framework to an edge-labeling one. Y. Wang \emph{et al.}~\cite{wang2018large} proposed to impose a large margin constraint and augment the final classification loss of Prototypical Nets or GNN with a margin loss such that the metric space could be more discriminative. In consideration of the different importance of
one feature element in different tasks, H. Li \emph{et al.} introduced Category Traversal Module (CTM)~\cite{li2019finding} to select the most task-relevant dimensions of feature embeddings. CTM~\cite{li2019finding} can be used as a plug-and-play module for other approaches like Matching Nets~\cite{vinyals2016matching}, Prototypical Nets~\cite{snell2017prototypical} and Relation Net~\cite{yang2018learning}.

\subsubsection{Learn-to-Finetune}~\label{sec:dmb:meta:finetune}
L2F approaches suggest to finetune a base learner for task $T$ using its few support samples and make the base learner converge fast on these samples within several parameter update steps. Generally, every L2F approach contains a base learner and a meta learner. The base learner is for a specific task, which takes the sample as input and outputs the prediction probability. The base learner is learned by the higher-level meta-learner that is learned on a bunch of meta-train tasks to maximize the combined generalization power of the base learner on all tasks. Let $\theta_b$ and $\theta_m$ denote the parameters of base learner and meta learner, respectively. The learning process of L2F occurs at two levels. Gradual learning is performed across tasks, which aims to optimize the meta learning parameters $\theta_m$ and then facilitate the rapid learning of base learner for each specific task. Two milestone L2F approaches are MAML~\cite{finn2017model} and Meta-Learner LSTM~\cite{ravi2016optimization}. 


MAML~\cite{finn2017model} is an elegant meta learning framework with strong interpretability, which has a profound influence on the field of meta learning and FSL. Its core idea is to search for a good parameter initialization for $\theta_b$ by cross-task training strategy such that the base learner with this initialization can rapidly generalize  new tasks using a few support samples. Concretely, as the base learner copes with a task $T$, the one-step updated parameter $\theta'_b$ of base learner is computed as
\vspace{-0.5em}
\begin{equation}
\theta^T_b=\theta_b-\alpha \nabla_{\theta_b}L(D_{\mathrm{trn}}^T,\theta_b),
\label{eq:rapid_learning}
\end{equation}
where $\alpha$ is learning rate and $L(D_{\mathrm{trn}}^T,\theta_b)$ is the loss on support set of task $T$ when base learner parameter starts with $\theta_b$. On the meta level, MAML optimizes meta learner  by balancing the loss with updated base learner $\theta^T_b$ over many tasks:
\begin{equation}
\theta_m=\theta_m-\beta \nabla_{\theta_m}\sum\nolimits_{T\sim P(T)}L(D_{\mathrm{tst}}^T,\theta^T_b).
\label{eq:gradual_learning}
\end{equation}
Note that the meta learner in MAML~\cite{finn2017model} is actually the base learner, that is, the meta learner parameter satisfy $\theta_m=\theta_b$. Eq.~(\ref{eq:rapid_learning}) is the rapid learning process that aims to finetune the base learner towards the specific task, while Eq.~(\ref{eq:gradual_learning}) is the gradual learning process that is intended to distill an appropriate parameter initialization for base learner.

Additionally, many L2F approaches belonging to MAML variants~\cite{li2017meta,lee2018gradient,nichol2018reptile,zhou2018deep,grant2018recasting,finn2018probabilistic,yoon2018bayesian, jamal2019task,rusu2019meta,jiang2019learning} have been developed recently. The relationship between MAML and them is shown in Fig.~\ref{fig:L2F}. Meta-SGD~\cite{li2017meta} proposed to meta-learn not just the base learner initialization, but also the base learner update direction and learning rate. Thus, Meta-SGD modified the learning rate $\alpha$ in Eq.~(\ref{eq:rapid_learning}) into a learnable vector $\bm{\alpha}$ and adds it into meta learner parameters:
\begin{equation}
\begin{aligned}
\theta^T_b&=\theta_b-\bm{\alpha}\circ \nabla_{\theta_b}L(D_{\mathrm{trn}}^T,\theta_b),\\
(\theta_b,\bm{\alpha})&=(\theta_b,\bm{\alpha})-\beta \nabla_{(\theta_b,\bm{\alpha})}\sum\nolimits_{T\sim P(T)}L(D_{\mathrm{tst}}^T,\theta^T_b).
\label{eq:Meta-SGD}
\end{aligned}
\end{equation}
Follow this line, DEML~\cite{zhou2018deep} made an incremental change for Meta-SGD, which equipped the meta-learner of Meta-SGD with a concept generator to enable learning to learn in a high-level concept space. In contrast, MT-net~\cite{lee2018gradient} proposed to reduce the meta learner parameter space of MAML into a subspace that is composed of each layer's activation space and perform the rapid learning on this subspace. To avoid the computation of second-order derivative in MAML during gradual learning, A. Nichol \emph{et al.} developed Reptile~\cite{nichol2018reptile} that directly moves the meta learner parameter $\theta_m$ towards the base learner parameters $\theta_b^T$ that are updated on many tasks:
\begin{equation}
\theta_b=\theta_b-\beta\sum\nolimits_{T\sim P(T)}(\theta_b-\theta_b^T).
\label{eq:gradual_learning}
\end{equation} 
LLAMA~\cite{grant2018recasting} reformulated MAML by  hierarchical Bayesian and made an extension to MAML from the perspective of Bayesian posterior estimation.
In consideration of the issue of task ambiguity when learning from small amounts of samples, PLATIPUS~\cite{finn2018probabilistic} extended MAML using probabilistic graphical model and reframes it as a graph model inference problem, enabling simple and effective sampling of base learners for new tasks at meta-test time. In contrast, BMAML~\cite{yoon2018bayesian} coped with the model uncertainty when learning from a few samples by combining MAML with a non-parametric variational inference, i.e., Stein Variational Gradient Descent (SVGD)~\cite{liu2016stein}. In addition, BMAML proposed a novel Chaser loss~\cite{yoon2018bayesian} during gradual learning to optimize the meta learner parameters $\theta_m$. 
TAML~\cite{jamal2019task} imposed an unbiased task-agnostic prior, which is achieved by an entropy-maximization/reduction or an inequality-minimization, on the initial model to prevent it from over-performing on meta-train tasks.  
LEO~\cite{rusu2019meta} designed a  latent embedding $\mathbf{z}$, which is produced from support set of task $T$ via an encoder $\mathbf{z}$$=$$E(D^T_{\mathrm{trn}};\theta_E)$, to generate the base learner parameter $\theta_b$, i.e., $\theta_b$$=$$ G(\mathbf{z}; \theta_G)$, and performed the rapid learning in the low-dimensional latent space instead of the high-dimensional base learner parameter space like Eq.~(\ref{eq:rapid_learning}):
\begin{equation}
\mathbf{z}^T=\mathbf{z}-\bm{\alpha}\circ \nabla_{\mathbf{z}}L(D_{\mathrm{trn}}^T,G(\mathbf{z}; \theta_G)).
\end{equation}
Obviously, the combination of encoder $E(\cdot;\theta_E)$ and generator $G(\cdot;\theta_G)$ plays the role of meta learner of LEO, and thus its gradual learning process can be described as
\begin{equation}
\theta_m=\theta_m-\beta \nabla_{\theta_m}\sum\nolimits_{T\sim P(T)}L(D_{\mathrm{tst}}^T,G(\mathbf{z}^T; \theta_G)),
\end{equation}
where the meta learner parameter $\theta_m$$=$$(\theta_E, \theta_G, \bm{\alpha})$. Compared with MAML, CAML~\cite{jiang2019learning} leverages the label structure to modulate the representations of base learner in light of the current task. Specifically, a parametric conditional transformation module is designed to perform the representation modulation. The combination of this module and the base learner acts as the meta learner of CAML, which is meta-learned via the gradual learning strategy adopted by MAML.

\begin{figure}[t!]
\centering
\includegraphics[width=0.98\linewidth]{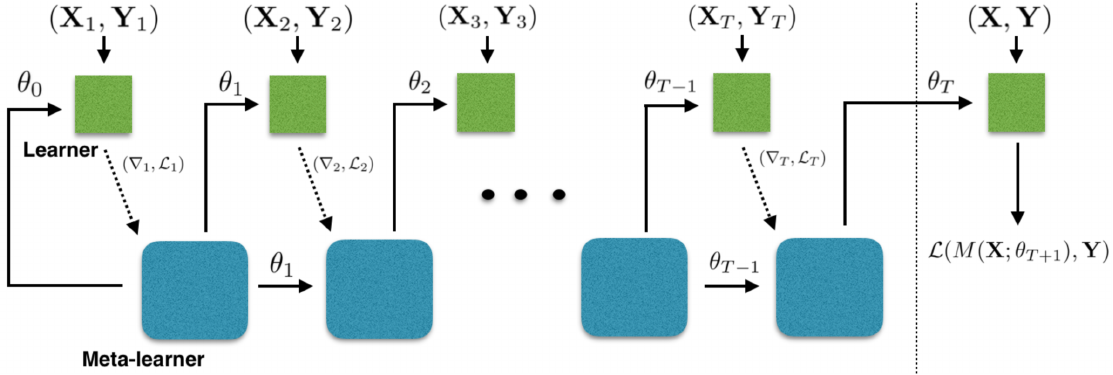}
\vspace{-0.5em}
\caption{Forward computational process of Meta-Learner LSTM for one task~\cite{ravi2016optimization}. Green box is base learner and blue box is meta learner.}
\label{fig:meta-learner}
\end{figure}

Another representative L2F approach is Meta-Learner LSTM~\cite{ravi2016optimization}, which suggests finetuning the base learner on the few support samples by a LSTM-based meta learner. As shown in Fig.~\ref{fig:meta-learner}, the LSTM-based meta-learner takes as input the loss and gradient of base learner with respect to each support sample, and its hidden state is treated as the updated base learner parameter, which would be used to handle the next support sample. In this framework, the vanilla gradient-based optimization for base learner parameters is superseded by an LSTM in the hope of learning appropriate parameter updates specifically for the scenario where a few updates will be made. J. Nie \emph{et al.} extended Meta-Learner LSTM  to a dual version, called Meta-Learner Dual-LSTM~\cite{nie20203d}, and applied it to  3D model few-shot classification tasks.
In addition, a recent L2F approach is MTL~\cite{sun2019meta}, which developed a light-weight scaling and shifting network attached to the frozen base learner to reduce the probability of overfitting when finetuned on the few support samples.

\begin{figure}[b!]
\centering
\includegraphics[width=0.75\linewidth]{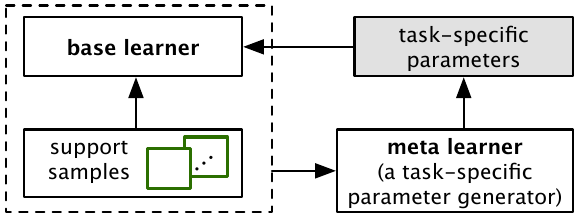}
\vspace{-0.5em}
\caption{General framework of Learn-to-Parameterize FSL approaches.}
\label{fig:L2P}
\end{figure}

\subsubsection{Learn-to-Parameterize}~\label{sec:dmb:meta:parameterrize}
L2P is another kind of popular meta learning based FSL approach, which adheres to a straightforward idea: parameterizing the base learner or some subparts of base learner for a novel task so that it can address this task specifically. As shown in Fig.~\ref{fig:L2P}, most L2P approaches also contain both base learner and meta learner like L2F approaches, but the difference is that, for L2P approaches, the two learners are trained synchronously within each task and the meta learner is essentially a task-specific parameter generator. For a task $T$, the meta learner is expected to generate some $T$-specific parameters for the base learner or its subparts in light of the few support samples of task $T$ and the state of current base learner that is handling these support samples. At this point, L2P approaches attempt to  learn how to parameterize the base learner to render it applicable to the specific task.

\begin{table}[t!]
\begin{center}
\caption{Summary of Learn-to-Parameterize FSL approaches.}
\vspace{-1.8em}
\label{tab:L2P}
\scalebox{0.78}{
\begin{tabular}{lll}
\toprule[1pt]
\textbf{Approaches} & \textbf{Parameter Generation Module} & \textbf{Generated Parameters}   \\
\cmidrule[0.5pt](lr){1-3}
Siamese Learnet~\cite{bertinetto2016learning} & single-stream Siamese Nets~\cite{koch2015siamese} &  Conv. layer weights\\
\cmidrule[0.5pt](lr){1-3}
Regression Nets~\cite{wang2016learning} & MLP-based weight transformation & SVM weights\\
\cmidrule[0.5pt](lr){1-3}
Dynamic Nets~\cite{gidaris2018dynamic} & attention-based weight composition & predictor weights\\
\cmidrule[0.5pt](lr){1-3}
Acts2Params~\cite{qiao2018few} & MLP-based parameter predictor &Softmax layer weights\\
\cmidrule[0.5pt](lr){1-3}
Imprinting~\cite{qi2018low} & MLP-based weight transformation & predictor weights\\
\cmidrule[0.5pt](lr){1-3}
DCCN~\cite{zhao2018dynamic} & LSTM embedding module & Conv. layer weights\\
\cmidrule[0.5pt](lr){1-3}
MeLA~\cite{wu2018meta} & Auto-Encoder & all Conv. layer weights \\
\cmidrule[0.5pt](lr){1-3}
DAE~\cite{gidaris2019generating} & graph neural network & predictor weights \\
\cmidrule[0.5pt](lr){1-3}
VERSA~\cite{gordon2019meta} & probabilistic amortization network & Softmax layer weights\\
\cmidrule[0.5pt](lr){1-3}
R2-D2~\cite{bertinetto2019meta} & ridge regression layer & predictor weights\\
\cmidrule[0.5pt](lr){1-3}
MetaOptNet~\cite{lee2019meta} & SVM & predictor weights\\
\cmidrule[0.5pt](lr){1-3}
LGM-Net~\cite{li2019lgm} & VAE-like weight generator & Conv. layer weights\\ 
\bottomrule[1pt]
\end{tabular}}
\end{center}
\end{table}

\begin{figure}[t!]
\centering
\includegraphics[width=0.85\linewidth]{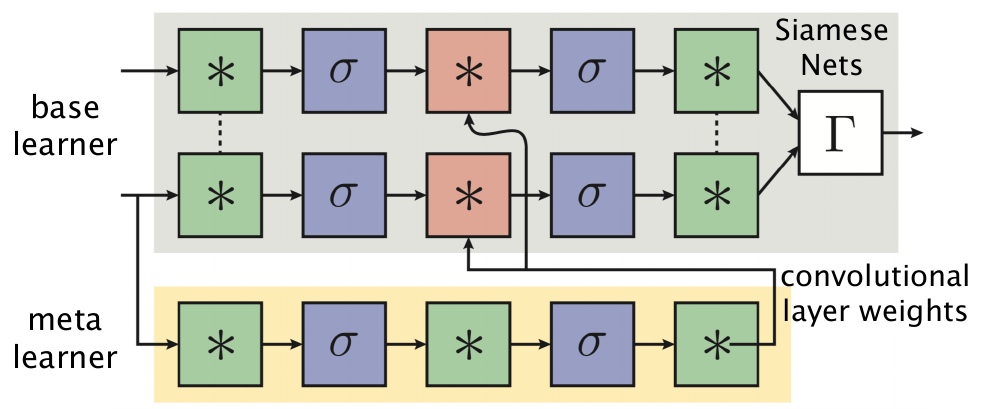}
\vspace{-0.8em}
\caption{Siamese Learnet architecture~\cite{bertinetto2016learning}. ``$*$'': convoluational layer.}
\label{fig:SLN}
\end{figure}

\begin{figure}[b!]
\centering
\includegraphics[width=0.99\linewidth]{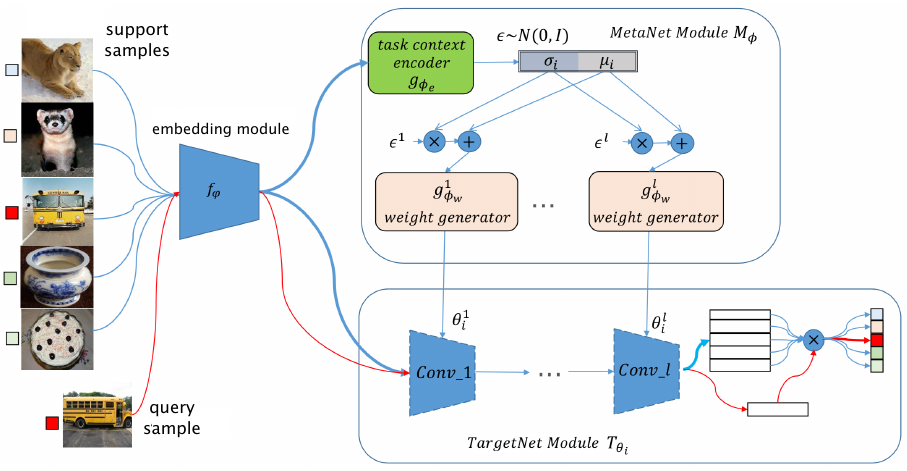}
\vspace{-0.8em}
\caption{LGM-Net architecture~\cite{li2019lgm} (5-way 1-shot task for example).}
\label{fig:LGM}
\end{figure}

Many L2P approaches have been proposed in recent years, which developed various parameter generation modules to parameterize different parts of base learner. They might parameterize the task-specific predictor in base learner~\cite{gidaris2018dynamic,qiao2018few,qi2018low,gidaris2019generating,bertinetto2019meta,lee2019meta,gordon2019meta}, or the intermediate feature extraction layers in base learner~\cite{bertinetto2016learning,zhao2018dynamic,li2019lgm}, even or the whole base learner~\cite{wang2016learning,wu2018meta}. The characteristics of different L2P approaches are summarized in Table~\ref{tab:L2P}. 

Siamese Learnet~\cite{bertinetto2016learning}, as shown in Fig.~\ref{fig:SLN}, used Siamese Nets~\cite{koch2015siamese} as its base learner where one intermediate convolutional layer (red block in Fig.~\ref{fig:SLN}) is designed to be dynamic to different tasks. Another single-stream Siamese Nets is deployed as its meta learner to generate  task-specific weights for this convolutional layer. 
LGM-Net~\cite{li2019lgm} is a state-of-the-art L2P  approach, which developed a MetaNet Module (i.e., meta learner) to generate the weights of TargetNet Module (i.e., base learner) on the basis of the few support samples in each FSL task, as shown in Fig.~\ref{fig:LGM}. Specifically, the MetaNet Module in LGM-Net takes the average embedding of support samples as input and produces the weights for each convolutional layer in base learner through an encoder-decoder model with multivariate Gaussian sampling. Once parameterized, the base learner of LGM-Net would make FSL inference similar to the classic Matching Nets~\cite{vinyals2016matching}.

Regression Nets~\cite{wang2016learning} pursued a task-agnostic transformation of base learner's weights from a small-sample model to a large-sample model. Through this weight transformation, one can obtain  more general weights for base learner albeit only on a few training samples. Dynamic Nets~\cite{gidaris2018dynamic} advocated parameterizing the task-specific predictor by combining the average representation of the few support samples and an attention-based weight composition on non-task predictor weights. Acts2Params~\cite{qiao2018few} learned an MLP-based parameter predictor that maps neuron activations into the weights of the final Softmax predictor. Once well trained, it could directly predict the task-specific Softmax weights by taking the activations of the few support samples of this task as its inputs. Similarily, Imprinting~\cite{qi2018low} also inherited the mapping idea which transforms the embeddings of support samples into the task-specific predictor weights via an MLP. VERSA~\cite{gordon2019meta} exploited a versatile amortization network  that accepts support samples as input and outputs the parameter distribution for task-specific Softmax predictor. DAE~\cite{gidaris2019generating} used a graph neural network based denoising Auto-Encoder (AE) to generate final predictor parameters. Comparably, R2-D2~\cite{bertinetto2019meta} adopted a differentiable ridge regression layer to parameterize the task-specific predictor, while MeteOptNet~\cite{lee2019meta} advocated a differentiable convex optimization on SVM for generating final predictor weights. MeLA~\cite{wu2018meta} is similar to LGM-Net~\cite{li2019lgm} since they both tried to customize the convolutional layers of base learner (MeLA parameterized several rear convolutional layers while LGM-Net parameterized all convolutional layers) for the specific task via an encoder-decoder based generator (MeLA used AE while LGM-Net used an AE variant like Variational Auto-Encoder~\cite{kingma2014auto}).

\begin{figure}[b!]
\centering
\includegraphics[width=0.85\linewidth]{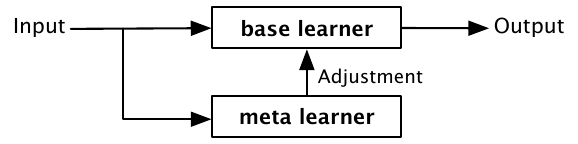}
\vspace{-0.5em}
\caption{General framework of Learn-to-Adjust FSL approaches.}
\label{fig:L2A}
\end{figure}

\subsubsection{Learn-to-Adjust}~\label{sec:dmb:meta:adjust}
As depicted in Fig.~\ref{fig:L2A}, the core ideology of L2A approaches is to adaptively adjust the computation flow or computing nodes in the base learner for a specific sample to make this sample compatible with the base learner. One may find that both L2P and L2A are similar since they all use the meta learner to change the base learner, but L2A approaches have two distinctive characteristics different from L2P. (1) The degree of change on the base learner by L2A approaches is lighter since they only make some incremental adjustments to base learner instead of a complete parameterization to the base learner or its subparts like L2P. (2) The change on the base learner by L2A approaches is more fine-grained since the adjustment of L2A is sample-specific while the parameterization of L2P is task-specific. 

Several typical L2A approaches include MetaNet~\cite{munkhdalai2017meta}, CSNs~\cite{munkhdalai2018rapid}, MetaHebb~\cite{munkhdalai2018metalearning} and  FEAT~\cite{ye2020few}, which differ in the selection for the parts needing to be adjusted as well as the design for the generated adjustment, as described in Table~\ref{tab:L2A}. MetaNet~\cite{munkhdalai2017meta} deployed a fast-weight layer attached to each layer of the base learner. The weights of each fast-weight layer are meta-generated by an external meta learner in light of the input sample. These collateral branch layers are used to adjust the intermediate values of the input sample during the feedforward process. CSNs~\cite{munkhdalai2018rapid} selected to adjust the neuron state (i.e., pre-activation) of each hidden node in the base learner. Specifically, it combined a memory module with an attention-based memory read mechanism to generate the condition shift for each neuron pre-activation in the base learner. MetaHebb~\cite{munkhdalai2018metalearning} added an auxiliary fast-weight matrix in pre-Softmax layer, which is meta-generated via Hebbian learning~\cite{hebb1949organization} and is expected to adjust the input's internal representation fed into final Softmax layer. FEAT~\cite{ye2020few} proposed to adapt the embeddings of support samples to render them more discriminative for the task in hand, and four kinds of set-to-set functions including BiLSTM~\cite{hochreiter1997long}, DeepSets~\cite{zaheer2017deep}, GCN~\cite{kipf2017semi} and Transformer~\cite{vaswani2017attention} that aim to transform the original embedding vector to the adapted vector were investigated by~\cite{ye2020few}.

\begin{table}[t!]
\begin{center}
\caption{Summary of Learn-to-Adjust FSL approaches.}
\vspace{-1.6em}
\label{tab:L2A}
\scalebox{0.78}{
\begin{tabular}{lll}
\toprule[1pt]
\textbf{Approaches} & \textbf{Parts to be Adjusted} & \textbf{Adjustment}  \\
\cmidrule[0.5pt](lr){1-3}
MetaNet~\cite{munkhdalai2017meta} & parametric layers of base learner & layer-wise fast weights\\
\cmidrule[0.5pt](lr){1-3}
CSNs~\cite{munkhdalai2018rapid} & neuron state of base learner & neuron-wise conditional shift\\
\cmidrule[0.5pt](lr){1-3}
MetaHebb~\cite{munkhdalai2018metalearning} &  pre-Softmax layer of base learner & pre-Softmax fast-weight matrix\\
\cmidrule[0.5pt](lr){1-3}
FEAT~\cite{ye2020few} & embeddings of support samples & vector-to-vector transformer\\
\bottomrule[1pt]
\end{tabular}}
\end{center}
\end{table}

\begin{figure}[t!]
\centering
\includegraphics[width=1\linewidth]{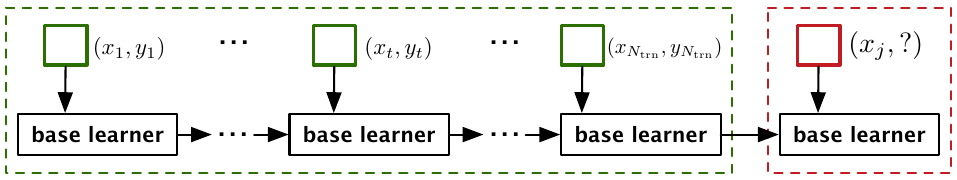}
\vspace{-1.8em}
\caption{General framework of Learn-to-Remember FSL approaches.}
\label{fig:L2R}
\end{figure}

\subsubsection{Learn-to-Remember}~\label{sec:dmb:meta:remember}
Several representative FSL approaches, such as MANN~\cite{santoro2016meta}, ARCs~\cite{shyam2017attentive}, SNAIL~\cite{mishra2018simple} and APL~\cite{ramalho2019adaptive}, belong to the kind of L2R. As shown in Fig.~\ref{fig:L2R}, its primary idea is to model the support set of an FSL task as a sequence and formulate the FSL task as a sequence learning task, where the query sample is required to match with previously seen information (i.e., support samples). Thus, the base learner of L2R approaches usually entails a temporal network to handle the few support samples. For example, MANN~\cite{Santoro2016One} utilized a memory-augmented Neural Turing Machine (NTM)~\cite{graves2014neural} to rapidly assimilate the support samples and then retrieve them when the query sample arrives. ARCs~\cite{shyam2017attentive} developed an attention based RNN to realize dynamic comparison between samples. SNAIL~\cite{mishra2018simple} devised a temporal convolutional network with soft attention to aggregate previously seen information and pinpoint specific information. APL~\cite{ramalho2019adaptive} designed a surprised-based memory network to remember the most informative support samples it has encountered. 

\subsubsection{Discussion}
The above five kinds of meta learning approaches all focus on dealing with FSL problems, however, each of them has their strengths or weaknesses. The L2M approaches would not be restricted by the specific settings of test scenarios since they only leverage the similarity between samples to make ultimate inference regardless of the number of classes and support samples per class (i.e., way/shot-agnostic). The L2F approaches need to be finetuned on each new task using the few support samples, which may yield a relatively long adaptation period to prepare for each task. One common challenge faced by L2P and L2A approaches is a large number of model parameters because they have to deploy another meta learner completely different from base learner to generate a series of model parameters or adjustment parameters. Besides, the model complexity of meta learner depends heavily on the parameter quantity needing to be generated, thus increasing the difficulty of model training. Due to the ceiling effect of long-term dependence in sequence learning~\cite{hochreiter2001gradient}, the L2A approaches are difficult to generalize the case with slightly more support samples in a task.

\subsection{Other Approaches}~\label{sec:dmb:other}
In addition to the aforementioned three mainstreams, i.e., augmentation (Section~\ref{sec:dmb:aug}), metric learning (Section~\ref{sec:dmb:metric}) and meta learning (Section~\ref{sec:dmb:meta}), there have also some niche discriminative FSL approaches from other perspectives.

Multi-task learning~\cite{zhang2017survey} advocates learning multiple tasks synchronously by making the upstream  embedding module implicitly or explicitly shared across tasks and the downstream task module specific in the hope of  rendering the internal representation more generic. Follow this line, several multi-task learning based FSL approaches~\cite{yan2015multi,zhang2018metagan,hu2018few} have been proposed. In~\cite{yan2015multi}, a  regularization penalty term is designed to force the parameters of different tasks to be similar. MetaGAN~\cite{zhang2018metagan} introduced a task-conditioning GAN, which generates and discriminates fake samples and casts it as an auxiliary task to sharpen the decision boundary formed by other meta learning based FSL approaches. Z. Hu \emph{et al.}~\cite{hu2018few} inserted an attribute prediction step before final class prediction and combined the attribute learning loss with the main task loss to jointly optimize the whole learner.

\begin{figure}[b!]
\centering
\includegraphics[width=0.98\linewidth]{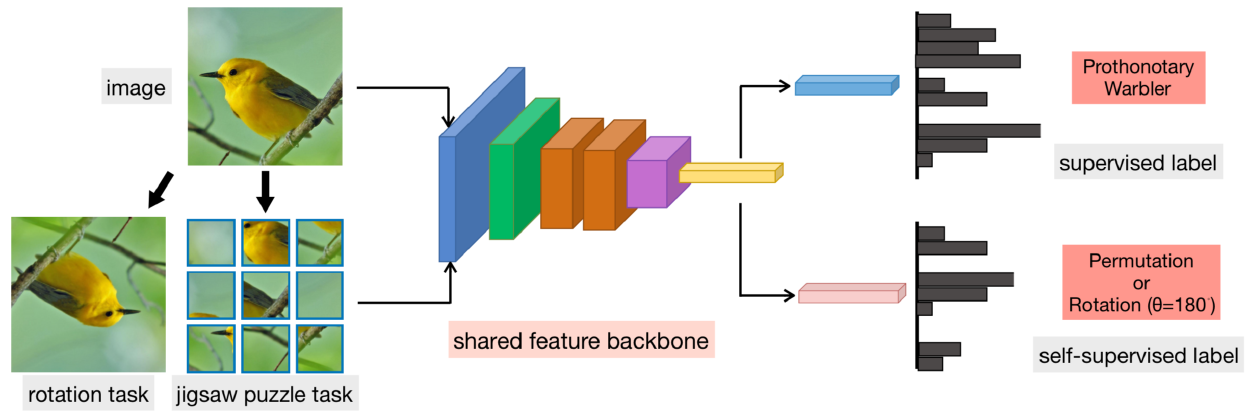}
\vspace{-0.5em}
\caption{Overview of self-supervised learning based FSL approach~\cite{su2019boosting}.}
\label{fig:SSFSL}
\end{figure}

Self-supervised learning becomes a popular technique nowadays, especially in the field of vision~\cite{doersch2015unsupervised,noroozi2016unsupervised,tung2017self,fernando2017self,gidaris2018unsupervised,noroozi2018boosting,nathan2018improvements,hendrycks2019using,kolesnikov2019revisiting,jing2020self} and language~\cite{wang2019self,kang2019recommendation,wang2019reinforced,lan2020albert}, which aims to learn semantically meaningful representations using only the inherent structural information contained by data itself instead  of expensive human labels. Researchers leverage the structural information of data as its self-supervision to train their networks. For instance, in~\cite{noroozi2016unsupervised,nathan2018improvements,noroozi2018boosting}, an unlabeled image is shuffled into several patches and the permutation of patches is treated as the self-supervision, and thus the goal of self-supervised learning is to solve jigsaw puzzles. Besides, other self-supervised learning tasks include predicting image rotation angles~\cite{gidaris2018unsupervised}, relative patch location~\cite{doersch2015unsupervised},  exemplar class of augmented samples~\cite{dosovitskiy2014discriminative}, etc. Recently, there are also several works attempting to address FSL problems in the spirit of self-supervised learning~\cite{su2019boosting,gidaris2019boosting,li2019learning,mangla2020charting}. 
For example, as shown in Fig.~\ref{fig:SSFSL},~\cite{su2019boosting} combined the supervised loss formed by the off-the-shelf Prototypical Nets~\cite{snell2017prototypical}  and the self-supervised losses formed by rotation tasks and jigsaw puzzle tasks to learn the feature representation. S2M2~\cite{mangla2020charting} leveraged self-supervised tasks (i.e., rotation prediction and exemplar prediction) and   Manifold Mixup~\cite{mangla2020charting} to regularize the feature manifold, which leads to an additional loss for  general-purpose representation. Both~\cite{gidaris2019boosting,li2019learning} focused on how to incorporate self-supervised learning into semi-supervised FSL tasks, which will be introduced in Section~\ref{sec:S-FSL}. In essence, these self-supervised FSL approaches construct auxiliary self-supervised tasks attached to the main FSL tasks, and thus they still belong to the scope of multi-task learning.

Inspired by transductive inference~\cite{vapnik1999overview}, Y. Liu \emph{et al.}~\cite{liu2019learning} assumed a transductive setting where all query samples in a task would arrive at once when testing. In this manner, Transductive Propagation Network (TPN)~\cite{liu2019learning} was developed, which realized a label propagation for all unlabeled query samples using a graph model. Besides, several works proposed to fully utilize extra available data or prior knowledge to facilitate FSL. For instance,
Z. Xu \emph{et al.}~\cite{xu2017few} drew support from large-scale machine-labeled web images, and M. Bauer \emph{et al.}~\cite{bauer2017discriminative} leveraged the concept information between different classes to build a probabilistic $K$-shot learning model. In addition, there also exist some approaches from other unique perspectives, such as feature replacement~\cite{bart2005cross}, LS-SVM-based model adaptation~\cite{tommasi2009more}, Bilevel Programming~\cite{franceschi2018bilevel}, knowledge distillation~\cite{kimura2018few}, dense classification~\cite{lifchitz2019dense} and saliency-guided data hallucination~\cite{zhang2019few}, etc.

With the recent emergence of many FSL solutions, some researchers switched their focus from method development to further analysis of existing methods. In~\cite{chen2019closer}, the effect of the base network's depth to FSL model capability was analyzed via a series of consistent comparative experiments  on several representative FSL approaches, such as Prototypical Nets~\cite{snell2017prototypical}, Matching Nets~\cite{vinyals2016matching}, Relation Net~\cite{yang2018learning} and MAML~\cite{finn2017model}. In~\cite{wang2019simpleshot}, the metric learning based FSL approaches were further explored and it was claimed that some simple feature pre-processing  (e.g., mean-subtraction and L2-normalization) could bring performance improvement.

\section{Extensional Topics}~\label{sec:extensional}
This section elaborates several emerging extensional topics of FSL including Semi-supervised FSL (S-FSL), Unsupervised FSL (U-FSL), Cross-domain FSL (C-FSL), Generalized FSL (G-FSL) and Multimodal FSL (M-FSL). Their mathematical descriptions have been presented in Section~\ref{subsec:notations}. Based on a variety of practical application environments, the five topics replan the application scenarios and task requirements of FSL, and they are becoming the hot directions of FSL researches.

\subsection{Semi-supervised Few Sample Learning}
\label{sec:S-FSL}
S-FSL postulates that the training set $D_{\mathrm{trn}}$ for an $N$-way $K$-shot task contains not only $NK$ labeled support samples, but also some unlabeled samples that belong or not belong to the $C$ task classes. Researchers are allowed to use the semi-supervised training set to build their FSL systems. 

\begin{figure}[t!]
\centering
\includegraphics[width=0.99\linewidth]{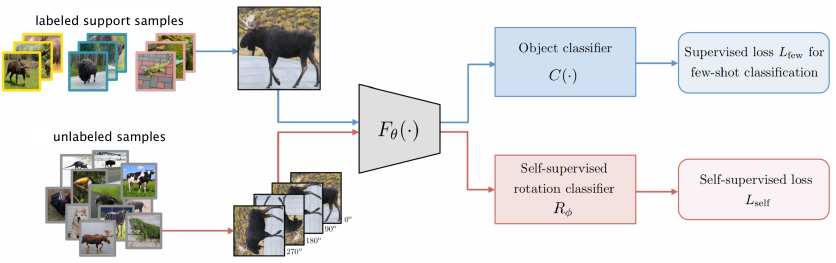}
\vspace{-0.5em}
\caption{Overview of self-supervised approach for S-FSL tasks~\cite{gidaris2019boosting}.}
\label{fig:SFSL}
\end{figure}

In~\cite{boney2017semi,ren2018meta,allen2019infinite},  Prototypical Nets~\cite{snell2017prototypical} were improved to cope with S-FSL via semi-supervised clustering. MetaGAN~\cite{zhang2018metagan} that has been discussed in Section~\ref{sec:dmb:other} was designed to be compatible with the S-FSL setting. In~\cite{gidaris2019boosting} and~\cite{li2019learning}, the self-supervised learning paradigm was exploited to ingest information from unlabeled samples. In particular, as shown in Fig.~\ref{fig:SFSL}, ~\cite{gidaris2019boosting} constructed self-supervised tasks (i.e.,  rotation prediction and relative patch location) on unlabeled images and added this self-supervised loss into the main FSL task loss, which is very similar to the approach in~\cite{su2019boosting} discussed in Section~\ref{sec:dmb:other} except that it built the self-supervised tasks using labeled support samples (see Fig.~\ref{fig:SSFSL}). Differently,~\cite{li2019learning} proposed a self-training strategy for S-FSL that iterates between predicting pseudo labels for unlabeled data and finetuning FSL models using the pseudo-labeled data.
Self-Jig~\cite{chen2019image} treated the labeled images as probe samples and the unlabeled images as gallery samples and then synthesized new images from them. Besides, several task-oriented S-FSL models have also been proposed recently, like SAMIE~\cite{wang2019semi} for question-answer tasks and  AffinityNet~\cite{ma2019affinitynet} for disease prediction tasks.

\subsection{Unsupervised Few Sample Learning}
U-FSL encourages a more general setting than vanilla FSL where the auxiliary set $D_A$ is fully unsupervised. The goal is to pursue a relatively mild condition for performing FSL and weaken the prerequisite for building an FSL learner, since collecting an unlabeled auxiliary set belonging to non-task classes is more easy-to-implement than collecting a labeled dataset. For example, one can readily acquire a large number of unlabeled images via web crawlers in today's big data era.

In~\cite{wang2016learning}, the top layers in the base learner were pre-trained as low-density separator (LDS) using the unlabeled samples, and they are encouraged to capture a more generic  representation space for downstream FSL tasks. CACTUs~\cite{hsu2019unsupervised} adopted a two-stage strategy: synthesizing meta-train tasks on the unlabeled set by unsupervised representation learning methods (e.g., ACAI~\cite{berthelot2018understanding} and BiGAN~\cite{khodadadeh2019unsupervised}) and clustering algorithms, and then running classic MAML~\cite{finn2017model} or Prototypical Nets~\cite{snell2017prototypical} on these synthetic tasks. Comparably, both UMTRA~\cite{khodadadeh2019unsupervised} and AAL~\cite{antoniou2019assume} synthesized meta-train tasks through augmenting the unlabeled samples and treating the ancestor, on which augmentation is performed, and the corresponding augmented data as the congener samples, which is followed by the ready-made MAML~\cite{finn2017model} algorithm. The commonness between CACTUs~\cite{hsu2019unsupervised}, UMTRA~\cite{khodadadeh2019unsupervised} and AAL~\cite{antoniou2019assume} is that they essentially focused on how to allocate pseudo labels to unlabeled samples such that the existing vanilla FSL models can work without modification.

\subsection{Cross-domain Few Sample Learning}
\label{sec:C-FSL}
\begin{figure}[t!]
\centering
\includegraphics[width=0.99\linewidth]{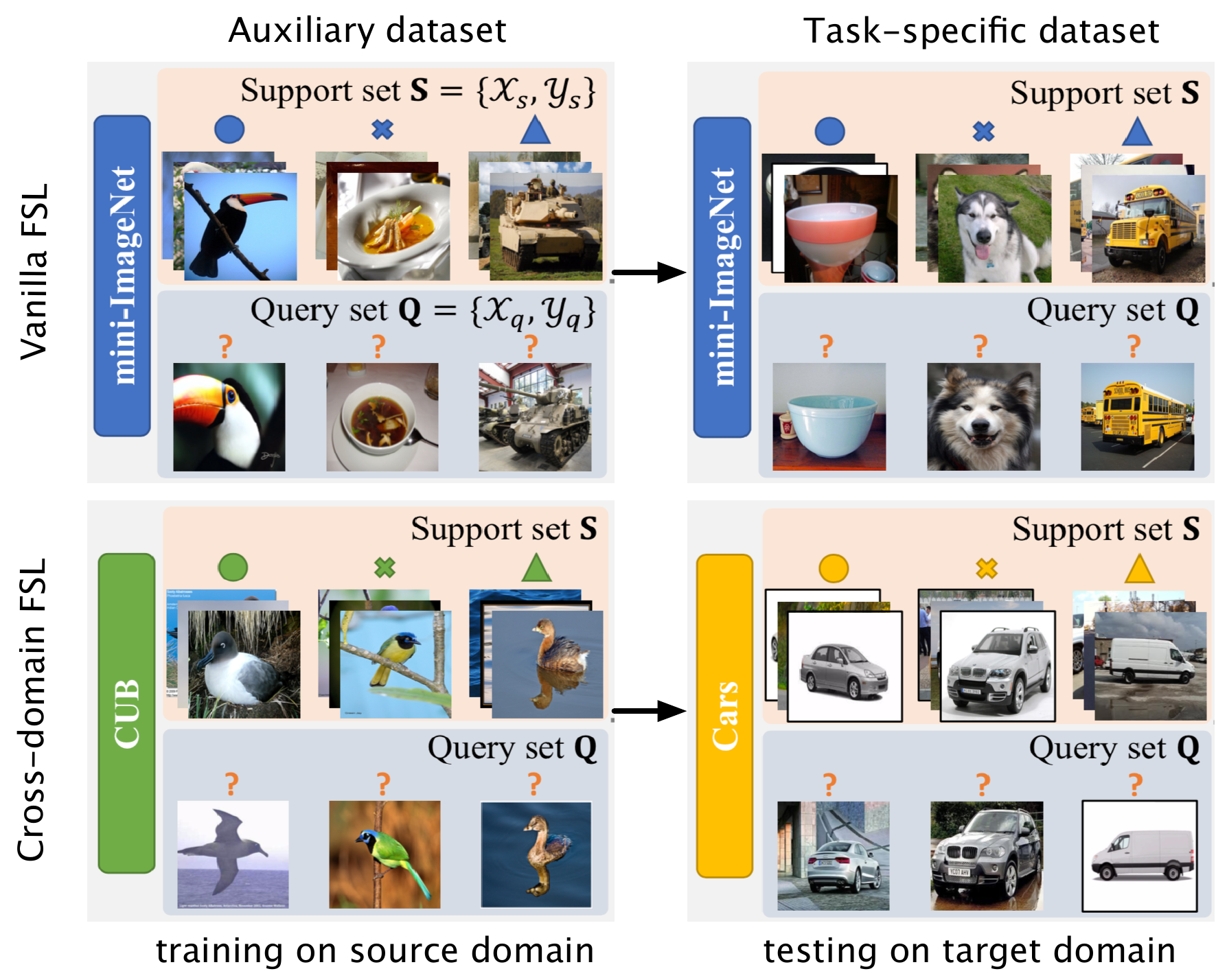}
\vspace{-0.8em}
\caption{Illustration for the task setting of Cross-domain FSL (C-FSL).}
\label{fig:C-FSL}
\end{figure}

Under the vanilla FSL setting, it is assumed that the samples in auxiliary dataset $D_{\mathrm{A}}$ and $T$-specific dataset $D_{T}$ all from the same data domain, as depicted in the top part of Fig.~\ref{fig:C-FSL}. However, when the FSL task to be handled is from a novel domain for which no relevant auxiliary samples are available, we have to leverage some cross-domain samples as the auxiliary data, as shown in the bottom part of Fig.~\ref{fig:C-FSL}. The domain shift between auxiliary dataset and task-specific dataset poses a higher challenge for FSL approaches.

C-FSL is naturally highly related to domain adaptation (DA)~\cite{patel2015visual}, which is a classic direction in the field of machine learning. Although there exists individual work~\cite{motiian2017few}  addressing DA with a few samples, its task setting is different from C-FSL: the label space in DA is shared between source and target domain, whereas that in C-FSL tasks  is disjoint between auxiliary dataset and task-specific dataset. Recently, several approaches were proposed to tackle C-FSL problems from various perspectives, such as adversarial training~\cite{luo2017label,dong2018domain,kang2018transferable,sahoo2018meta}, feature transformation~\cite{tseng2020cross,ye2020few}, domain alignment~\cite{lu2018boosting}, domain-specific finetuning~\cite{chen2019closer}, feature composition~\cite{hu2019weakly} and ensemble methods~\cite{dvornik2019diversity}, etc. To facilitate follow-up C-FSL related researches, we summarize in Table~\ref{tab:C-FSL} the specific cross-domain forms used by them.

\begin{table}[t!]
\begin{center}
\caption{Specific cross-domain forms studied by C-FSL approaches}
\vspace{-1.em}
\label{tab:C-FSL}
\scalebox{0.75}{
\begin{tabular}{ll}
\toprule[1pt]
\textbf{Approaches} & \textbf{Cross-domain Form} ($D_{A}\rightarrow D_{T}$)  \\
\cmidrule[0.5pt](lr){1-2}
~\cite{luo2017label} & SVHN~\cite{netzer2011reading} 0-4 $\rightarrow$ MNIST~\cite{lecun1998gradient} 5-9,  ImageNet~\cite{russakovsky2015imagenet} $\rightarrow$ UCF-101~\cite{soomro2012ucf101}\\
\cmidrule[0.5pt](lr){1-2}
~\cite{dong2018domain} & Omniglot~\cite{lake2015human} $\rightarrow$ EMNIST~\cite{cohen2017emnist}\\

\cmidrule[0.5pt](lr){1-2}
\multirow{4}{*}{\cite{kang2018transferable}} & \emph{Digit dataset}: MNIST~\cite{lecun1998gradient} $\rightarrow$ USPS~\cite{le1989handwritten}, MNIST $\rightarrow$ SVHN~\cite{netzer2011reading} \\
& $\qquad \qquad \qquad\ \ $USPS $\rightarrow$ MNIST, SVHN $\rightarrow$ USPS\\
\cmidrule[0.5pt](lr){2-2}
& \emph{Office dataset}\cite{saenko2010adapting}: Amazon $\rightarrow$ DSLR, Amazon $\rightarrow$ Webcam, \\
& $\qquad \qquad \qquad\qquad\quad $ DSLR $\rightarrow$ Webcam,  Webcam $\rightarrow$ DSLR\\

\cmidrule[0.5pt](lr){1-2}
\multirow{2}{*}{\cite{sahoo2018meta}} & \emph{Character dataset}: Omniglot $\rightarrow$ Omniglot-M, Omniglot-M $\rightarrow$ Omniglot \\
\cmidrule[0.5pt](lr){2-2}
& \emph{Office-Home dataset}~\cite{venkateswara2017deep}: Clipart $\rightarrow$ Product, Product $\rightarrow$ Clipart \\

\cmidrule[0.5pt](lr){1-2}
~\cite{ye2020few} & \emph{Office-Home dataset}~\cite{venkateswara2017deep}: Clipart $\rightarrow$ Real World, Real World $\rightarrow$ Clipart \\

\cmidrule[0.5pt](lr){1-2}
~\cite{tseng2020cross} & \emph{mini}ImageNet~\cite{vinyals2016matching} $\rightarrow$ CUB~\cite{wah2011caltech} / Cars~\cite{krause20133d} / Places~\cite{zhou2017places} / Plantae~\cite{van2018inaturalist}\\

\cmidrule[0.5pt](lr){1-2}
~\cite{lu2018boosting} & MNIST~\cite{lecun1998gradient} $\rightarrow$ Cifar-10~\cite{krizhevsky2009learning}, Cifar-10 $\rightarrow$ MNIST\\

\cmidrule[0.5pt](lr){1-2}
~\cite{chen2019closer} & \emph{mini}ImageNet~\cite{vinyals2016matching} $\rightarrow$ CUB~\cite{wah2011caltech}\\

\cmidrule[0.5pt](lr){1-2}
~\cite{hu2019weakly} & \emph{mini}ImageNet~\cite{vinyals2016matching} $\rightarrow$ CUB~\cite{wah2011caltech}, Kinetics-CMN~\cite{zhu2018compound} $\rightarrow$ Jester\\

\cmidrule[0.5pt](lr){1-2}
~\cite{dvornik2019diversity} & \emph{mini}ImageNet~\cite{vinyals2016matching} $\rightarrow$ CUB~\cite{wah2011caltech}\\

\bottomrule[1pt]
\end{tabular}}
\end{center}
\end{table}

\subsection{Generalized Few Sample Learning}
Vanilla FSL setting could easily lead to catastrophic forgetting issue~\cite{goodfellow2014empirical}, that is, most FSL models were trained to make inference for pre-defined classes of a novel task, but can not be continuously applied to the previous classes in the auxiliary set. However, in many applications where class concepts and samples arrive in a dynamic manner, learning systems are often faced with an extreme imbalance of training data among classes, which means some classes are provided with sufficient training samples while some have only a few. In this context, it is crucial and desirable to have the incremental learning ability for novel task classes with limited data while at the same time does not forget previous non-task classes. Thus, the focus of G-FSL is to enable FSL models to jointly handle all classes in both $D_A$ and $D_T$.

Several augmentation based FSL approaches mentioned in Section~\ref{sec:dmb:aug} including SH~\cite{hariharan2017low}, Hallucinator~\cite{wang2018low}, CP-ANN~\cite{gao2018low} and IDeMe-Net~\cite{chen2019image} are naturally applicable to both FSL and G-FSL settings since their learning processes were divided into two independent stages: first augmenting training samples for the sparse task classes and then training the models using the combination of raw and augmented samples. GcGPN~\cite{shi2019relational} extended vanilla Prototypical Nets~\cite{snell2017prototypical} to G-FSL setting using a GCN~\cite{kipf2017semi}, which models the relationship between all novel and existing classes by casting the classes as nodes and the inter-class dependencies as edges. Dynamic Nets~\cite{gidaris2018dynamic}, Acts2Params~\cite{qiao2018few}, DAE~\cite{gidaris2019generating} and AAN~\cite{ren2019incremental} adhered to a common principle of generating new weights for novel classes incrementally and combining them into the weights of existing classes to form a joint decision maker. In~\cite{rahman2018unified}, a notion of class adapting principal directions is introduced to enable efficient and discriminative embeddings for images from both novel and existing classes. 
CADA-VAE~\cite{schonfeld2019generalized} developed a Variational Auto-Encoder to create latent space features for novel classes and then trained the final predictor over all classes. The L2A approach, FEAT~\cite{ye2020few}, is also suitable for G-FSL setting since its set-to-set functions are class-agnostic and its embedding adaptation can operate on both novel and existing classes.

\subsection{Multimodal Few Sample Learning}
\begin{figure}[t!]
\centering
\includegraphics[width=1\linewidth]{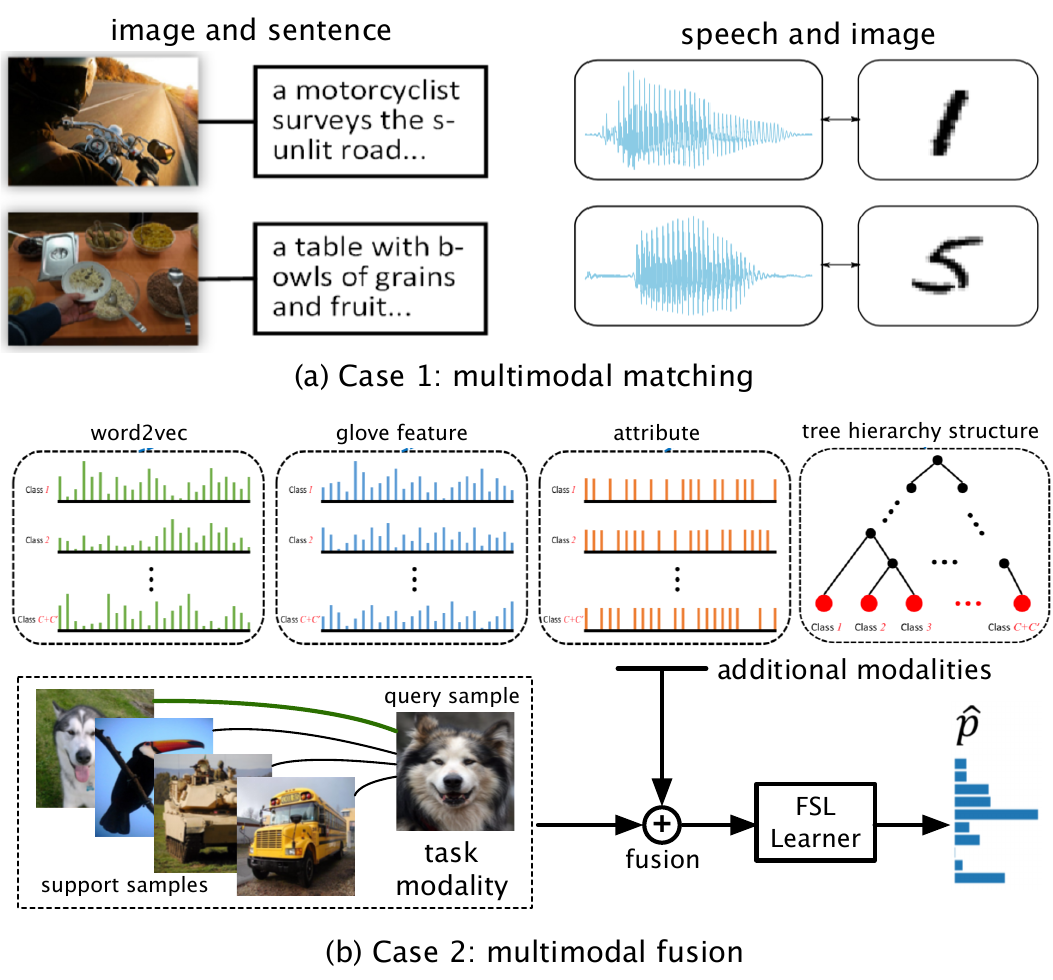}
\vspace{-2em}
\caption{Illustration for the two cases of Multimodal FSL (M-FSL).}
\label{fig:M-FSL}
\end{figure}

Different from vanilla FSL that only contains a task modality, M-FSL  involves information or data from additional modality. According to the role of additional modality, M-FSL setting can be further subdivided into two cases, as shown in Fig.~\ref{fig:M-FSL}.

\textbf{Multimodal Matching.} Vanilla FSL seeks for a mapping from the task modality to hard class label space, whereas the multimodal matching of M-FSL aims to learn the mapping from one modality to another modality~\cite{huang2019acmm,huang2019few,eloff2019multimodal}. For example, given a few image-sentence training pairs, FSL learners are required to determine the sentence correctly describing the query image~\cite{huang2019acmm,huang2019few}, or given a handful of speech-images training pairs, FSL learners need to find a correct visual image that contains the word spoken in the query speech~\cite{eloff2019multimodal}. These FSL based multimodal matching settings are meaningful, especially for robotic applications.

\vspace{0.2em}
\textbf{Multimodal Fusion.} It allows FSL learners to use extra  information from additional modalities to help the learning in task modality. Several rencent works~\cite{zhao2018dynamic,pahde2018discriminative,pahde2018cross,tsai2017improving,xing2019adaptive,fortin2019few,schonfeld2019generalized,vuorio2019multimodal,pahde2019self,li2019large} enhanced FSL model capability by fusing  various multimodal information, which include word2vec~\cite{tsai2017improving,schonfeld2019generalized}, text captions~\cite{pahde2018discriminative,pahde2018cross,pahde2019self,zhao2018dynamic}, attributes~\cite{tsai2017improving,schonfeld2019generalized,tokmakov2019learning}, glove features~\cite{tsai2017improving}, word embeddings~\cite{xing2019adaptive,fortin2019few}, tree hierarchy structure among classes~\cite{tsai2017improving,li2019large} and cross-style datasets~\cite{vuorio2019multimodal}, etc. These additional modalities bring more prior knowledge into FSL, providing a remedy for the lack of training samples.

\begin{table}[b!]
\begin{center}
\caption{Statistics of popular FSL benchmark datasets for image classification}
\vspace{-0.8em}
\label{tab:CV-datasets}
\scalebox{0.89}{
\begin{tabular}{lccc}
\toprule[1pt]
\textbf{Dataset} & \textbf{\# Images} & \textbf{\# Trn/Val/Tst cls.} & \textbf{Content}  \\
\cmidrule[0.5pt](lr){1-4}
Omniglot~\cite{lake2015human} & 129,840  & 4,800/-/1,692 & characters\\ 
\cmidrule[0.5pt](lr){1-4}
\emph{mini}ImageNet~\cite{vinyals2016matching} & 60,000 & 64/16/20 & common objects\\ 
\cmidrule[0.5pt](lr){1-4}
\emph{tiered}ImageNet~\cite{ren2018meta} & 779,165 & 351/97/160 & common objects\\ 
\cmidrule[0.5pt](lr){1-4}
CUB~\cite{hilliard2018few} & 11,788 & 100/50/50 & birds\\ 
\cmidrule[0.5pt](lr){1-4}
Stanford Dogs~\cite{li2019revisiting} &  20,580 & 70/20/30 & dogs\\ 
\cmidrule[0.5pt](lr){1-4}
Stanford Cars~\cite{li2019revisiting} &  16,185 & 130/17/49 & cars\\ 
\cmidrule[0.5pt](lr){1-4}
Caltech-256~\cite{zhou2018deep} &  30,607 & 150/56/50 & common objects\\ 
\cmidrule[0.5pt](lr){1-4}
Oxford-102~\cite{zhang2019scheduled} &  8,189 & 82/-/20 & flowers\\ 
\cmidrule[0.5pt](lr){1-4}
FC100~\cite{oreshkin2018tadam} &  60,000 & 60/20/20 & common objects\\ 
\cmidrule[0.5pt](lr){1-4}
CIFAR-FS~\cite{bertinetto2019meta} &  60,000 & 64/26/20 & common objects\\ 
\cmidrule[0.5pt](lr){1-4}
Visual Genome~\cite{fortin2019few} &  $\sim$108,000 & 1,211/-/829 & common objects\\ 
\cmidrule[0.5pt](lr){1-4}
SUN397~\cite{tokmakov2019learning} &  108,754 & 197/-/200 & scenes\\ 
\cmidrule[0.5pt](lr){1-4}
ImageNet1K~\cite{hariharan2017low} &  $\sim$1,000,000 & 389/-/611 & common objects\\

\bottomrule[1pt]
\end{tabular}}
\end{center}
\end{table}

\begin{table*}[t!]
\begin{center}
\caption{Image classification accuracy (\%) on \emph{mini}ImageNet~\cite{vinyals2016matching} of existing FSL approaches. Results are cited from there original articles, which were reported by averaging hundreds of random meta-test $C$-way $K$-shot tasks with 95\% confidence intervals (``-'': not available) }
\vspace{-1em}
\label{tab:benchmark}
\scalebox{0.93}{
\begin{tabular}{c|l|l||c|l|l}
\hline
\textbf{Approaches} & \textbf{5-way 1-shot} & \textbf{5-way 5-shot} & \textbf{Approaches} & \textbf{5-way 1-shot} & \textbf{5-way 5-shot}  \\
\hline
\hline
Matching Nets~\cite{vinyals2016matching} & 43.56 $\pm$ 0.84 & 55.31 $\pm$ 0.73 & Resnet PN~\cite{boney2017semi} & 54.05 $\pm$ 0.47 & 70.92 $\pm$ 0.66 \\ \hline
Meta-Learner LSTM~\cite{ravi2016optimization} & 43.44 $\pm$ 0.77 & 60.60 $\pm$ 0.71 & MetaHebb~\cite{munkhdalai2018metalearning} & 56.84 $\pm$ 0.52 & 71.00 $\pm$ 0.34 \\ \hline
MAML~\cite{finn2017model} & 48.70 $\pm$ 1.84 & 63.11 $\pm$ 0.92 &  STANet~\cite{yan2019dual} & 58.35 $\pm$ 0.57 & 71.07 $\pm$ 0.39  \\
\hline
MACO~\cite{hilliard2018few} & 41.09 $\pm$ - & 58.32 $\pm$ - & CSNs~\cite{munkhdalai2018rapid} & 56.88 $\pm$ 0.62 & 71.94 $\pm$ 0.57  \\ \hline
Gauss (MAP pr.) HMC~\cite{bauer2017discriminative} & 50.00 $\pm$ 0.50 & 64.30 $\pm$ 0.60 & SalNet~\cite{zhang2019few} & 57.45 $\pm$ 0.88 & 72.01 $\pm$ 0.67   \\
\hline
Meta-SGD~\cite{li2017meta} & 50.47 $\pm$ 1.87 & 64.03 $\pm$ 0.94 &  Dynamic Nets~\cite{gidaris2018dynamic} & 56.20 $\pm$ 0.86 & 72.81 $\pm$ 0.62  \\ \hline
Reptile~\cite{nichol2018reptile} & 48.21 $\pm$ 0.69 & 66.00 $\pm$ 0.62 & Dual TriNet~\cite{chen2018semantic} & 58.12 $\pm$ 1.37 & 76.92 $\pm$ 0.69 \\ \hline
MetaNet~\cite{finn2017model} & 49.21 $\pm$ 0.96 & - & Acts2Params~\cite{qiao2018few} & 59.60 $\pm$ 0.41 & 73.74 $\pm$ 0.19 \\
\hline
LLAMA~\cite{grant2018recasting} & 49.40 $\pm$ 1.83 & - & TADAM~\cite{oreshkin2018tadam} & 58.50 $\pm$ 0.30 & 76.70 $\pm$ 0.30 \\ \hline
Prototypical Nets~\cite{snell2017prototypical} & 49.42 $\pm$ 0.78 & 68.20 $\pm$ 0.66 & Deep Comparison Net~\cite{zhang2018deep} & 62.88 $\pm$ 0.83 & 75.84 $\pm$ 0.65 \\ \hline
IMP~\cite{allen2019infinite} & 49.60 $\pm$ 0.80 & 68.10 $\pm$ 0.80 & IDeMe-Net~\cite{chen2019image} & 59.14 $\pm$ 0.86 & 74.63 $\pm$ 0.74 \\ \hline
GNN~\cite{garcia2018few} & 50.33 $\pm$ 0.36 & 66.41 $\pm$ 0.63 & K-tuplet Nets~\cite{li2020revisiting} & 58.30 $\pm$ 0.84 & 72.37 $\pm$ 0.63 \\ \hline
Triplet Ranking Nets~\cite{ye2018deep} & 50.58 $\pm$ - & - & Self-Jig~\cite{chen2019image}  & 58.80 $\pm$ 1.36 & 76.71 $\pm$ 0.72 \\
\hline
mAP-Nets~\cite{triantafillou2017few} & 50.32 $\pm$ 0.80 & 63.94 $\pm$ 0.72 & CAML~\cite{jiang2019learning} & 59.23 $\pm$ 0.99 & 72.35 $\pm$ 0.71 \\ \hline
Relation Net~\cite{yang2018learning}  & 50.44 $\pm$ 0.82 & 65.32 $\pm$ 0.70 & CFA~\cite{hu2019weakly} & 58.50 $\pm$ 0.80 & 76.60 $\pm$ 0.60 \\ \hline
Cross-Modulation Nets~\cite{prol2018cross}  & 50.94 $\pm$ 0.61 & 66.65 $\pm$ 0.67 & SoSN~\cite{zhang2019power} & 59.22 $\pm$ 0.91 & 73.24 $\pm$ 0.69\\ \hline
Hyper-Represent~\cite{franceschi2018bilevel} & 50.54 $\pm$ 0.85 & 64.53 $\pm$ 0.68 & DAE~\cite{gidaris2019generating}& 61.07 $\pm$ 0.15 & 76.75 $\pm$ 0.11 \\ \hline
CovaMNet~\cite{li2019distribution} & 51.19 $\pm$ 0.76 & 67.65 $\pm$ 0.63 &  LEO~\cite{rusu2019meta} & 61.76 $\pm$ 0.08 & 77.59 $\pm$ 0.12   \\ \hline
TAML~\cite{jamal2019task} & 51.73 $\pm$ 1.88 & 66.05 $\pm$ 0.85 & AAM~\cite{hao2019instance} & 62.24 $\pm$ 0.20 & 77.24 $\pm$ 0.15 \\ \hline
Large Margin~\cite{wang2018large} & 51.41 $\pm$ 0.68 & 67.81 $\pm$ 0.64 & MTL~\cite{sun2019meta} & 61.20 $\pm$ 1.80 & 75.50 $\pm$ 0.80 \\
\hline
SARN~\cite{hui2019self} & 51.62 $\pm$ 0.31 & 66.16 $\pm$ 0.51 & EGNN~\cite{kim2019edge} & - & 76.37 $\pm$ -   \\ \hline
MT-net~\cite{lee2018gradient} & 51.70 $\pm$ 1.84 & - & Principal Characteristic Nets~\cite{zheng2019principal} & 63.29 $\pm$ 0.76 & 77.08 $\pm$ 0.68 \\\hline
MM-Net~\cite{cai2018memory} & 53.37 $\pm$ 0.48 & 66.97 $\pm$ 0.35 & AM3~\cite{xing2019adaptive} & 65.30 $\pm$ 0.49 & 78.10 $\pm$ 0.36 \\ \hline
MetaGAN~\cite{zhang2018metagan} & 52.71 $\pm$ 0.64 & 68.63 $\pm$ 0.67 & DC~\cite{lifchitz2019dense} & 62.53 $\pm$ 0.19 & 78.95 $\pm$ 0.13  \\ \hline
VERSA~\cite{gordon2019meta} & 53.40 $\pm$ 1.82 & 67.37 $\pm$ 0.86 & CC+rot~\cite{gidaris2019boosting} & 62.93 $\pm$ 0.45 & 79.87 $\pm$ 0.33  \\ \hline
BMAML~\cite{yoon2018bayesian} & 53.80 $\pm$ 1.46 & - & MetaOptNet~\cite{lee2019meta} & 64.09 $\pm$ 0.62 & 80.00 $\pm$ 0.45  \\ \hline
SNAIL~\cite{mishra2018simple} & 55.71 $\pm$ 0.99 & 68.88 $\pm$ 0.92 & CTM~\cite{li2019finding} & 64.12 $\pm$ 0.82 & 80.51 $\pm$ 0.13 \\ \hline
DA-PN~\cite{lu2018boosting} & 50.56 $\pm$ 0.85 & 69.62 $\pm$ 0.76 & LGM-Net~\cite{li2019lgm} & 69.13 $\pm$ 0.35 & 71.18 $\pm$ 0.68 \\ \hline
R2-D2~\cite{bertinetto2019meta} & 51.90 $\pm$ 0.20 & 68.70 $\pm$ 0.20 & Diversity with Cooperation~\cite{dvornik2019diversity} & 63.73 $\pm$ 0.62 & 81.19 $\pm$ 0.43 \\ \hline
TPN~\cite{liu2019learning} & 55.51 $\pm$ - & 69.86 $\pm$ - & FEAT~\cite{ye2020few} & 66.78 $\pm$ - & 82.05 $\pm$ -   \\ \hline
SRPN~\cite{mehrotra2017generative} & 55.20 $\pm$ - & 69.60 $\pm$ - & SimpleShot~\cite{wang2019simpleshot} & 64.29 $\pm$ 0.20 & 81.50 $\pm$ 0.14 \\ \hline
$\Delta$-encoder~\cite{schwartz2018delta} & 59.90 $\pm$ - & 69.70 $\pm$ - & S2M2~\cite{mangla2020charting} & 64.93 $\pm$ 0.18 & 83.18 $\pm$ 0.11$^\star$ \\ \hline
DN4~\cite{li2019revisiting} & 51.24 $\pm$ 0.74 & 71.02 $\pm$ 0.64 & LST~\cite{li2019learning} & 70.10 $\pm$ 1.90$^\star$ & 78.70 $\pm$ 0.80 \\ \hline

\end{tabular}}
\end{center}
\end{table*}

\begin{table*}[t!]
\begin{center}
\caption{Summary of FSL applications in various fields and their representative publications}
\vspace{-1.em}
\label{tab:CV-app}
\scalebox{0.9}{
\begin{tabular}{cccc}
\toprule[1pt]
\textbf{Fields} & \multicolumn{3}{c}{\textbf{Subfields \& References}}   \\
\cmidrule[0.5pt](lr){1-4}

\multirow{20}{*}{\tabincell{c}{\textbf{Computer}\\\textbf{Vision} }} 

&\multirow{12}{*}{\textbf{Image}} 

& \multirow{3}{*}{image classification}  & general image classification (see Table~\ref{tab:gmb_comparison},~\ref{tab:supervised_aug_comparison},~\ref{tab:metric_comparison},~\ref{tab:L2P},~\ref{tab:L2A},~\ref{tab:C-FSL},~\ref{tab:benchmark}, Fig.~\ref{fig:L2M},~\ref{fig:L2F}),  multi-label classification~\cite{alfassy2019laso}, \\ 
& & &fine-grained  recognition~\cite{ali2018few,pahde2018cross,pahde2019low,wei2019piecewise,huang2019low,hu2019weakly,li2019distribution,pahde2019self,li2019revisiting},\\
&  & &hyperspectral image classification~\cite{liu2018deep,zhang2020deep},  3D object/model classification~\cite{alfassy2019laso,nie20203d}\\

 \cmidrule[0.5pt](lr){3-4}
 & & \multirow{3}{*}{image segmentation}  & semantic segmentation~\cite{oliver2014one,shaban2017one,rakelly2018few,dong2018few,zhang2018sg,dong2019multi,siam2019amp,hu2019attention,zhang2019canet},  \\
 & &&instance segmentation~\cite{michaelis2018one,michaelis2018one,bhunia2019deep}, texture segmentation~\cite{ustyuzhaninov2018one,zhu2019one},\\
 & && medical/biological image segmentation~\cite{mondal2018few,zhao2019data,dietlmeier2019few,roy2020squeeze}  \\

 \cmidrule[0.5pt](lr){3-4}
 & & object  detection  & general objects\cite{dong2018fewo,chen2019few,fan2019few,karlinsky2019repmet}, air vehicles~\cite{kang2019few}, RGB-D objects~\cite{sun2018one}\\ 

 \cmidrule[0.5pt](lr){3-4}
 & &\multirow{3}{*}{other applications}    & image generation~\cite{lake2015human,rezende2016one,ali2018few,benaim2018one,antoniou2018data,clouatre2019figr,liu2019few,gao2019artistic,hong2020matchinggan}, image retrieval~\cite{triantafillou2017few,wang2017few}, \\
 &&& gaze estimation~\cite{yu2019improving}, depth estimation~\cite{li2019fewdept}, localization~\cite{wertheimer2019few},  scene graph prediction~\cite{dornadula2019visual},\\  
 && &image-based person re-identification~\cite{xiang2019incremental,xu2019feature}, image colorization~\cite{yoo2019coloring}, color constancy~\cite{mcdonagh2018formulating}\\

 \cmidrule[0.5pt](lr){2-4} 
 &\multirow{6}{*}{\textbf{Video}} 

& \multirow{2}{*}{video classification}  & general video classification\cite{yan2015multi,zhu2018compound,khodadadeh2018unsupervised,cao2019few}, gesture recognition~\cite{li2018one,lu2019one},\\
& && action recognition~\cite{mishra2018generative,xu2018dense,bishay2019tarn,hu2019weakly,dwivedi2019protogan,coskun2019domain,memmesheimer2020signal}\\ 

 \cmidrule[0.5pt](lr){3-4} 
& & video detection & action localization~\cite{yang2018one,goo2019one}, activity detection~\cite{xu2018similarity}\\

 \cmidrule[0.5pt](lr){3-4}
 & &\multirow{3}{*}{other applications}    & 
video prediction~\cite{wu2018meta,gui2018few},  video object segmentation~\cite{caelles2017one,xiao2019online}, semantic indexing~\cite{inoue2018few},\\

&&&video retargeting~\cite{lee2020metapix}, video generation~\cite{zakharov2019few}, video-based person re-identification~\cite{wu2019few}, \\

&&&object tracking~\cite{park2018meta},  motion capture~\cite{mason2018few} \\

 \cmidrule[0.5pt](lr){1-4}
 \multirow{4}{*}{\tabincell{c}{\textbf{Natural}\\\textbf{Language}\\\textbf{Processing} }} & 
 \multicolumn{3}{c}{text classification~\cite{yan2018few,yu2018diverse,liu2018few,jiang2018importance,rios2018few,bailey2018few,bao2019few,zhang2019improving,geng2019induction,sun2019hierarchical,pan2019few,schick2020exploiting}, dialogue system~\cite{vlasov2018few,madotto2019personalizing,qian2019domain},}\\

& \multicolumn{3}{c}{relation learning and knowledge graphs~\cite{han2018fewrel,xiong2018one,zhang2019fewnlp,ye2019multi,chen2019metaNLP}, word representation learning~\cite{lampinen2017one,li2018context,sun2018memory,liu2019second,hu2019few},}\\

 &\multicolumn{3}{c}{named entity recognition~\cite{hofer2018few,fritzler2019few,florez2019learning}, word prediction~\cite{vinyals2016matching,munkhdalai2018rapid,munkhdalai2018metalearning},  natural language generation~\cite{chen2019fewl,mi2019meta,peng2020few,kale2020few}, }\\

  &\multicolumn{3}{c}{information extraction~\cite{wang2019semi}, matchine translation~\cite{kaiser2017learning},  charge prediction~\cite{hu2018few}, sequence labeling~\cite{hou2019few} }\\

 \cmidrule[0.5pt](lr){1-4}
 \multirow{3}{*}{ \tabincell{c}{\textbf{Audio\&Speech} }  } & 
 \multicolumn{3}{c}{audio/speech/sound classification~\cite{lake2014one,pons2019training,zhang2019fewaudio,cheng2019multi,chou2019learning,naranjo2020open}, }\\

 &\multicolumn{3}{c}{text-to-speech~\cite{chen2019sample,moss2020boffin,zhang2020adadurian,choi2020attentron}, acoustic/sound event detection~\cite{wang2020few,shimada2020metric,shi2020fewaudio}, speech generation~\cite{lake2014one,arik2018neural}, }\\
 &\multicolumn{3}{c}{keyword/command recognition~\cite{higy2018few}, keyword spotting~\cite{seth2019prototypical}, human-fall detection~\cite{droghini2018few}, speaker recognition~\cite{anand2019few}, }\\

  \cmidrule[0.5pt](lr){1-4}
 \multirow{2}{*}{\tabincell{c}{\textbf{Reinforcement}\\\textbf{ Learning\&Robotic} }} & 
 \multicolumn{3}{c}{imitation learning~\cite{duan2017one,finn2017oneIM,yu2018one,yu2018one2,aytar2018playing,james2018task,huang2019continuous,shao2020object,bonardi2020learning}, locomotion~\cite{frans2017meta, mishra2018simple,finn2017model}, policy learning~\cite{goo2019one,vuorio2019multimodal},  }\\
  &\multicolumn{3}{c}{visual navigation~\cite{finn2017model,li2017meta,xie2018few,mishra2018simple,jamal2019task}, robot manipulation~\cite{xie2018few,xu2018neural}, multi-armed bandits~\cite{mishra2018simple}, tabular MDPs~\cite{mishra2018simple}}\\

   \cmidrule[0.5pt](lr){1-4}
 \multirow{1}{*}{\textbf{Data Analysis}} & 
 \multicolumn{3}{c}{data regression~\cite{finn2017model,li2017meta,nichol2018reptile,santoro2016meta,yoon2018bayesian,lee2018gradient,rusu2019meta,finn2018probabilistic,grant2018recasting,wu2018meta,vuorio2019multimodal}, anomaly/error detection~\cite{koizumi2019sniper,heidari2019holodetect,koizumi2020spidernet}}\\

  \cmidrule[0.5pt](lr){1-4}
 \multirow{1}{*}{\textbf{Cross-Field}} & 
 \multicolumn{3}{c}{image captioning~\cite{dong2018fast}, visual question answering~\cite{teney2018visual,dong2018fast}}\\

  \cmidrule[0.5pt](lr){1-4}
 \multirow{2}{*}{\tabincell{c}{\textbf{Other}\\\textbf{Applications} }} & 
 \multicolumn{3}{c}{disease prediction~\cite{ma2019affinitynet,prabhu2019few,paul2020fast,zhu2020alleviating,rajan2020self,ali2020additive,li2020fewdisea}, biometrical recognition (e.g., palmprint~\cite{du2019low}, ear~\cite{zhang2019few}),}\\
 & \multicolumn{3}{c}{ drug discovery~\cite{altae2017low}, spectrum classification~\cite{liu2019dynamic}, precision agriculture~\cite{li2020few}, internet security~\cite{chowdhury2017few}, mobile sensing~\cite{gong2019metasense}}\\

\bottomrule[1pt]
\end{tabular}}
\end{center}
\end{table*}

\section{Applications}~\label{sec:applications}
Since the ubiquitous demand of machine learning systems for large-scale training samples and the vigorous advances of FSL studies in recent years, the methods and ideas of FSL are being widely applied to various research areas such as computer vision, natural language processing, audio and speech, reinforcement learning and robotic, and data analysis, etc. Table~\ref{tab:CV-app} summarizes the fields and subfields of FSL applications as well as their representative publications.

\vspace{0.2em}
\textbf{Computer Vision.}
Thanks to the intuitiveness and intelligibility of visual data, computer vision has always been the main testbed for machine learning algorithms, and it is no exception to FSL. From the earliest Congealing model~\cite{miller2000learning} to today's meta learning approaches, visual tasks have always acted as the touchstone of FSL approaches, especially the few sample based (or few-shot) image classification tasks. In Table~\ref{tab:CV-datasets}, we enumerate several popular FSL benchmark datasets for image classification and summarize their statistics. Two most commonly used benchmarks are Omniglot~\cite{lake2015human} and \emph{mini}ImageNet~\cite{vinyals2016matching}. Due to the simplicity of grayscale characters images and the sufficient meta-train classes, many FSL approaches have achieved good performance on Omniglot that is close to saturation. As a result, researchers are inclined to utilize \emph{mini}ImageNet to evaluate the performance of FSL approaches. For better reference to follow-up studies, we summarize in Table~\ref{tab:benchmark} the performance of all FSL approaches that have reported their results on \emph{mini}ImageNet. We can observe that, within only three years from 2016 to 2019, the $5$-way $1/5$-shot accuracy increased by more than 20\%, which indicates the rapid development of FSL researches.
Besides, FSL has  been incorporated into image segmentation~\cite{fu1981survey}, object detection~\cite{liu2020deep} and other image-based vision tasks. At the level of  video data, FSL also has many rising applications in video classification~\cite{brezeale2008automatic}, video detection~\cite{shou2016temporal}, video object segmentation~\cite{yao2019video}, etc. More FSL applications in the vision domain can be found in the first part of Table~\ref{tab:CV-app}.

\textbf{Natural Language Processing.} It is the  second largest field of FSL applications. One common FSL application in natural language processing is text classification~\cite{korde2012text}, which seeks to utilize a few documents or words to infer document labels. In addition, the FSL regime was also brought into fundamental research topics of natural language processing, such as word representation learning~\cite{camacho2018word}, relation learning and knowledge graphs~\cite{nickel2015review}. The second part of Table~\ref{tab:CV-app} details more FSL applications in natural language processing.

\textbf{Audio and Speech.} Acoustic data is a more complex data form, and generally the large-scale collection and annotation for them are more difficult than that for images or texts, which leads to a more urgent need for FSL approaches. At present, FSL has been used to address many acoustic tasks covering from the basic audio classification and keyword recognition to the challenging text-to-speech and speech generation. The third part of Table~\ref{tab:CV-app}  summarizes existing FSL applications and corresponding FSL references.


\vspace{0.2em}
\textbf{Reinforcement Learning and Robotic.} An ideal robotic system should possess the ability of learning novel tasks with a few demonstrations and without long task-specific training time for a task, however, a new situation could make robots vulnerable to the dilemma of limited observation samples, which makes FSL an indispensable skill for future advanced robotic systems. As FSL approaches grew in popularity, many researchers have reconsidered the applications of reinforcement learning and robotic~\cite{kober2013reinforcement} under the regime of FSL, which include imitation learning~\cite{hussein2017imitation}, visual navigation~\cite{bonin2008visual} and policy learning~\cite{deisenroth2013survey}, etc. More related applications are presented in the fourth part of Table~\ref{tab:CV-app}.

 \vspace{0.2em}
\textbf{Data Analysis.}  As is well known, 
effectively analyzing data and mining underlying rules in data via sparse training data is a goal tirelessly pursued by data science researchers. Fortunately, FSL is gradually being applied to some classical data analysis applications like data regression and anomaly detection~\cite{chandola2009anomaly}, as described in the fifth part of Table~\ref{tab:CV-app}.

\vspace{0.2em}
\textbf{Cross-Field Applications.} Recently, FSL has been integrated into two popular cross-field applications, i.e., image captioning~\cite{dong2018fast} and visual question answering~\cite{teney2018visual,dong2018fast}. Given only a few image-text training pairs, the former tries to generate a proper textual description for an image, while the latter seeks to output an accurate natural language answer to a textual question about an image.

\vspace{0.2em}
\textbf{Other Applications.} Besides the above several common application fields of machine learning, FSL has also been introduced into other professional areas, such as medicine, chemometrics, agriculture, sensors and internet security, etc. Please refer to the last part of Table~\ref{tab:CV-app} for more details.

\begin{figure}[t!]
\centering
\includegraphics[width=0.99\linewidth]{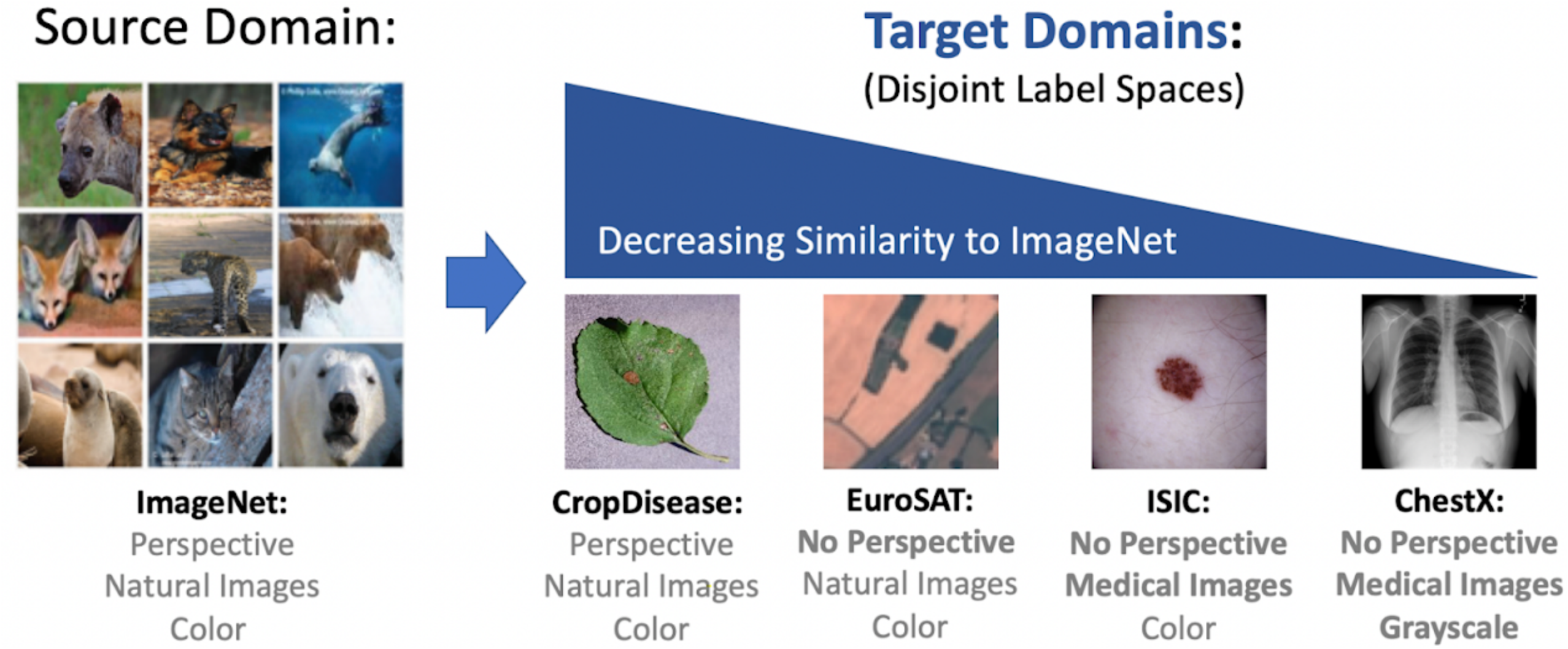}
\vspace{-0.7em}
\caption{Illustration for the cross-domain few-shot learning  competition (\url{https://www.learning-with-limited-labels.com/challenge}).}
\label{fig:Competition}
\end{figure}

\vspace{0.2em}
\textbf{Open Competitions.} With the growing attention towards FSL, several related competitions are emerging. To our best knowledge, \emph{Few-Shot Verb Image Classification} (\url{http://www.lsfsl.net/cl/}) that was  published in a workshop of ICCV 2019 is the first FSL competition, which focused on large-scale verb image classification and proposed a high-quality few-shot verb image dataset. Recently, in the Visual Learning with Limited Labels Workshop of CVPR 2020, a more challenging competition, \emph{Cross-Domain Few-Shot Learning Challenge} (\url{https://www.learning-with-limited-labels.com/challenge}), was proposed, which keeps aligned with the C-FSL tasks discussed in Section~\ref{sec:C-FSL}. As depicted in Fig.~\ref{fig:Competition}, this competition requires participants to train FSL models on ImageNet but perform evaluation on the other four datasets from varying domains, such as plant disease images, satellite images, dermoscopic images of skin lesions, and X-ray images. This competition includes two main tracks that use or not use unlabeled images from the target domain for training.










\section{Future Directions}~\label{sec:future}
Though recent years have witnessed considerable progress of FSL in both methodology and applications, challenges still exist due to the intrinsic difficulty from sparse samples. In this section, we suggest four future directions of FSL.

\vspace{0.2em}
\textbf{Robustness.}
Most of the current FSL studies are based on an ideal data hypothesis, but it is hard to hold true for all practical scenes. In many realistic applications, one may be faced with  uncertain disturbances that destroy the ideal setting of FSL. For instance, the few training data may suffer from outlier interference (e.g., noisy samples or label-wrong data)~\cite{lu2020robust} due to instrumental malfunction or perfunctory errors. It raises the question of whether existing FSL models can  effectively alleviate the influence from such outliers and still maintain an acceptable generalization. In addition, the possible domain shift between auxiliary data and task-specific data as described in~Section~\ref{sec:C-FSL} is  another kind of disturbance to the ideal setting of FSL. Therefore, improving the robustness of FSL models against  various potential disturbance factors is substantially meaningful. 

\vspace{0.2em}
\textbf{Universality.}
The universality mentioned here is twofold. The first is the model-level generality and scalability of FSL approaches. For now, most of FSL approaches are excessively  designed for the specific benchmark tasks and datasets, weakening their applicability to other more general tasks. An ideal FSL framework should be able to deal with various learning tasks with different data complexity and diverse data forms. The second is the application-level versatility and flexibility of FSL approaches.  The majority of current FSL studies focus on the plain application scenario with small-scale task classes and large-scale labeled auxiliary data. However, real-world problems may bring more complex application scenarios, such as large-scale task classes,  long-tail phenomenon of data distribution~\cite{park2008long},  dynamicity of task classes,  unavailability of labeled auxiliary data, and even a mixture of these scenarios. They raise higher requirements and challenges for the universality of FSL approaches.

\vspace{0.2em}
\textbf{Interpretability.}
The surge and success of FSL in recent years mainly lie in deep learning technology, which is often criticized for its lack of interpretability. Model interpretability is a key issue for deep learning~\cite{bau2017network,zhang2018visual}. We believe that the impressive few sample learning ability of humans benefit from many aspects including the rational use of empirical knowledge and the ingenious exploration of underlying knowledge  behind task data (e.g.,  compositional relationship~\cite{lake2015human,wong2015one}, structural correspondence between data components~\cite{lu2019self}, etc). Therefore, how to capitalize on the fusion of external prior knowledge and internal data knowledge to enhance the interpretability of FSL models could be a future research direction.

\vspace{0.2em}
\textbf{Theoretical System.}
As analyzed in Section~\ref{sec:introduction}, the fundamental difficulty caused by sparse training samples is that the search space of learning function $f$ is very huge  due to the lack of effective function regularization formed by training samples. If we re-look current FSL approaches from this theoretical view of point, it can be found that, in essence, all FSL solutions are to realize function regularization through specific technologies. For instance, the augmentation based FSL approaches reach this goal by directly increasing the training samples, while the meta learning approaches suggest introducing other irrelevant learning tasks to regularize the learning function across tasks. Thus, building  a systematic theoretical system for FSL from the perspective of regularizing learning function space under sparse training samples could bring new inspiration to FSL researchers.

\section{Conclusions}~\label{sec:conclusion}
Enabling learning systems to learn from very few samples  is crucial for the further development of machine learning and artificial intelligence. This article conducts a comprehensive survey on few sample learning (FSL). In particular, the evolution history and current advances of FSL are reviewed, and all FSL approaches are grouped via a succinct and understandable taxonomy. An in-depth analysis is made to shed light into the underlying development relationship between  mainstream meta learning based FSL approaches. Also, several emerging extensional research topics of FSL, existing FSL applications in various fields, current benchmark datasets and performance, together with several potential research directions are systematically summarized. This survey is expected to promote the grasp of FSL related knowledge and the collaborative development of FSL research area.

\ifCLASSOPTIONcompsoc
  \section*{Acknowledgments}
\else
  \section*{Acknowledgment}
\fi
The authors would like to thank the pioneer researchers in few sample learning and other related fields. 
This work was funded by the Natural Science Fundation of China (NSFC) and the German Research Foundation (DFG) in Project Crossmodal Learning, NSFC 62061136001/DFG TRR-169, NSFC 62176132.

\ifCLASSOPTIONcaptionsoff
  \newpage
\fi



%
\bibliographystyle{IEEEtran}
\bibliography{ref}

%
\begin{IEEEbiography}[{\includegraphics[width=1in,height=1.25in,clip,keepaspectratio]{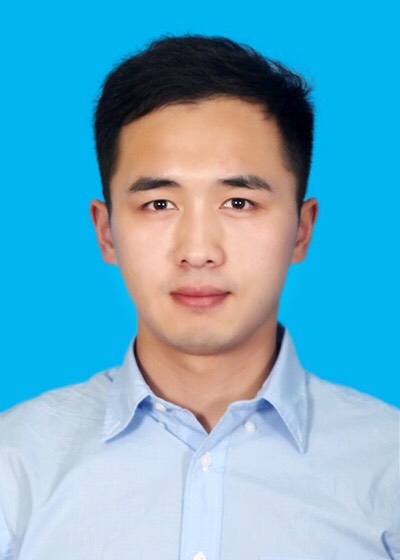}}]{Jiang Lu}
received the B.S. and Ph.D. degree  from the Department of Automation, Tsinghua University, Beijing, China, in 2013 and 2020, respectively. He is currently an Associate Research Assistant with China Marine Development and Research Center (CMDRC), Beijing, China. 

He has served as a Reviewer for the IEEE Transactions on Pattern Analysis and Machine Intelligence, and Pattern Recognition. His research interests include machine learning, deep learning and computer vision.
\end{IEEEbiography}

\begin{IEEEbiography}[{\includegraphics[width=1in,height=1.25in,clip,keepaspectratio]{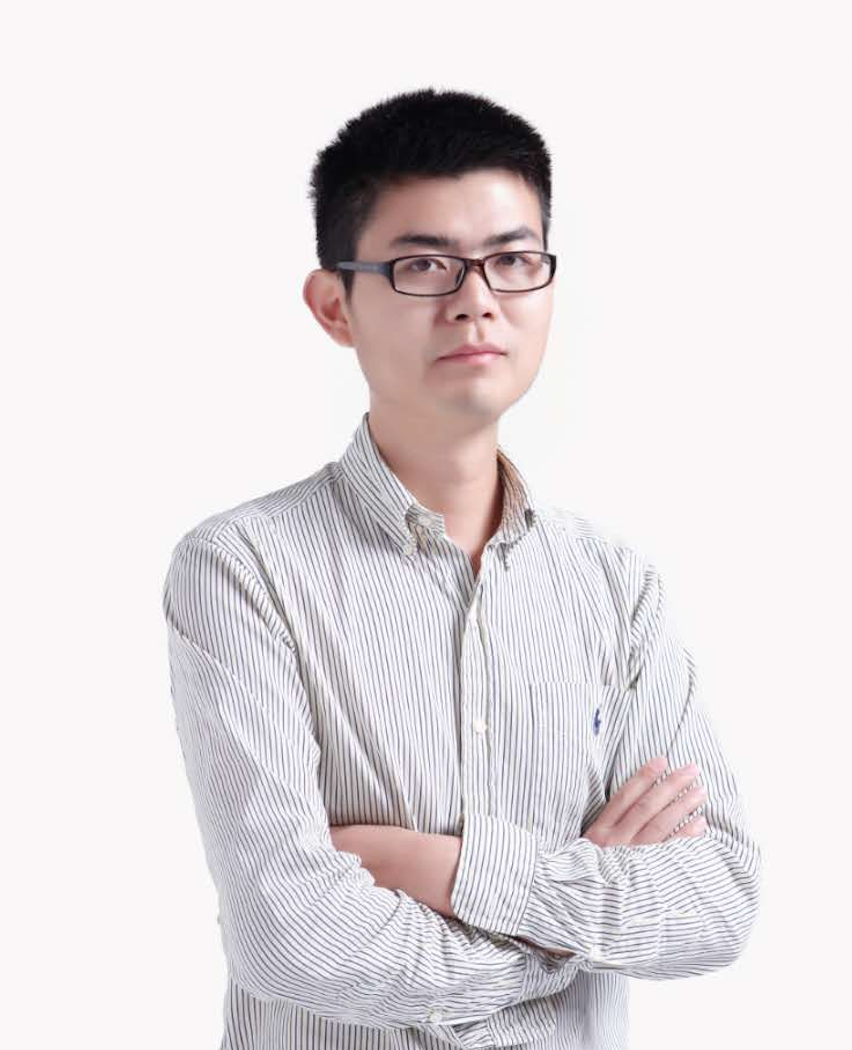}}]{Pinghua Gong}
received the B.S. degree from the Department of Automation, Xi’an Jiaotong University, Xi’an, China, in 2008, and the Ph.D. degree from the Department of Automation, Tsinghua University, Beijing, China, in 2013. 

He is currently a Distinguished Scientist \& Senior Director at Didi Chuxing, Beijing. He has published more than 20 papers in top-tier journals and conferences, including JMLR, NIPS, ICML, KDD, IJCAI, and AAAI. His research interests include machine learning and data mining.
\end{IEEEbiography}

\begin{IEEEbiography}[{\includegraphics[width=1in,height=1.25in,clip,keepaspectratio]{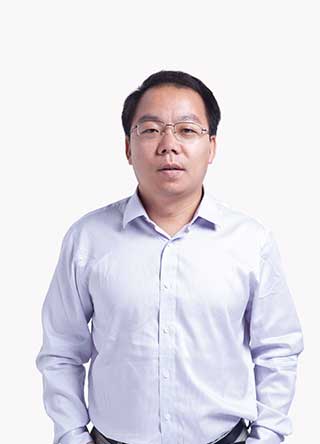}}]{Jieping Ye} (Fellow, IEEE)
received the Ph.D. degree in computer science from the University of Minnesota, Twin Cities, MN, USA, in 2005. 

He is the Head of Didi AI Labs, VP of Didi Chuxing, and a Didi Fellow. He is also a Professor with the University of Michigan, Ann Arbor, MI, USA. His research interests include big data, machine learning, and data mining with applications in transportation and bio-medicine. He was a recipient of the NSF CAREER Award in 2010. His papers have been selected for the Outstanding Student Paper at ICML in 2004, the KDD Best Research Paper Runner Up in 2013, and the KDD Best Student Paper Award in 2014. He has served as a Senior Program Committee/Area Chair/Program Committee Vice Chair of many conferences, including NIPS, ICML, KDD, IJCAI, ICDM, and SDM. He serves as an Associate Editor for Data Mining and Knowledge Discovery, the IEEE Transactions on Knowledge and Data Engineering, and the IEEE Transactions on Pattern Analysis and Machine Intelligence.
\end{IEEEbiography}

\begin{IEEEbiography}[{\includegraphics[width=1in,height=1.25in,clip,keepaspectratio]{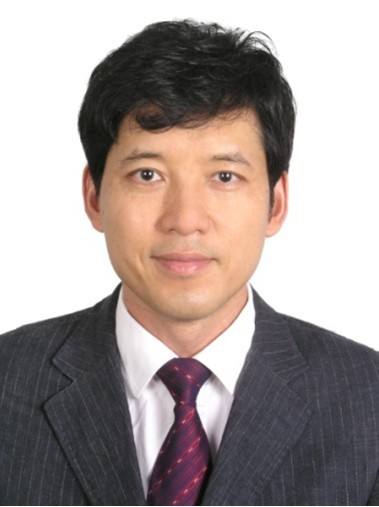}}]{Jianwei Zhang} (M’95) received the B.Eng. (Hons.) and M.Eng. degrees from the Department of Computer Science, Tsinghua University, Beijing, China, in 1986 and 1989, respectively, the Ph.D. degree from the Institute of Real-Time Computer Systems and Robotics, Department of Computer Science, University of Karlsruhe, Karlsruhe, Germany, in 1994, and the Habilitation from the Faculty of Technology, University of Bielefeld, Bielefeld, Germany, in 2000.

He is currently a Professor and the Head of the TAMS Group, Department of Informatics, University of Hamburg, Hamburg, Germany. He has published about 300 journal and conference papers (winning four best paper awards), technical reports, four book chapters, and five research monographs. He has been coordinating numerous collaborative research projects of EU and German Research Council, including the Transregio-SFB TRR 169 “Crossmodal Learning.” His current research interests include cognitive robotics, sensor fusion, dexterous manipulation,
and multimodal robot learning.
Dr. Zhang is a Life-Long Academician of the Academy of Sciences,
Hamburg. He is the General Chair of the IEEE MFI 2012, the IEEE/RSJ IROS 2015, and the IEEE Robotics and Automation Society AdCom from 2013 to 2015.
\end{IEEEbiography}

\begin{IEEEbiography}[{\includegraphics[width=1in,height=1.25in,clip,keepaspectratio]{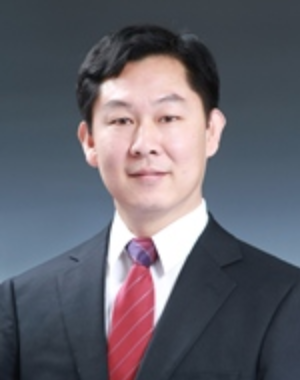}}]{Changshui Zhang} (Fellow, IEEE) received the B.S. degree in mathematics from Peking University, Beijing, China, in 1986, and the M.S. and Ph.D. degrees in control science and engineering from Tsinghua University, Beijing, in 1989 and 1992, respectively.

He is currently a Professor with the Department of Automation, Tsinghua University. His research interests include artificial intelligence, image processing and machine learning.
\end{IEEEbiography}

\end{document}